\documentclass[preprint,3p,authoryear]{elsarticle}

\usepackage[utf8]{inputenc}

\usepackage{amssymb}

\usepackage{lineno}
\modulolinenumbers[10]

\usepackage{subfigure}
\usepackage{amsfonts}
\usepackage{natbib}
\usepackage{graphicx}
\usepackage[hidelinks]{hyperref}
\usepackage{color}
\usepackage{cancel}
\usepackage{multirow}
\usepackage{blkarray}
\usepackage{subfigure}
\usepackage{ulem}
\usepackage{amssymb}
\usepackage{booktabs}
\usepackage{eurosym}
\usepackage{tcolorbox}

\usepackage{tkz-graph}
\usepackage{amsthm}
\theoremstyle{definition}
\newtheorem{theorem}{Theorem}[section]

\newenvironment{problema}[3]{
\begin{equation}
\nonumber
 {\bf #1}
\begin{array}{cll}
 \hbox{Minimize}& #2\\[0.12cm]
 \hbox{Subject to:}& #3
 \end{array}
\end{equation}}{\noindent}

\usepackage[acronym]{glossaries} 
\makeglossaries

\newacronym{acc}{Acc}{Accuracy}
\newacronym{ai}{AI}{Artificial Intelligence}
\newacronym{ampca}{AMPCA} {Arithmetic Mean Probability of Correct Assignment}
\newacronym{bag}{BAG}{Bagging Trees}
\newacronym{boost}{BOOST}{Boosting Trees}
\newacronym{cart}{CART}{Classification And Regression Trees}
\newacronym{cnn}{CNN}{Convolutional Neural Network}
\newacronym{cdf}{CDF}{Cumulative Distribution Function}
\newacronym{dcm}{DCM}{discrete choice models}
\newacronym{dnn}{DNN}{Deep Neural Network}
\newacronym[plural=GPs,firstplural=Gaussian Processes]{gp}{GP}{Gaussian Process}
\newacronym{gbdt}{GBDT}{Gradient Boosting Decision Trees}
\newacronym{gmpca}{GMPCA} {Geometric Mean Probability of Correct Assignment}
\newacronym{hpo}{HPO}{Hyperparameter Optimisation} 
\newacronym{kf}{KF}{Kernel Factory}
\newacronym{klr}{KLR}{Kernel Logistic Regression}
\newacronym{knn}{KNN}{K-Nearest Neighbors}
\newacronym{map}{MAP}{Maximum a posteriori probability}
\newacronym{ml}{ML}{Machine Learning}
\newacronym{mle}{MLE}{Maximum Likelihood Estimation}
\newacronym{ffnn}{FFNN}{Feed Forward Neural Network}
\newacronym{mnl}{MNL}{Multinomial Logit}
\newacronym{mnp}{MNP}{Multinomial Probit}
\newacronym{mxl}{MXL}{Mixed Logit}
\newacronym{nb}{NB}{Naive Bayes}
\newacronym{nl}{NL}{Nested Logit Model}
\newacronym{nn}{NN}{Neural Network}
\newacronym{pqll}{PMLE}{Penalised Maximum Likelihood Estimation}
\newacronym{rf}{RF}{Random Forests}
\newacronym{rkhs}{RKHS}{Reproducing Kernel Hilbert Spaces}
\newacronym{rnll}{RNLL}{Regularised Negative Log-Likelihood}
\newacronym{rum}{RUM}{Random Utility Model}
\newacronym{shap}{SHAP}{SHapley Additive exPlanation}
\newacronym{svm}{SVM}{Support Vector Machine}
\newacronym{vot}{VOT}{Value of Time}
\newacronym{wtp}{WTP}{Willingness To Pay}
\newacronym{xai}{XAI}{Explainable AI}
\newacronym{xgboost}{XGBoost} {Extreme Gradient Boosting}
\newacronym{tpe}{TPE}{Tree-structured Parzen Estimators}

\journal{Transportation Research Part C: Emerging Technologies}

\begin{document}

\begin{frontmatter}



\title{
A prediction and behavioural analysis of machine learning methods for modelling travel mode choice
} 


\author[inst1]{José Ángel Martín-Baos\corref{cor1}}
\ead{JoseAngel.Martin@uclm.es}
\cortext[cor1]{Corresponding author}
\author[inst2]{Julio Alberto López-Gómez}
\author[inst2]{Luis Rodriguez-Benitez}
\author[inst3]{Tim Hillel}
\author[inst1]{Ricardo García-Ródenas}

\affiliation[inst1]{organization={Department of Mathematics, Escuela Superior de Inform\unexpanded{á}tica},
      addressline={University of Castilla-La Mancha}, 
      city={Ciudad Real},
      postcode={13071}, 
      country={Spain}}
\affiliation[inst2]{organization={Department of Information and System Technologies, Escuela Superior de Inform\unexpanded{á}tica},
      addressline={University of Castilla-La Mancha}, 
      city={Ciudad Real},
      postcode={13071}, 
      country={Spain}}
\affiliation[inst3]{organization={Department of Civil, Environmental and Geomatic Engineering},
      addressline={University College London}, 
      city={London},
      postcode={WC1E 6BT}, 
      country={UK}}

\begin{abstract}

The emergence of a variety of \gls{ml} approaches for travel mode choice prediction poses an interesting question to transport modellers: which models should be used for which applications?
The answer to this question goes beyond simple predictive performance, and is instead a balance of many factors, including behavioural interpretability and explainability, computational complexity, and data efficiency. 
There is a growing body of research which attempts to compare the predictive performance of different \gls{ml} classifiers with classical \glspl{rum}. 
However, existing studies typically analyse only the disaggregate predictive performance, ignoring other aspects affecting model choice. 
Furthermore, many existing studies are affected by technical limitations, such as the use of inappropriate validation schemes, incorrect sampling for hierarchical data, a lack of external validation, and the exclusive use of discrete metrics.
In this paper, we address these limitations by conducting a systematic comparison of different modelling approaches, across multiple modelling problems, in terms of the key factors likely to affect model choice (out-of-sample predictive performance, accuracy of predicted market shares, extraction of behavioural indicators, feature importance analysis, and computational efficiency). 
The modelling problems combine several real world datasets with synthetic datasets, where the data generation function is known. 
The results indicate that the models with the highest disaggregate predictive performance (namely \gls{xgboost} and \gls{rf}) provide poorer estimates of behavioural indicators and aggregate mode shares, and are more expensive to estimate, than other models, including \glspl{dnn} and \gls{mnl}.
It is further observed that the \gls{mnl} model performs robustly in a variety of situations, though \gls{ml} techniques can improve the estimates of behavioural indices such as \gls{wtp}. 
\end{abstract}





\end{frontmatter}


\section{Introduction}
\glspl{rum} have played a central role in the analysis of individual decision making, with wide applications to economics, marketing, and transportation \citep{McFa73,BLL85,Tra09}. 
They provide simple explanations of decision processes, as well as statistical tools to test the relationships which determine the choice process. 
In recent years, \gls{ml} techniques have been studied as an alternative to \glspl{rum} for choice modelling.
Several studies have indicated that \gls{ml} models outperform \glspl{rum} in terms of their prediction capability \citep{WMH21,HaH17,ZYY20,LBD19,MGL20,ZYY20,MGR21,Omr15,SMM16}.
However, these studies only consider disaggregate predictive performance, and treat \gls{ml} classifiers as \textit{black-box} models, from which it is not possible to extract behavioural indicators to inform real-world policy decisions. 
There are therefore two aspects that have received little attention in the literature: i)~how to derive behavioural indicators from the \gls{ml} models, and ii)~how to evaluate the quality of the outcomes derived from these \gls{ml} models. 

Furthermore, many existing studies are affected by methodological pitfalls that bias their results, as indicated by a recent systematic literature review into \gls{ml} approaches for mode choice modelling \cite{HBE21}, in which the authors identify and compare the methodologies of 70 peer-reviewed articles, and in doing so identify several major limitations with existing work. 
Notably, the methodologies used across the articles identified are highly fragmented, and are affected by common methodological limitations that introduce bias in the estimation of the model performance. 

These limitations mean that it is difficult to gain a clear understanding from the literature of which approaches should be used in different modelling scenarios.
To address this, in this paper we conduct an in-depth systematic experimental comparison of state-of-the-art \gls{ml} methods that have given good results in previous studies on travel mode choice, alongside traditional \gls{dcm} models. 
The paper makes use of appropriate methodologies for data validation, and includes a diverse set of modelling scenarios composed of both real-world datasets and synthetic datasets, where the data generation function is known.
The paper also applies a homogeneous optimisation methodology for the estimation of the hyperparameters of the \gls{ml} methods.
Finally, we analyse not only the out-of-sample predictive performance of the models, but also the accuracy of predicted market shares, the extraction of behavioural indicators, and the computational efficiency of the methods.
To the best of the authors' knowledge, this is the first study that systematically compares both the predictive performance and behavioural consistency of multiple modelling approaches across \gls{ml} and utility-based models. 

The primary contributions of the paper can be summarised as follows:
\begin{enumerate}

    \item analysing, from a theoretical point of view, the difficulties faced by \gls{ml} models to derive econometric indicators such as elasticities or \gls{wtp};
 
    \item conducting a systematic assessment of the highest performing \gls{ml} models and \glspl{dcm} in modelling travel mode choice, addressing the common failures found in similar works in the literature, as identified in \cite{HBE21};

    \item studying how different model outputs, such as the choice probabilities of each alternative, market-share estimates, feature importance analysis, and latent behavioural indicators like the \gls{wtp}, can be extracted from different \gls{ml} models, and comparing their performance against \glspl{dcm};

    \item evaluating the models using both real-world datasets employed in previous comparative studies, and synthetic datasets where the data generation process (and therefore ground-truth latent variables) are known.

\end{enumerate}

A GitHub repository containing the codes and datasets developed in this paper is provided as complementary material. This repository is available under a MIT license at \url{https://github.com/JoseAngelMartinB/prediction-behavioural-analysis-ml-travel-mode-choice}.


The rest of this paper is organised as follows: Firstly, Section \ref{sect:relatedwork} details the background.
Next, Section \ref{sect:RUMvsML} provides a theoretical analysis of the \gls{rum} and \gls{ml} approaches when used with a focus on obtaining behavioural indicators. Section \ref{sect:setupexperiments} then details the setup of the experiments including the description of the methodology used to generate the datasets. Section \ref{sect:experiments} presents the results of the experiments and their discussion. Finally, the conclusions and future research are given in Section \ref{sect:conclusions}.

\section{Related Work}\label{sect:relatedwork}

The key factors when comparing \gls{ml} and \gls{dcm} methods typically discussed in the literature of travel behaviour are predictive performance and the ability of each approach to derive indicators of decision-maker behaviour. These two aspects are covered in Sections \ref{sect:background1} and \ref{sect:background2}, respectively.

\subsection{Performance analysis of ML classifiers}\label{sect:background1}
Whilst several papers look at \gls{ml} approaches for modelling travel mode choice, there is not yet an established standardised methodology for their application. 
Furthermore, datasets, model specifications, and metrics used to compare the different methods vary greatly across different studies. 
These issues limit the insights gained from meta-analysis of results from across the literature, and so we must approach the problem as a collection of largely disparate evidence that defines a certain pattern in the performance of the algorithms. 
In the \gls{ml} community, accuracy is the most widely used index for benchmark comparisons. 
A comprehensive review of the topic can be found in \cite{HBE21}.

In this section we try to establish a set of traditional classification algorithms for travel mode choice modelling. 
The starting-point of this comparison is the results obtained in other domains. 
These include predicting stock price direction \citep{BVH15} where ensemble classifiers (such as \gls{rf}, \gls{kf}, AdaBoost, \gls{boost} or \gls{bag}) are ranked in three of the four top places of the ranking. 
Regarding single classifier models (such as \glspl{nn}, \glspl{svm}, \gls{knn}, \gls{cart} or \gls{nb}) the study concluded that \gls{svm} are ranked in the second position, but the difference between this model and the \gls{rf} (which was ranked first) was not significant. These conclusions agree with the results obtained in \cite{FCB14}, with 17 families of classifiers in 171 datasets.

Several empirical studies carried out on mode choice modelling problems identify that ensemble methods and \glspl{svm} outperform \gls{mnl} \citep{ZYY20,MGL20,MGR21,ZYY20,Omr15,HaH17,SMM16} and it offers a baseline to which future approaches can be compared. One of these ensemble methods is the \gls{gbdt}, which is based on an ensemble of decision-trees. Perhaps the most well-known implementation of \gls{gbdt} is the \gls{xgboost} algorithm, which has received a great deal of attention and use because of its high predictive performance. \cite{WaR18} show that the \gls{xgboost} model has overall higher predictive performance than the \gls{mnl} model. In \cite{HBE19}, it is shown that \gls{xgboost} outperforms other \gls{ml} methods, even discrete choice models with assisted specification, but it has the drawback that it is difficult to interpret.

Nowadays, the most widely used \gls{ml} method for mode choice modelling is the \gls{nn}. \cite{KaV11} summarised 86 studies in six fields of transportation in which \gls{nn} were applied. The review of \cite{HBE21} also reveals that \gls{nn} are the most commonly applied technique. A theoretical result found in \cite{Hornik89} to motivate the use of these techniques is that \gls{nn} are a universal approximator. The \gls{dcm} determine the probability function of choosing each alternative in terms of the distribution of the error term and the specification of the utility. If the resulting function is continuous, \gls{nn} can act as universal function approximators under fairly general conditions, and a \gls{nn} with a single hidden layer and a sufficient number of hidden nodes can universally approximate arbitrary functions for the probability function and their derivatives \citep{HSW90}. \cite{RGM14} concluded that \gls{nn} have been successfully used, in particular when dealing with multidimensional non-linear data. Meanwhile, \glspl{dnn} belong to state-of-the-art classification algorithms, and \cite{RaB20,WWZ20,Wang2021} illustrate how to obtain behavioural information from these models. Because of these capabilities, \gls{dnn} should be included in any comparison of \gls{ml} methods.

Finally, numerous studies suffer from deficiencies when it comes to splitting the data into training and validation sets. \cite{Hil21} identifies that trip-wise sampling with panel data results in data leakage of the chosen alternative from the training data to the validation data for matching return/repeated/shared trips, and so violates the key principle of validating \gls{ml} classifiers with unseen out-of-sample data. 

Despite these methodological problems, the cumulative and concordant empirical findings do provide a snapshot of the phenomenon under study. For instance, \cite{WMH21} presented an exhaustive empirical benchmark for the performance of \gls{ml} and \gls{dcm} classifiers in predicting travel behaviour, with $6,970$ experiments (an experiment here consists of a combination of a set of hyperparameters, an algorithm, a sample size and a dataset). The conclusions drawn from this extensive numerical experience are close to a definitive approach, and establish the ensemble methods (including boosting, bagging, and random forests) and the \gls{dnn} as the models that achieve the highest predictive performance. Further, they conclude that discrete choice models can offer an accuracy only $3$ or $4$ percent lower than the top \gls{ml} classifiers.

In this study, we have avoided the methodological problems identified by \cite{Hil21},  such as the deficiencies in trip-wise sampling with panel data, through the implementation of appropriate methodologies for data validation that prevent data leakage when splitting the data into training and test sets.
Moreover, our comparison is not limited to prediction and computation, instead we include the perspectives of interpretability and robustness, which are critical for successfully deploying the \gls{ml} methods in practice. When studying the implications of prediction by repeating the experiments of past papers using the methodology proposed by \cite{Hil21}, we obtain similar results to those of \cite{WMH21} on the model performance. However, the results obtained for the behavioural aspect indicate that the \gls{ml} models are not as robust as \gls{dcm}.

\subsection{Behavioural analysis}\label{sect:background2}
From the point of view of behavioural analysis \glspl{dcm} have the important advantage of being interpretable, in that the regression coefficients can be used directly to calculate key behavioural indicators, such as \gls{wtp}. 
This makes \glspl{dcm} appear more attractive than \gls{ml} models, which are known for their black-box nature. Conversely, \gls{dcm} must make a priori assumptions regarding the underlying functional form of a representative utility function and the distribution of unobserved heterogeneity (error terms). Such theoretical restrictions may leave the postulated model statistically misspecified, which often leads to a sacrifice in the predictive power of the model. The \gls{ml} techniques weaken the distributional assumptions of the model to avoid potential functional misspecification. These models directly estimate the choice probabilities of alternatives without the need to define the utility function and based on this knowledge obtain the econometric indices of interest. The aim of this section is to review existing studies that have explored the possibility of extracting interpretable information from \gls{ml} models.

\cite{ZYY20} can be considered one of the few papers making a comprehensive comparison between logit models and \gls{ml} covering both prediction and behavioural analysis. These authors show that the \gls{rf} method gives the highest accuracy, however, they also show that computing marginal effects and elasticity with the standard \gls{rf} may lead to unreasonable behavioural results.

\cite{HeT00} compares artificial \gls{dnn} with \gls{nl} in a commuting mode choice problem. The research finds that the \gls{nl} model predicts more accurate aggregate market shares; however the \gls{dnn} models are more accurate at predicting individual choices at a disaggregate level. The authors conclude that there is no clear indication of which approach is better. Furthermore, \cite{WWZ20} and \cite{RaB20} show that \glspl{dnn} can provide economic indicators similar to those from classical discrete choice models. These authors interpret the hidden layers as the utility functions (multiple layers in the case of \cite{WWZ20} and a single hidden layer in the case of \cite{RaB20}) and implement the definitions of marginal effects and \gls{wtp} measures on them. 

In \cite{WWZ20}, it is shown that the use of the \glspl{dnn} involves three challenges associated with automatic learning capacity: high sensitivity to hyperparameters, model non-identification, and local irregularity. The first challenge is related to the statistical problem of balancing the approximation and estimation errors of \glspl{dnn}, the second to the optimisation challenge of identifying the global optimum in the \gls{dnn} training, and the third to the robustness challenge of mitigating locally irregular patterns of estimated functions. These issues lead to situations where the economic information obtained from a \gls{dnn} can be unreliable when the sample size is small.

\cite{WWZ20} and \cite{RaB20} have proven the difficulty of computing stable and accurate \glspl{dnn}. For unstable training algorithms, small changes in the training data set can lead to large changes in estimation results. These authors apply a bootstrapping \gls{dnn} framework, which obtains multiple estimates of the parameters through resampling with replacement. Reported estimation results are averaged across bootstrap samples.

Conversely, \cite{Wang2021} propose a theory-based residual neural network (TB-ResNet) framework. This model considers a convex combination of the utility specified by \gls{dcm} and by \gls{dnn}. The training procedure proposed for TB-ResNets is similar to the boosting method with multiple stages, in which the \gls{dcm} and the \gls{dnn} models are trained sequentially. These authors point out that TB-ResNets, when compared with a pure \gls{dnn}, can modestly improve prediction, and significantly improve interpretation and robustness, because the \gls{dcm} component in the TB-ResNets stabilises the utility functions and input gradients.

\cite{Wong2021} proposes the ResLogit model, which specifies the systematic utility as the sum of a linear utility used as \gls{dcm} and a residual utility obtained by applying a \gls{dnn} model. This latter term incorporates the non-linear cross-effects between alternatives. The model is estimated by a mini-batch stochastic gradient descent algorithm applied to the log-likelihood function, which allows for a fast and efficient optimisation of the model. These authors determined that the ResLogit model yielded a smaller standard error for each econometric model parameters than the baseline \gls{mnl} model. Moreover, the integration of residual layers reduced model sensitivity to cross-effects and heterogeneity.

\cite{MGL20} and \cite{MGR21} propose the use of the \gls{klr} method as a way of specifying non-parametric utilities in \gls{rum}. \gls{klr} enables the unbiased estimation of economic indicators such as \gls{wtp}. In the numerical results, the authors show that \gls{klr} obtains unbiased \gls{wtp} estimates for both linear and non-linear utility functions, whilst the linear \gls{mnl} only obtains unbiased \gls{wtp} estimates for linear utility functions.

This section will end by linking the interpretability of \gls{ml} models, an increasingly popular approach to address the problem of opacity in black-box \gls{ml} models. It should first be clarified that \gls{rum} can be considered as inherently interpretable, as the importance of a variable in the choice of a certain mode of transport can be identified by analysing the sign and magnitude of the associated parameters and, furthermore, statistical tests can be conducted on the parameters. However, the issue of interpretability poses a challenge for \gls{ml} models, which are typically considered black-box models. In response, \gls{xai} is currently experiencing a significant development through the creation of transparent AI systems. Two paths of development can be identified: (i) directly conveying the workings of the model (so-called transparency), and (ii) justifying how/why the model arrived at its predictions (so-called post-hoc explanation). 
Regarding \gls{ml} applications in transport, \cite{SDA22} employ permutation-based feature importance to provide an understanding of the scale of the impact of different variables for the decision-making process, in terms of the contribution to predictive performance. However, there exist techniques which further indicate both the direction and scale of the impact. For example, \citet{WaJ22} use an explainable model on Bayesian Networks to explain travel behaviour. Also, in \citet{Li23}, big trip data is integrated with \gls{ml} and \gls{xai} to understand the factors that influence willingness to take shared rides. To conclude, in a similar way to \gls{rum}, \gls{xai} methods offer insights into the underlying mechanisms of the \gls{ml} model's decision-making process, and can provide an estimate of the importance, direction, and magnitude of each feature (or variable) in constructing the classifier.

\section{Theoretical analysis of RUM and ML approaches for obtaining behavioural indicators}
\label{sect:RUMvsML}
In this section we point out the peculiarities of \gls{rum} and \gls{ml} approaches for performing a behavioural analysis. Furthermore, Section \ref{sect:learning} aims to address the learning problem that emerges in \gls{ml} systems and the different sources of error that come up during the development of these models and Section \ref{sect:DerivationEconometric} introduces a theoretical analysis of the training problem of the models in relation to the derivation of econometric metrics.

\subsection{RUM vs ML approach}
In \gls{rum} the utility function is defined for a decision maker $n$ when choosing an alternative $i$ from the choice set $C$ and is given by
\begin{equation}
  \label{eq_U_in}
  U_{ni} = V_{i}(\mathbf{x}_{ni}) + \varepsilon_{ni},
\end{equation}
where $V_{ni}$ is the deterministic (also called systematic) component of the utility that depends on a set of attributes measured in the alternative $(\mathbf{x}_{ni})$, and $\varepsilon_{ni}$ is the unobserved component, which is a random term used to include the impact of all the unobserved variables on the utility function. Hence, the probability that a decision maker $n$ chooses an alternative $i$ from the choice set $C$ is
\begin{equation}
  \label{eq_P_i}
  \mathbb{P} (i | \mathbf{x}_n) = \mathbb{P}\left ({ U }_{ ni }\ge { U }_{ nj }\quad \forall j\in C \right )= \mathbb{P}\left ({ U }_{ ni }=\max _{ j\in C}{ U }_{ nj } \right).
\end{equation}
where $\mathbf{x}_{n}$ denotes the set of features measured for individual $n$.

Some assumptions are necessary to make the random utility model operational. The hypothesis about the error distribution $\varepsilon_{ni}$ determines the probability of choosing each alternative. \glspl{rum} calculate the probability by finding an expression dependent on the systematic utility, that is
\begin{equation}
\label{eq:parametrico}
 \mathbb{P} (i | \mathbf{x}_n) = F_i(V_1 ({ \mathbf{x}_{n1}}),\ldots, V_{|C|} ({ \mathbf{x}_{n|C|}})),
\end{equation}
\noindent where $|C|$ is the cardinality of the set $C$.

There exist several types of \glspl{rum}, depending on error distribution and structure of the data. In this paper, we are going to consider the \gls{mnl} model. This model assumes that $\varepsilon_{ni}$ is independently Gumbel distributed with variance $\sigma^2=\frac{\pi^2}{6} \cdot \beta^2$, where $\beta$ is the Gumbel scale parameter and the location parameter is set such that the expected value of the Gumbel error term is zero. Therefore, in this case the probability of each alternative is given by the expression:
\begin{equation}
  \label{eq2}
  \mathbb{P} (i | \mathbf{x}_n)=\frac{ \exp(\frac{1}{\beta}V_{i}({ \mathbf{x}_{ni}}))}{\sum_{j\in C} \ \exp(\frac{1}{\beta} V_{j} ({ \mathbf{x}_{nj}}))}.
\end{equation}

\gls{ml} models compute the probability that the decision maker $n$ chooses the alternative $i$ directly without building the systemic utility functions $V_i(\mathbf{x})$. Then, \gls{ml} methods are distribution-free approaches\footnote{The term ``distribution-free'' pertains to the explicit absence of probabilistic assumptions in ML models, notwithstanding the fact that some models do implicitly assume probabilities. A case in point is when neural networks employ a softmax function in the output layer, thereby encoding Gumbel distributions (as a logit model) within the model structure.} that avoid defining $F_i$ and $V_j$ in Equation~(\ref{eq:parametrico}) by trying to estimate the compound function directly:
\begin{equation}
\label{eq:ML}
   \mathbb{P} (i | \mathbf{x}_n)=s_i({\bf x}_n).
\end{equation}

The expression shown in Equation~(\ref{eq:ML}) relieves the modeller of the task of providing an a priori expression for the systematic utility function ${V}(x)$, and the non-linear relationship between attributes $\mathbf{x}_n$ and probabilities $\mathbb{P} (i|\mathbf{x}_n)$ is inferred from the data. On the other hand, having an expression of the utility function allows analytical work to be done to obtain information about the behaviour of the individual $n$. A second difference between the two approaches is that \gls{ml} methods use all available information $\mathbf{x}_{n}$ to estimate the probability of the class $i$, while the \gls{rum} methods are restricted to the information dependent on the alternative and the characteristics of the individual, $\mathbf{x}_{ni}$.

Some \gls{ml} focus exclusively on classification tasks and do not provide estimates of the probability of choosing a certain class. Others, even when they provide estimates of the probabilities, are not especially accurate and require calibration. To obtain added value from these models, it is necessary to estimate the probabilities accurately. Once the parameters of the model (\gls{rum} or \gls{ml}) have been estimated, they are used to obtain the econometric indices that describe the phenomenon under study. Several econometric indices are based on probability information (e.g. market shares), other indices use information from the gradient of the probability function (e.g. elasticities) or from the utility function (e.g. \gls{wtp} or \gls{vot}). Table \ref{tab:indicators} shows the definition of some of these indices, and the form they take for logit models with linear utilities. In this table, the notation $P_i$ is used to refer to the probability $\mathbb{P}(i | \mathbf{x})$.

\begin{table}[h]
    \caption{Definition of economic indicators using linear \gls{mnl} models}
    \label{tab:indicators}
    \centering
    \begin{tabular}{lcc}
    \hline
    {\bf Description} & {\bf Definition} & {\bf Linear  MNL}\\\hline
         \multicolumn{1}{p{5cm}}{
        \vspace*{0em}{Marginal effects of feature $x_{ik}$ for alternative $j$}
       }
         & 
         \multicolumn{1}{p{3cm}}{
        \vspace*{-1.5em}$$M_{j,ik}=\frac{\partial P_{j}}{\partial x_{ik}}$$
       }
         &
        \multicolumn{1}{p{5cm}}{
        \vspace*{-1.5em}$$ =\left \{ \begin{array}{cc}
         P_{i}\left ( 1-P_{i}\right )\beta_k & \hbox{ if } j=i\\
         -P_{j} P_{i} \beta_k & \hbox{ if } j \ne i\\
         \end{array}
         \right.$$
       }
         \\ 
        \multicolumn{1}{p{5cm}}{
        \vspace*{0em}{Arc elasticity of feature $x_{ik}$ for alternative $j$}
       }
         & 
         \multicolumn{1}{p{3cm}}{
        \vspace*{-1.5em}$$E_{j,ik}=\frac{\partial P_{j}\left /\partial x_{ik} \right .}{P_{j}}$$
       }
         &
        \multicolumn{1}{p{5cm}}{
        \vspace*{-1.5em}$$ =\left \{ \begin{array}{cc}
        \left ( 1-P_{i}\right )\beta_k & \hbox{ if } j=i\\
         -P_{i}  \beta_k & \hbox{ if } j \ne i\\
         \end{array}
         \right.$$
       }
         \\
        \multicolumn{1}{p{5cm}}{
        \vspace*{0em}{Willingness-to-pay for the attribute $k$ for alternative $i$}
       }
         & 
         \multicolumn{1}{p{3cm}}{
        \vspace*{-1.5em}$$WTP_{ik}=\frac{\partial V_{i}\left /\partial x_{ik} \right .}{\partial V_{i}\left /\partial I_{i} \right .}$$
       }
         &
        \multicolumn{1}{p{5cm}}{
        \vspace*{-1.5em}$$ =\frac{\beta_k}{\beta_I}$$
       }
         \\
        \multicolumn{1}{p{5cm}}{
        \vspace*{0.5em}{Value of time for alternative $i$ ($t$ time and $c$ cost)}
       }
         & 
         \multicolumn{1}{p{3cm}}{
        \vspace*{-1.5em}$$VOT_{i}=\frac{\partial V_{i}\left /\partial t_{i} \right .}{\partial V_{i}\left /\partial c_{i} \right .}$$
       }
         &
        \multicolumn{1}{p{5cm}}{
        \vspace*{-1.5em}$$ =\frac{\beta_t}{\beta_c}$$
       }
         \\\hline
    \end{tabular}
\end{table}

In the experiments (Section \ref{sect:experiments}), we consider one of the most important marginal effects, the so called \gls{wtp}, which is often considered in discrete choice modelling (see \cite{McF96}). According to behavioural economics, a consumer's \gls{wtp} is the highest price at or below which they will unquestionably purchase one unit of an amenity. They are commonly estimated as the ratio of the marginal utility of a particular choice attribute $x_{ik}$ to the marginal utility of income $I_i$ (or other form of economic payment/incentive) (see \cite{RaB20}). The mathematical expression of this index is:
\begin{equation}
\label{eq:def-wtp}
WTP_{ik}=\frac{\partial V_{i}\left /\partial x_{ik} \right .}{\partial V_{i}\left /\partial I_{i} \right .}
\end{equation}
which is usually estimated econometrically when linear utilities are employed as the ratio of two model parameters (e.g. $\beta_k/ \beta_I$) where $k$ represents the choice attribute of interest and $I$ represents the income.

It is noted that Equation~(\ref{eq:def-wtp}) depends on the utility function, and therefore it would not be a priori calculable using \gls{ml} methods. This definition can also be made operational for \gls{ml} models by means of the expression:
\begin{equation}
\label{eq:wtp-trick}
\frac{\partial P_{i}\left /\partial x_{ik} \right .}{\partial P_{i}\left /\partial I_{i} \right .}=
\frac{ \cancel{\partial P_{i}\left /\partial V_{i} \right .}\cdot \partial V_{i}\left /\partial x_{ik} \right .}{\cancel{\partial P_{i}\left /\partial V_{i} \right .}\cdot \partial V_{i}\left /\partial I_{i} \right .} = \gls{wtp}_{ik}.
\end{equation}

The expression in Equation~(\ref{eq:wtp-trick}) shows that \gls{wtp} can be computed from the derivatives of $P_i$. This makes it possible to compute \gls{wtp} from the derivatives of the \gls{ml} models.

\subsection{The learning problem and error sources}\label{sect:learning}
This section aims to address the learning problem that arises in \gls{ml} models and the diverse sources of error that emerge during the construction of these models. We denote $f$ as the operational model that is assumed to hold in the empirical domain, while $h$ represents the classifier learned through our study. We assume that $f$ emerges from latent probability functions, which unfortunately evade direct empirical observation. Furthermore, we assume that the classifier $h$ has a functional expression analogous to that of $f$, reflecting the underlying structural consistency between the two. Formally, a classifier can be abstracted as a function $h$ (the so-called prediction function) which, given a sample $\mathbf{x}_n$, and depending on a parameter vector $\boldsymbol{\omega}$, assigns a label $i\in C$ to this sample, where $C$ is the set of classes of the classification problem. It can be formalised as $h(\mathbf{x}_n;\boldsymbol{\omega}) \in C$. Mathematically, these models are expressed as:
\begin{eqnarray}
    f(\mathbf{x}) = \underset{i\in C }{ \hbox{arg max } } P_i(\mathbf{x}),\\
    h(\mathbf{x; \boldsymbol \omega}) = \underset{i\in C }{ \hbox{arg max } } s_i(\mathbf{x};\boldsymbol \omega).
 \end{eqnarray}

In this supervised learning process, the target function $f(\mathbf{x})$ is defined on a domain $\cal X$ and should be learned from a finite set of data samples $\{\mathbf{x}_n, y_n)\}_{n=1}^N$. At this stage, the goal is to find the prediction function that minimises the losses incurred from inaccurate predictions. For this purpose, a given loss function $\ell$ is assumed, that, given an input-output pair $(\mathbf{x}, y)$, yields the loss $\ell(h(\mathbf{x}; \boldsymbol{\omega}), y)$ when $h(\mathbf{x}; \boldsymbol{\omega})$ and $y$ are the predicted and true outputs, respectively. Three parameter vector estimations, obtained by minimising the risk function, either for the population or for empirical data, are identified:
\begin{eqnarray}
\label{eq:trainning}
\boldsymbol \omega_D &=&\underset {\boldsymbol \omega } {\hbox{arg min }} R_D(\boldsymbol{\omega}):=\mathbb{E}_{(\mathbf{x},y) \sim \mathbb{P}({\cal X},C)} \,\, \ell (h(\mathbf{x}; \boldsymbol \omega), y),\\
\boldsymbol \omega_S&=& \underset {\boldsymbol \omega } {\hbox{arg min }}  R_S(\boldsymbol \omega):=   \frac{1}{N} \sum_{n=1}^N \ell (h(\mathbf{x}_n; \boldsymbol \omega),y_n),\\
 \label{eq:training_problem}
  \boldsymbol \omega_*&=& \underset{\boldsymbol{\boldsymbol{ \omega}}}{\hbox{arg min }} R_{S^*}(\boldsymbol{\omega}):= \frac{1}{N} \sum_{n=1}^N \ell (h(\mathbf{x}_n;\boldsymbol{\omega}), y_n) +\frac{\lambda}{2} \Omega(\boldsymbol{\omega}).
\end{eqnarray}

The overall inference error  is estimated by $R_D(\boldsymbol \omega_D)$, where  $(\mathbf{x},y)\sim \mathbb{P}({\cal X},C)$ indicates that attributes and choices are randomly distributed throughout the space ${\cal X}\times C$.  In reality, both the set  $\cal X$ and its corresponding probability distribution $\mathbb{P}({\cal X})$  are unknown, as well as the marginal distribution $\mathbb{P}(i|\mathbf{x})=P_i(\mathbf{x})$.   Only a finite set of sample points  $\{(\mathbf{x}_n,y_n)\}_{n=1}^N$, obtained in accordance with the aforementioned distribution, is available. For this reason, the empirical loss $R_S(\boldsymbol \omega)$ is constructed solely based on available data, and subsequently minimised in an attempt to obtain $h(\mathbf{x};\boldsymbol \omega_S)$, instead of minimising the population loss $R_D(\boldsymbol \omega)$ to obtain $h(\mathbf{x};\boldsymbol \omega_D)$. 

In practice, it is the regularised empirical risk that is minimised, which involves augmenting the empirical risk with a {\sl regularisation term} $\Omega(\boldsymbol{\omega})$, thereby obtaining $R_{S^*}(\boldsymbol{\omega})$.  This formulation allows many estimation methods to be described, defined by the choice of functions $\ell$ and $\Omega$.

The estimator $\boldsymbol \omega_*$  cannot be considered a global minimum of the empirical risk for two reasons: firstly, due to the presence of the penalty term, and secondly, owing to the choice of optimisation algorithm employed.  

The model actually learned for inferring $f(\mathbf{x})$ is $h(\mathbf{x};\boldsymbol \omega_*)$, and its associated inference error is measured by $R_D(\boldsymbol \omega_*)$. Quantifying $R_D(\boldsymbol \omega_*)$ is crucial to assess the performance of the learned model $h(\mathbf{x};\boldsymbol \omega_*)$, as $R_D(\boldsymbol \omega_*)$ represents the expected inference error over all possible data samples. A bound on this value is:
\begin{eqnarray}
\nonumber
    R_D(\boldsymbol \omega_*) \le \underbrace{R_D(\boldsymbol \omega_D)}_{\text{approximation}} +  \underbrace{[R_S(\boldsymbol \omega_*)-R_S(\boldsymbol \omega_S)]}_{\text{optimisation}}+ \\
    \underbrace{[R_D(\boldsymbol \omega_*)-R_S(\boldsymbol \omega_*)]+[R_S(\boldsymbol \omega_D)-R_D(\boldsymbol \omega_D)]}_{\text{generalisation}}.
    \label{eq:bound}
\end{eqnarray}

The above inequality follows from the fact that $R_S(\boldsymbol \omega_D)-R_S(\boldsymbol \omega_S)\ge 0$, since $\boldsymbol \omega_S$ is a global optimiser of the empirical risk function. Note that in practice, $\boldsymbol{\omega}_*$ is used instead of $\boldsymbol\omega_{S}$ because, although the so-called optimisation error in (\ref{eq:bound}) of this estimator is greater, its generalisation error is much lower.

Several studies have indicated that very deep neural networks possess greater expressive power than their shallower counterparts (i.e. \gls{dnn} with fewer layers), resulting in lower approximation errors. This characteristic potentially explains the superior performance of \gls{dnn} across various applications. Empirical evidence has indeed demonstrated the advantages of deep networks over their shallower counterparts in computer science, see \citet{BiS14}. However, in the domain of scientific computing, empirical experience has revealed that shallower but wider networks (i.e. \gls{dnn} with fewer layers but more nodes in each layer) exhibit superior performance compared to deep and narrow neural networks, as demonstrated in the work of \citet{LKS18}. Some theoretical understanding of this is provided in \citet{LTR17}. The reasoning behind this phenomenon can be understood by considering the constraints outlined in Equation~(\ref{eq:bound}). These constraints suggest that when data is scarce, training deeper networks can become more challenging, leading to higher optimisation and generalisation errors.

\subsection{Theoretical analysis of the training problem regarding the derivation of econometric indicators}\label{sect:DerivationEconometric}
To obtain appropriate econometric indicators from an estimated model $s_i(\mathbf{x}; \boldsymbol{\omega }_*)$, it is necessary to ensure that the model satisfies the following two essential conditions:
\begin{eqnarray}
\label{eq:probabilidadad_cota}
\left |P_i (\mathbf{x})-s_i(\mathbf{x}; \boldsymbol{\omega }_*) \right | \le  \varepsilon_0, \,\, \hbox{ for all } i\in C,\\
\left |\frac{\partial P_i (\mathbf{x})}{\partial x_{ik}} - \frac{\partial s_i(\mathbf{x}; \boldsymbol{\omega }_*)}{\partial x_{ik}} \right | \le \varepsilon_1, \,\, \hbox{ for all } i\in C.
\label{eq:derivada_cota}
\end{eqnarray}

The condition in Equation~(\ref{eq:probabilidadad_cota}) enables estimates of market shares of order $O(\varepsilon_0)$ to be derived, whereas the condition in Equation~(\ref{eq:derivada_cota})  allows for the estimation of elasticities and \gls{wtp}  of order  $O(\varepsilon_1)$. This relationship can be observed by setting $e_k=O(\varepsilon_1)$ and $e_I=O(\varepsilon_1)$ and examining the resulting estimates.

\begin{equation}
   WTP_{ik}= \frac{\frac{\partial P_i}{\partial x_{ik}}}{\frac{\partial P_i}{\partial I_{i}}}= \dfrac{\frac{\partial s_i}{\partial x_{ik}}+e_k}{\frac{\partial s_i}{\partial I_{i}}+e_I}\approx \dfrac{\frac{\partial s_i}{\partial x_{ik}}}{\frac{\partial s_i}{\partial I_{i}}}+ \dfrac{e_k \frac{\partial s_i}{\partial I_{i}}-e_I \frac{\partial s_i}{\partial x_{ik}} }{(\frac{\partial s_i}{\partial I_{i}})^2}=
   \dfrac{\frac{\partial s_i}{\partial x_{ik}}}{\frac{\partial s_i}{\partial I_{i}}}+ O(\varepsilon_1).
\end{equation}

Hence, it is important to consider these conditions carefully when conducting empirical analyses in the context of demand modelling, as they can have significant implications for the accuracy of the resulting estimates. By ensuring these two conditions are met, researchers can derive reliable and informative insights into consumer behaviour. The issue addressed in this section is whether the problem of learning the target function $f(\mathbf{x})$ is sufficient to ensure both conditions. The following result analyses the approximation error $R_D(\boldsymbol \omega_D)$ for an arbitrary \gls{ml} model $s_i$:
\begin{theorem}
\label{th1}
Let $\mathbb{P}(\mathbf{x},y)$ be  an arbitrary joint probability distribution on  ${\cal X } \times C$. Let us assume that  its  marginal distribution $\mathbb{P}(y=i|\mathbf{x})$ is given by the functions $P_i(\mathbf{x})$ that are continuous on the compact set $\cal X$ and that the probability density function of the random variable $\mathbf{x}$ is a certain function $\psi(\mathbf{x})$ that is bounded on ${\cal X}$.

Let it be further assumed that the set ${\cal X}_o=\{\mathbf{x} \in {\cal X}: \exists \,\, i, j \hbox{ with } i\ne j  \,\, \hbox{ and }P_i(\mathbf{x})=P_j(\mathbf{x})\}$ has measure zero. Suppose that for any $\varepsilon >0$ there exists a parameter vector $\boldsymbol \omega$ such that:
\begin{eqnarray}
\left |P_i (\mathbf{x})-s_i(\mathbf{x}; \boldsymbol{\omega }) \right | \le  \varepsilon,\,\,  \hbox{ for all } i \in C,
\end{eqnarray}
then we have $R_D(\boldsymbol \omega_D) =\mathbb{E}_{(\mathbf{x},y) \sim \mathbb{P}({\cal X},C)} \,\, \ell (f(\mathbf{x}), y)$. 
\end{theorem}
\begin{proof}
See \ref{App:Proof-th1}.
\end{proof}

A corollary from Theorem \ref{th1}, is that the criteria for choosing a \gls{ml} model to perform a given demand analysis should not only rely on its performance but also include its ability to be suitable for robust and reliable economic information extraction. This matter is examined through numerical analysis in Experiment $3$, detailed in Section \ref{sect:Exp3}. However, we introduce a short theoretical discussion below.

Assume an ideal situation with a {\sl successful} optimisation algorithm that always finds the global optimum $\boldsymbol{\omega}_*=\boldsymbol{\omega}_S$. Moreover, suppose that the hypothesis space ${\cal H}=\{ h(\cdot, \boldsymbol{\omega}):{\cal X} \mapsto C \}$ has a Vapnik-Chervonenkis dimension of $d_{VC}$. Then, with probability of at least $1-\delta$, the following holds \citep{Vap98}:
\begin{eqnarray}
    |R_D(\boldsymbol \omega_*)-R_S(\boldsymbol \omega_*)|+|R_S(\boldsymbol \omega_D)-R_D(\boldsymbol \omega_D)|  \le  2 \sqrt{\frac{d_{VC}}{N} \log \left ( \frac{N}{d_{VC}} \right ) -\frac{1}{N} \log \delta}.
\end{eqnarray}
In this ideal situation, it would be possible to learn the function $f(\mathbf{x})$ as $N\rightarrow \infty$ without imposing any conditions on the convergence of the derivatives of the model $s_i$.

The learning problem posed is only a necessary condition for the fulfilment of Equation~(\ref{eq:probabilidadad_cota}), and nothing may be said about the fulfilment or not of the condition in Equation~(\ref{eq:derivada_cota}). This can be seen in two extreme situations. The first arises when applying neural network models that allow for simultaneous approximation of probability functions and their derivatives of different orders  ($\varepsilon=\varepsilon_0=\varepsilon_1$) . There are theoretical results for various types of activation function that allow the necessary depth, width, and sparsity of a deep neural network to be determined, for approximating any smooth function up to a specified approximation error in certain norms \citep{BPN23, HoY22, RLM21}. Therefore, such models may be able to satisfy the two aforementioned conditions. However, the downside is that the bounds on the number of non-zero weights provided in the literature are of the order of $O(\varepsilon^{-d/\beta})$, where $d$ is the dimension of the feature vector and $\beta$ is the order of smoothness. This indicates an exponential growth in the number of parameters as a function of dimensionality, which is a type of curse of dimensionality.

With respect to ensemble approaches based on decision trees, such as \gls{rf} and \gls{xgboost}, it can be stated that they present a contrasting situation. These techniques may not be the most appropriate for estimating differentiable probabilities due to their estimation procedure, which typically involves aggregating local estimators based on rule-based approaches. The probability of each class is estimated by computing the proportion of votes for each class, which is the standard estimation procedure for ensemble approaches based on decision trees, such as \gls{rf} and \gls{xgboost}. The resulting probabilities $s_i(\mathbf{x},\boldsymbol{\omega} )$  take on only a finite set of values, i.e. $s_i(\mathbf{x},\boldsymbol{\omega} ) \in \left \{0, \frac{1}{\hbox{n-estimators}},\frac{2}{\hbox{n-estimators}},\cdots,  1 \right \}$. As a result, the function $s_i$ becomes discontinuous and therefore  non-differentiable, with partial derivatives equalling either $0$ or undefined. Therefore, it is evident that the condition in Equation~(\ref{eq:derivada_cota}) cannot be satisfied in these models.

The above discussion highlights that \gls{ml} models do not guarantee the existence of derivatives. As a result, analytical methods are replaced with numerical approximations to calculate gradient information.  Let $\mathbf{x}_{n}$ be the current value of the feature vector and consider two scenarios where all attributes take the same value, except for the attribute under study, which takes $x^+_{nik}=x_{nik}+h$ and $x^-_{nik}=x_{nik}-h$. These new vectors are denoted by $\mathbf{x}^+_{nik}$ and $\mathbf{x}^-_{nik}$, respectively. The Center Divided Difference Method approximates the $k-$th partial derivative as follows:
\begin{equation}
  \label{numerical_derivative}
  \cfrac{\partial P_{i}}{ \partial x_{ik}} (\mathbf{x}_{n})= \cfrac{P_i (\mathbf{x}^+_{nik})-P_i (\mathbf{x}^-_{nik})}{2h}+O(h^2)\approx \cfrac{s_i (\mathbf{x}^+_{nik})-s_i (\mathbf{x}^-_{nik})}{2h}.
\end{equation}

Equation~(\ref{numerical_derivative}) assumes that $s_i$ approximates a differentiable function $P_i$, and by applying the above formula,  an approximation to the derivative of $P_i$ is obtained. These approximations will be used in all the models for the numerical experiments.

To sum up, this section discusses the fact that there is no theoretical basis to guarantee the uniform approximation of the derivatives of the probability function and only  numeric experimentation can properly evaluate the performance of the different \gls{ml} models used for modelling travel mode choice. This is the purpose of the next section.

\section{Methodology and setup of experiments}\label{sect:Method}
\label{sect:setupexperiments}
\begin{figure}[h]
	\begin{center}
		\includegraphics[width=0.95\textwidth, angle=0, origin=c]{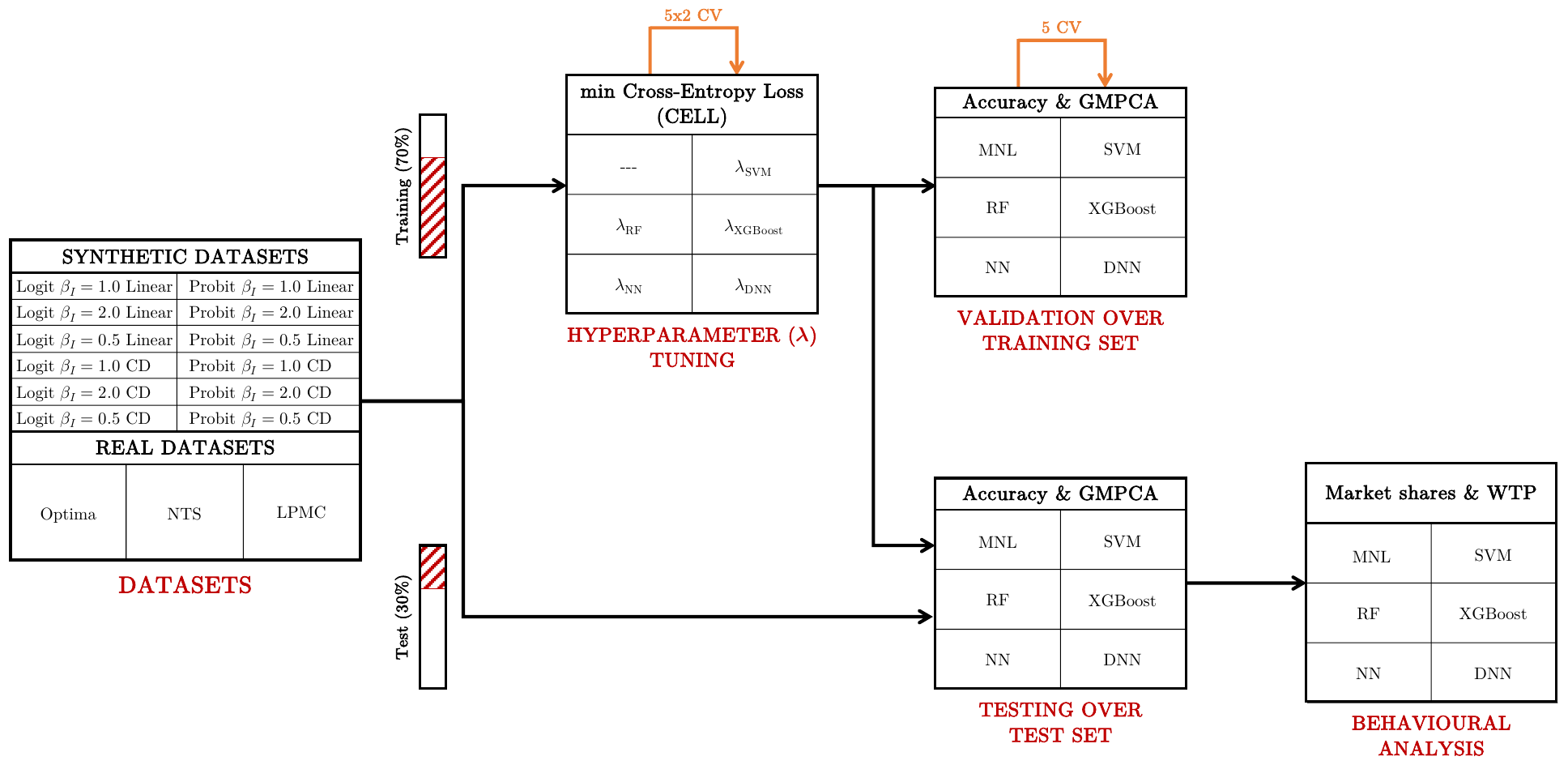}
		\caption{Graphical summary of the methodology employed in this study}
		\label{fig:methodology}
	\end{center}
\end{figure}

Figure \ref{fig:methodology} summarises the methodology used in this work. It can be observed that experiments are conducted on two types of datasets: synthetic and real. More specifically, twelve synthetic datasets were generated, and three real datasets of travel mode choice were used. The details of the dataset properties are explained in Section \ref{sect:Datasets}. These datasets are split into training and test subsets. $70\%$ of the data were used for model training, while the remaining $30\%$ were used for testing and validation. It should be noted that according to \citet{Hil21}, when using survey data, responses from the same individuals should belong entirely to either the training or the test set.

Once the training and test datasets are prepared, the different models are defined. Section \ref{sect:Techniques} describes all the algorithms and models used in this study. Before training the models, there is a hyperparameter tuning stage, carried out to obtain an optimal set of hyperparameters for each combination of model and dataset where a $5-$fold cross-validation (denoted as $5-$fold CV) was performed to minimise the cross-entropy loss function. The hyperparameter tuning stage is a time-consuming process, and to expedite it, a subset of $25\%$ of the original dataset is used. Subsequently, the most promising hyperparameter is selected, and the model is trained on the complete dataset. The details of the hyperparameter tuning can be found in Section \ref{sect:HPO}. 

Next, all the models are trained (Section \ref{sect:Train}) and assessed using different performance metrics. More specifically, the models are tested using a $5-$fold CV procedure over the training set where the accuracy and \gls{gmpca} metrics are reported. This is known as model validation.

The models are then evaluated on the unobserved data contained in the test set. This allows the generalisation capabilities of the models to be seen. 
 The accuracy and \gls{gmpca} metrics are also reported. 
Finally, a behavioural analysis is accomplished through the calculation and estimation of market shares and \gls{wtp}.

\subsection{Datasets}\label{sect:Datasets}
One limitation present in many literature studies is the use of a single real-world dataset in the comparatives. This limits the validity of the conclusions, since it is not possible to determine whether or not what is observed in such dataset is a pattern of the \gls{ml} methods or is a characteristic of the analysed dataset. For this reason $15$ datasets were used in this comparison. Firstly, three real datasets, which differ in the nature of their data and their sizes, were considered. This kind of dataset is typically used in the literature and is intended to evaluate the performance of the classifiers in their context of application. In addition to these datasets, $12$ synthetic datasets with different systematic utilities and error terms were randomly generated using a Monte Carlo simulation. The most important difference with respect to real datasets is that they permit the existence of a ground-truth with which to compare the results obtained by the different models. These two kinds of dataset are described in Sections \ref{sect:RealDatasets} and \ref{sect:SyntheticDatasets}, respectively.

\subsubsection{Real-world datasets}\label{sect:RealDatasets}
To provide a comprehensive comparison of \glspl{rum} and \gls{ml} models, three different datasets have been used: Optima \citep{Bie18}, NTS \citep{HaH17}, and LPMC \citep{HEJ18}. The variables used in each dataset, along with a concise description of each, are presented in \ref{App:Dataset_variables}.

Optima is a revealed preference dataset containing a total of $1,124$ surveys obtained in order to analyse travel behaviour between the years 2009 and 2010 in Switzerland. Later, the data was processed to extract $1,906$ total travel sequences from the collected surveys (which correspond to the number of observations in the dataset) with $115$ variables each. Therefore, more than one travel sequence can be associated with the same individual. For this study, the dataset was preprocessed to remove the observations with missing values and convert categorical variables into one-hot encoding, and finally $14$ variables were selected. As additional information, the variable of choice in this dataset takes the following values and distribution: public transport (which represents $28.12\%$ of the instances), private modes (which accounts for $65.90\%$ of the data) and soft modes (which represents $5.98\%$ of the total instances). 

NTS is a \gls{ml}-focused dataset comprising the results of the Dutch National Travel Survey (NTS) conducted between the years 2010 and 2012, in which environmental data were also included. The dataset includes $230,608$ trips obtained from a total of $69,918$ individuals. However, only a subset of $100,000$ randomly selected trips were used for the experiments for computational reasons. NTS is therefore a larger dataset than Optima. In this study, all sixteen variables of NTS have been selected. This dataset has also been preprocessed to convert categorical variables into one-hot encoding. The choice variable in this dataset takes the following values and distribution: walk (which corresponds to $16.29\%$ of the total instances), bike (which is $24.41\%$ of the data), public transport (which represents $4.03\%$ of the instances) and car (which is $55.26\%$ of the data). 

Finally, the London Passenger Mode Choice (LPMC) dataset consists of single day travel diary data obtained from the London Travel Demand Survey (LTDS) during the years 2012 to 2015. This data was augmented with additional variables collected through a directions API, including accurate cost and travel time estimates. The dataset contains $81,096$ samples, each of which corresponds to one trip made by one of the $17,616$ participants in the survey. This dataset contains a total of $31$ variables, $20$ of them were selected for this study, as the rest contained redundant or correlated information. Similarly to previous datasets, categorical variables were transformed into one-hot encoding. The choice variable takes the following values: walk (which corresponds to $17.6\%$ of the total instances), bike (which is $2.98\%$ of the data), public transport (which is $35.28\%$ of the data) and car (which represents $44.16\%$ of the instances).

To obtain the training and test sets from the Optima and NTS datasets, as shown in Figure \ref{fig:methodology}, they were randomly split into subsets of size $70\%$ and $30\%$ of the data, respectively. For the LPMC dataset, the last year of data (2014/15) was taken for testing, and the remaining years for training (2012/13-2013/14); that is, approximately $70\%$ training and $30\%$ test, and so it is consistent with the division in the other two datasets. All variables in the three datasets have been normalised to zero mean and unit standard deviation using the training set as a reference.

\cite{Hil21} demonstrated that trip-wise sampling with panel data causes a bias in model performance estimates, particularly for flexible non-linear \gls{ml} classifiers. Therefore, all the datasets were preprocessed to take account this, and the responses of the same individual are used exclusively in the training set or the test set, to fulfil the key principle of validating \gls{ml} classifiers on unseen out-of-sample data. Stated differently, the training and test sets are designed to avoid having observations from the same individual or household distributed across both sets. This same principle is applied to split the data for the cross-validation at the validation stage.

The Optima dataset contains one attribute that identifies individuals while the LPMC dataset contains attributes for identifying individuals and households. However, the NTS dataset does not contain such information. To solve this, the procedure described in \cite{Hil21} was applied, which consists in identifying the households by the unique combination of geographic area (inferred using the variables \textit{diversity} and \textit{green}) and the socio-economic information of the individuals (which combines the variables of \textit{age}, \textit{gender}, \textit{ethnicity}, \textit{education}, \textit{income}, \textit{cars}, \textit{bikes}, and \textit{driving license ownership}). Due to the improvement in the validation scheme of the models for the NTS dataset, the performance reported in this study is not comparable to that obtained, for example, in \cite{HaH17}.

To conclude, it is considered interesting to refer to some aspects related to the so-called panel effect. The three real-world datasets can have multiple responses from the same individual, Optima and LPMC data contain only one-day travel diaries, which means it is not possible to investigate habitual behaviours or changes in these over time at the level of individual travellers. On the contrary, NTS contains trips over six days, and this panel effect is taken into account in the assessment of the methods. One aspect that needs to be investigated is how to incorporate the panel effect in the estimation of \gls{ml} methods.  Re-sampling’ methods have been applied for random utility modelling with panel data in a seemingly very reliable manner \citep{DaH11}, and they could be extended to deal with panel effects  in the estimation of  \gls{ml}.

\subsubsection{Synthetic datasets}\label{sect:SyntheticDatasets}
Firstly, a Monte Carlo simulation study was designed to control the error term and the utility specification. For each synthetic dataset, two explanatory variables and three alternatives are considered, where the utility of the alternative $i$ for the individual $n$ is given by the expression
\begin{equation}
\label{eq_U_j}
  U_{ni}=V_i(x_{ni},I_{ni})+\varepsilon_{ni}; \hbox{ with } i\in \{1,2,3\},
\end{equation}
\noindent where the error terms $\varepsilon_{ni}$ are i.i.d. random variables drawn from i)~a Gumbel distribution with scale parameter $\beta=\frac{1}{\sqrt{12}}$ and location parameter $\mu=0$ or ii)~a Normal distribution with mean $\mu=0$ and standard deviation $\sigma=\frac{1}{\sqrt{12}}$. These models are respectively a logit (i) and probit (ii) model, and they constitute the ground-truth needed to compare the performance of the different approaches to be analysed.

For the simulation experiment two systematic utilities have been considered:
\begin{eqnarray}
	V_i(x_{ni},I_{ni})=& \left ( \beta_x x_{ni}+\beta_I I_{ni}   \right )    &\hbox{\quad Linear}\\
	V_i(x_{ni},I_{ni})=& \left( {x_{ni}}^{\beta_x} \right ) \left ({I_{ni}}^{\beta_I} \right )  &\hbox{\quad Cobb-Douglas (CD)}\label{eq_CD}
\end{eqnarray} 
where $x_{ni}$ y $I_{ni}$ are the quantity of a commodity $x$ consumed and the user's income $n$ when choosing alternative $i$, respectively.

Three pairs of parameters can be obtained by the combination of $\beta_x=1$ and $\beta_I\in \{0.5,1,2\}$\footnote{Note that for the estimation of the methods implemented, the \gls{mnl} assume that $\beta=\sigma=1$. This means that the estimation of the systematic utilities for logit models is $V'=\frac{V}{\beta}$, or $V'=\frac{V}{\sigma}$ in the case of the probit models. Therefore, in the case of linear utilities, the parameter estimates produced by the \gls{mnl} are the original parameters divided by $\beta$ or $\sigma$.
}. Therefore, 12 different models can be generated by combinations of: two error distributions (Gumbel or Normal) $\times$ two systematic utilities (Linear or Cobb-Douglas) $\times$ three pairs of parameters. For each model, a total of $10,000$ individuals were generated to train the methods compared in this experiment. Those individuals were drawn randomly from a uniform distribution of $(x_{ni},I_{ni})$ on the square $[0,1] \times [0,1]$ for $i=1,2,3$. This leads to a vector of features $\mathbf{x}_{n}=(x_{n1},I_{n1},x_{n2},I_{n2},x_{n3},I_{n3})$ for each individual. Finally, the same procedure was repeated to obtain a further $1,000$ samples that were used to test the performance of those methods.

\subsection{Selection of Travel Mode Choice techniques}\label{sect:Techniques}
In this section we provide a brief introduction to the selected methods for modelling travel mode choice.
  
\begin{itemize}
  \item \acrfull{mnl}. It is the baseline model used for discrete choice analysis. In this paper, we have used linear utility functions for each dataset, and each \gls{mnl} model was estimated using maximum likelihood estimation. More specifically, with respect to the synthetic datasets, linear utilities specified over all the attributes have been considered where these utilities use different parameters for each alternative. The same procedure was applied for defining the utility functions of the NTS dataset. However, for the Optima and LPMC datasets\footnote{The reader is encouraged to look at \cite{Bie18}, \cite{HaH17}, and \cite{HEJ18} to obtain the complete list of attributes of the Optima, NTS, and LPMC datasets, respectively.}, a utility function was defined with individual specific attributes and alternative varying attributes. In the case of the Optima dataset, the following features were selected for each alternative:
  \begin{itemize}
    \item Public transport: \textit{TimePT}, \textit{MarginalCostPT}, and \textit{distance\_km}.
    \item Private modes: \textit{TimeCar}, \textit{CostCarCHF}, and \textit{distance\_km}.
    \item Soft modes: \textit{distance\_km}.
  \end{itemize}
  The remaining features were selected as individual specific attributes. Similarly, for the LPMC dataset, all the features were selected as individual specific, except for the following features, which were selected per alternative:
  \begin{itemize}
    \item Walk: \textit{distance} and \textit{dur\_walking}.
    \item Bike: \textit{distance} and \textit{dur\_cycling}.
    \item Public transport: \textit{dur\_pt\_access}, \textit{dur\_pt\_rail}, \textit{dur\_pt\_bus}, \textit{dur\_pt\_int\_waiting}, \textit{dur\_pt\_int\_walking}, \textit{pt\_n\_interchanges}, and \textit{cost\_transit}.
    \item Car: \textit{dur\_driving} and \textit{cost\_driving\_total}.
  \end{itemize}
  
  \citet{KBD21} compare the unconditional out-of-sample predictive abilities of standard logit, standard mixed logit, and mixed logit with unobserved inter- and intra-individual heterogeneity. The results of both the simulation study and the real data application indicate that the two types of mixed logit do not offer significant improvements in unconditional predictive accuracy compared to standard logit. These findings are consistent with \citet{SDA22} and suggest that while mixed logit is useful for explaining behaviour, it may not necessarily provide more accurate unconditional out-of-sample predictions than standard logit.

  \item \acrfull{svm}. In its binary version, \gls{svm} considers the following labelled training data $y_{n}\in \{-1,1\}$. Then, the decision function $h(\mathbf{x})=\hbox{sgn}\left (\sum_{n=1}^N y_n \omega_n K(\mathbf{x}_n, \mathbf{x}) + \rho \right )$ is built, where $K(\mathbf{x}_n,\mathbf{x})$ is the so-called kernel function. In this paper, the kernel chosen is the radial basis function (RBF) kernel, which is widely used when working with \gls{svm}. This kernel is defined by $K(\mathbf{x}_n,\mathbf{x})=\exp \left (-\sigma \|\mathbf{x}_n-\mathbf{x}\|^2_2 \right)$. Regarding parameter estimation, quadratic programming is used to solve the training problem and obtain the $\omega_n$ parameters. In a multi-class problem where $n_C=|C|$, such as travel mode choice, $\frac{n_C(n_C-1)}{2}$ binary classifiers are trained to separate all pairs of $n_C$ classes. Thus, the predicted class is the most voted among all binary classifiers. \gls{svm} does not directly provide probability estimates, and so they have been calculated using Platt's method, which consists of training the \gls{svm} using $5-$fold CV and subsequently applying a logistic regression on the \gls{svm} scores. Furthermore, due to the large size of some datasets, it is not possible to calculate the kernel matrix directly, as it would be computationally expensive and it would not fit in memory. One approach to improving the running time of kernel-based methods is to build a small sketch of the kernel matrix and use it to generate a low-rank matrix approximation. The Nyström \cite{SZZ15} method is the most widely used and is based on the random selection of $m$ rows and $m$ columns of the kernel matrix. This method has been used for the NTS and LPMC datasets, taking the value of $m$ as a $5\%$ of the total observations. We have observed numerically that if the value of $m$ increases the improvement of the procedure is residual.
  
  \item \acrfull{rf}. Decision tree-based models are widely used classifiers whose structure makes it possible to interpret and explain the decisions made by the model in a simple manner. In a decision tree, the nodes are model binary decision rules and, for classification, the leaves represent the different classes (mode choices in this case). At each split, the variables or attributes that maximise the separation between instances of the dataset are chosen. The tree is extended recursively, starting from the best split, until a stopping criterion is satisfied. 
  \gls{rf} is a tree-based ensemble method that builds a set of decision trees in parallel, each estimated to perform the same predictive task on different bootstrap samples of the data, with further subsampling of the features. By averaging out the different sources of error across the ensemble, the classification obtained by the final model tends to be more accurate than the classification obtained by an individual decision tree. Whilst the \gls{rf} model does not output true probabilities for classification, the proportion of each class at the leaf nodes of each tree can be averaged to output probability-like values.  
  
  \item \acrfull{xgboost}. Like \gls{rf}, \gls{gbdt} is a tree-based ensemble method that can be used for both regression and classification. However, in the \gls{gbdt} model, the trees are estimated sequentially, with each tree predicting the residual error of the previous trees in the ensemble, through use of a gradient. The model was first proposed in \cite{Fri01}. \gls{xgboost} is an efficient implementation of the \gls{gbdt} model, incorporating several methodological advances. Unlike \gls{rf}, where each tree performs the same predictive task as the ensemble (i.e. classification trees are used for a classification task, regression trees for a regression task), \gls{gbdt} always makes use of regression trees to predict a continuous residual. For classification, the regression values are passed through the softmax\slash logistic function to generate well-calibrated probabilities. The model is therefore functionally similar to an \gls{mnl} model, with each linear-in-parameters utility specification replaced by an ensemble of regression trees.

  \item \acrfull{nn}. Neural models are widely used in function approximation and classification problems, due to their simplicity and their ability to capture the non-linear relationships between the input variables and the target variable. For multinomial classification, a shallow \gls{ffnn} (defined here as containing a single hidden layer) consists of an input layer, which contains a number of neurons equal to the number of attributes of the dataset. This is followed by a layer called the hidden layer, which contains a variable number of neurons, each fully connected to the neurons of the input layer with a non-linear activation function. Finally, the output layer of the model contains as many neurons as the number of classes (alternatives) in the problem, each fully connected to the neurons in the hidden layer. As with the \gls{mnl} and \gls{xgboost} model, the regression values in the output layer are passed through the softmax\slash logistic function to generate well-calibrated probabilities. The training of the model is carried out using backpropagation, which tries to minimise a loss function. 
  
  We represent the shallow \gls{nn} model formally in Equation~\ref{Eq_NN}. Let $\mathbf{x}$ be the vector of features (or attributes), hence, the probability that instance $\mathbf{x}$ belongs to class $i$, denoted $s_i(\mathbf{x})$, is computed as follows:
  \begin{equation}
  \label{Eq_NN}
   s_i(\mathbf{x}) = \sigma' \left ( \sum_{j=1}^{n_2}\omega_{ji} \cdot \sigma\left (\sum_{k=1}^{n_1}\omega_{kj}\cdot x_k + b_{0j} \right )+b_{0i} \right ),
  \end{equation}
  \noindent where the integers $n_1$ and $n_2$ correspond to the number of neurons in the hidden and output layer (such that $n_1$ is a parameter decided by the modeller and $n_2=|C|$), the parameters $\omega_{ji}$ are the weights of the connections between the neurons from the input and hidden layer,  $\omega_{kj}$ are the weights of the connections between the hidden and output layer, and $\sigma$ and $\sigma'$ are the activation functions of the hidden layer and output layer respectively. As discussed, it is standard practice to use the softmax\slash logistic function in the output layer ($\sigma'$) to produce class probabilities. For the activation function of the hidden layer ($\sigma$), we use the Rectified Linear Unit (ReLU) function, which has been demonstrated to work well for a wide range of problems. Finally, $b_{0j}$ and $b_{0i}$ model the biases or threshold values to compute the activation function in hidden and output layers. 
  
  \item \acrfull{dnn}. Sometimes, the performance of shallow \gls{nn} is not sufficient in multiple application domains. In particular, applications where non-structured data like images, video, documents, etc. are manipulated, often make use of \gls{dnn} with complex structures. In this study, we use the term \gls{dnn} to indicate a \gls{ffnn} which contains more than one hidden layer, as opposed to a \gls{nn} which indicates a shallow \gls{ffnn} with only a single hidden layer. The main advantage of this model is the increase in its predictive capability over the shallow \gls{nn} model. On the contrary, the growth of the model parameters and the overfitting are the main challenges in the implementation of this type of model.
  
\end{itemize}

\subsection{Hyperparameter optimisation}\label{sect:HPO}
During the training process of a \gls{ml} algorithm, several parameters must be optimised. Furthermore, depending of the algorithm type, \gls{ml} models have another set of pre-configured parameters that control the learning process. These are the so-called hyperparameters and they depend on the algorithm type: \gls{svm}, decision-tree based, \gls{nn}, etc. The so-called \gls{hpo} problem is defined, in order to adjust the hyperparameters. It allows the following: i)~Decreasing developer intervention in algorithm application, leading to automatic \gls{ml} systems. ii)~Tailoring hyperparameters to the problem at hand, increasing the performance of \gls{ml} algorithms. iii)~Providing reproducible studies, which ensures future fair comparisons and analysis of different \gls{ml} algorithms. 

The default hyperparameters values provided by the most widely used \gls{ml} libraries \citep{HKV19} are not often the most appropriate, which can affect the performance of the classifiers and the quality of the behavioural outputs derived from them. Many of the state-of-the-art benchmarks concerning the use of \gls{ml} classifiers applied to transport mode choice lack this procedure. Hence, it is proposed the application of a homogeneous optimisation methodology for the estimation of the hyperparameters of the \gls{ml} methods.

According to \cite{HKV19}, the \gls{hpo} problem can be formalised as follows: let $D$ be a dataset of the problem at hand. Consider a \gls{ml} algorithm with its hyperparameters denoted by $\mathcal{A}_\lambda$. Note that the vector of hyperparameters $\lambda$ belongs to the feasible region of hyperparameters, i.e. $\lambda \in \Lambda$. The optimal set of hyperparameters $\lambda^*$ is computed as
\begin{equation}
  \label{eq:HPO}
  \lambda^{*}=\underset{\lambda \in \Lambda}{\operatorname{argmin }\hspace{0.1cm}} \mathbb{E}_{\left(D_{t r a i n}, D_{v a l i d}\right) \sim \mathcal{D}} \mathbf{V}\left(\mathcal{A}_{\lambda}, D_{t r a i n}, D_{v a l i d}\right).
\end{equation}

In Equation~(\ref{eq:HPO}), $\mathbf{V}\left(\mathcal{A}_{\lambda}, D_{\text{train}}, D_{\text{valid}}\right)$ is a performance metric of $\mathcal{A}_\lambda$ on training data $D_{\text{train}}$ and assessed on validation data $D_{\text{valid}}$. The expectation of Equation~(\ref{eq:HPO}) must be approximated, in practice. A cross-validation approach is used for this purpose, where the dataset $D$ is partitioned into $K$ different subsamples (also known as folds) and denoted by $D_1,\ldots,D_K$. Next, for each $k \in \{1,\ldots,K\}$, the $\mathcal{A}_{\lambda}$ classifier is trained on $D^k_{\text{train}}=\cup_{j\ne k} D_j$ and then validated on the remaining set $D_k$. Finally, the sample mean is considered as an estimator of $\mathbb{E}_{\left(D_{\text{train}}, D_{\text{valid}}\right) \sim \mathcal{D}} \mathbf{V} \left(\mathcal{A}_{\lambda}, D_{\text{train}}, D_{\text{valid}} \right ).$ Briefly, this cross-validation process is denoted by $K-$fold CV. 

In the computational experiments carried out in this study, $5-$fold CV is used to approximate the mathematical expectation. The performance metric for $\mathbf{V}$ is the cross-entropy loss function and $1,000$ function evaluations of \gls{tpe} were run as optimisation algorithms for the problem (\ref{eq:HPO}). This way, \gls{hpo} allows to address the problem of expensive function evaluations for large models, the high-dimensional search space when there are many hyperparameters and the absence of large training datasets. The reader is referred to \ref{App:HPO} to check the implementation details of the \gls{hpo} problem and the optimal values of the hyperparameters.

\subsection{Training with optimal hyperparameters}\label{sect:Train}
The optimisation problem for training a \gls{ml} model can be stated as the {\sl penalised empirical risk}, as expressed in Equation~(\ref{eq:training_problem}). The objective function is composed of two terms, with the first term being the empirical risk and the second term being the regularisation term, which is a convex function. This formulation provides a flexible framework for various estimation methods, and the choice of functions $\ell$ and $\Omega$ determines the specific method used for training the model.

Regarding the models used in this paper, the \gls{mnl} is estimated using the Maximum Likelihood approach, which is equivalent to taking as the loss function the {\sl negative log-likelihood} and $\lambda=0$. For the binary \gls{svm} we use the {\sl hinge loss}, $\ell(h(\mathbf{x}_n; \boldsymbol{\omega}),y_n)=\max\{0,1-h(\mathbf{x}_n; \boldsymbol{\omega})y_n\}$. In the case of \gls{rf}, the loss function is taken as a hyperparameter of the model which can take two possible values: Gini or Entropy loss function. In the case of \gls{xgboost}, the Squared Log Error (SLE) loss function is employed. This is defined by $\ell(h(\mathbf{x}_n; \boldsymbol{\omega}),y_n)= \frac{1}{2}\sum_{i\in C} [log(s_{i}(\mathbf{x}_n; \boldsymbol{\omega})+1)-log \left (\mathbb{I}(y_n=i)+1 \right )]^2$. Finally, \gls{nn} and \gls{dnn} consider the log-loss function defined by $\ell(h(\mathbf{x}_n; \boldsymbol{\omega}),y_n)= \sum_{ i\in C } \mathbb{I}(y_n=i) \cdot log(s_{i}(\mathbf{x}_n; \boldsymbol{\omega}))$ in a multi-class problem with $|C|$ classes, where $\mathbb{I}(\cdot)$ is a function that takes value $1$ if the equality is satisfied, and $0$ otherwise.

The regularisation term is introduced to avoid {\sl overfitting}. Two common regularisation terms are the Lasso method, which uses the $L_1$-norm $\Omega(\boldsymbol{\omega})=\|\boldsymbol{\omega}\|_1$, or the Ridge method, which uses the Euclidean or $L_2$-norm $\Omega(\boldsymbol{\omega})=\|\boldsymbol{\omega}\|_2$. In order to validate the training process of the models, $5-$fold CV was carried out. Subsequently, in order to analyse the generalisation capability of the previously trained models, they were run on the test set.

\section{Experimental results}\label{sect:experiments}

This section describes the results of the numerical experiments carried out following the methodology described in Section \ref{sect:Method}. Four experiments are introduced, set out from Section \ref{sect:exp1} to Section \ref{sect:exp4}, where the first analyses the performance of the different models over the synthetic datasets. The second experiment analyses the quality of the probability function estimated by each model (which is of great importance for the behavioural analysis). Then, in the third, a comparison is made of the models to obtain behavioural indicators. Finally, in the last experiment, the previous analysis is repeated on the three real datasets, including a feature importance analysis.

With respect to the software and hardware resources, it must be indicated that the \gls{mnl} models were estimated using the open-source Python package Biogeme \citep{Bie03}. Regarding \gls{ml} models, Scikit-learn Python package \citep{Scikit11} was used for \gls{svm}, \gls{rf} and \gls{nn}. In the case of \gls{xgboost}, the Python package with the same name was used \citep{Chen16} and Keras Python library \citep{Chollet15,Gulli17} was selected for \gls{dnn}. Finally, regarding the hyperparameter tuning, the Python hyperopt package \citep{BYC13,Komer2014} was employed. All numerical tests were run on a Linux computer running Ubuntu 20.04 LTS with a $3.8$ GHz $12$ core AMD Ryzen 3900xt processor and $32$ GB of RAM. To reproduce the experiments in this paper, please refer to \url{https://github.com/JoseAngelMartinB/prediction-behavioural-analysis-ml-travel-mode-choice}, where the code is available under a MIT license.

\subsection{Experiment 1: Performance assessment}\label{sect:exp1}
In this experiment, the comparisons already made in the literature are repeated, but on synthetic data. The most commonly-used index in the literature is the \gls{acc}, which gives us the percentage of correctly predicted observations obtained by a model. This index is defined as
\begin{equation}
  \gls{acc}= \dfrac{1}{N_{\hbox{\tiny test}}} \sum_{n}^{N_{\hbox{\tiny test}}} \mathbb{I} (y_n=\widehat y_n),
\end{equation}
\noindent where $\widehat y_n$ is the label predicted by the algorithm, i.e $\widehat y_n=h(\mathbf{x}_n)=\underset{i\in C} {\hbox{arg max }\,} s_i(\mathbf{x}_n)$ and $N_{\hbox{\tiny test}}$, the number of observations in the test set. In this study, the \gls{acc} is expressed as a percentage.

The use of synthetic data has two essential advantages. The first is that it eliminates the correlations that exist in many studies between the data used to train the models and the test data. The second is that it allows the maximum accuracy that can be achieved by any \gls{ml} method to be known. The error term is controlled in the generation of the synthetic data, and it is therefore possible to determine when the choice of an individual corresponds to the systematic part and when it is originated by the random term. The most that can be expected from a model is that it recovers all the information associated with the systematic term. The percentage of individuals whose choice is determined by the systematic part has been computed for all synthetic datasets on the corresponding training and test data.

In \cite{Hil19}, it is argued that the performance measures have to collect probability-based indices because this information is essential in the behavioural analysis of the models. These authors recommend the \gls{gmpca}, hence we have added this indicator to the analysis and it is given by the expression:
\begin{equation}
  \gls{gmpca}= \left ( \prod_{n}^{N_{\hbox{\tiny test}}} P_{n, i_n } \right ) ^{\frac{1}{N_{\hbox{\tiny test}}}}.
\end{equation}
In this work, this index is expressed as a percentage.

Experiment 1 is designed for several purposes. The first is to find out whether, when the assumptions on which the \glspl{rum} are built are met, \glspl{rum} are able to estimate the model correctly. For this reason, simulated data from three logit models with linear utilities are included. The second purpose is to find out if the models are able to perform adequately, even if any of the assumptions fail, such as through poor specificity of utility or those related to the distribution of the error term. Then, we have also included simulated data using non-linear utilities and data with normal error terms, but maintaining the \gls{mnl} with linear utilities as our benchmark model. The third purpose is to analyse the performance of the \gls{ml} methods in these situations and to test whether or not the large differences in performance compared to the \gls{mnl} models that are reported on the literature are reproduced.

In the experiments, the above performance measures were estimated on both training and test data and they are shown in Tables \ref{tab:experiment-1-train} and \ref{tab:experiment-1-test}, respectively, and the method giving the best results is marked in bold.

To sum up, the first thing to observe is that, in the training set, the \gls{mnl} models achieve the highest values for the \gls{gmpca} metric in all cases with linear utilities, as expected. Using the previously trained models on the test data, it can be seen that some of the \gls{ml} methods are better than the \gls{mnl} models. This highlights the fact that even if the assumptions of the \gls{mnl} models are satisfied, the \gls{ml} methods can still perform as well as those models. The second observation is that the \gls{ml} and \gls{mnl} methods reach the maximum accuracy, i.e., they can fully estimate the systematic part of the decision process. This also suggests that a proper optimisation of the hyperparameters can make all the \gls{ml} techniques, under the performance point of view, equivalent to the \gls{mnl} methods. \gls{rf} is the \gls{ml} method that has slightly lower performance. It is observed that \gls{mnl} is robust and performs reasonably well on the non-linear utilities. In fact, in the test dataset it achieves the maximum accuracy in two types of non-linear utilities. However, when obtaining the \gls{gmpca} indicator, its performance is lower than other \gls{ml} methods, such as \gls{svm}, \gls{nn} and \gls{dnn}.

\begin{table}
\centering
\caption{Accuracy and GMPCA results by model for each synthetic dataset on the training set (using 5-CV)}
\label{tab:experiment-1-train}
\resizebox{\textwidth}{!}{
\begin{tabular}{rlllllllllllll}
\toprule
{} &  Maximum & \multicolumn{2}{c}{MNL} & \multicolumn{2}{c}{SVM} & \multicolumn{2}{c}{RF} & \multicolumn{2}{c}{XGBoost} & \multicolumn{2}{c}{NN} & \multicolumn{2}{c}{DNN} \\
\cmidrule(lr){3-4} \cmidrule(lr){5-6} \cmidrule(lr){7-8} \cmidrule(lr){9-10} \cmidrule(lr){11-12} \cmidrule(lr){13-14} 
{} & Accuracy & Accuracy &  GMPCA & Accuracy &  GMPCA & Accuracy &  GMPCA & Accuracy &  GMPCA & Accuracy &  GMPCA & Accuracy &  GMPCA \\
\midrule
logit linear $\beta_I=1.0$  &    66.84 &    66.85 &  \textbf{47.68} &    \textbf{66.87} &  47.55 &    65.63 &  46.37 &    66.72 &  47.41 &    67.09 &  47.66 &    \textbf{66.87} &  47.52 \\
logit linear $\beta_I=2.0$  &    76.97 &    77.16 &  \textbf{58.10} &    \textbf{77.17} &  57.91 &    75.90 &  56.08 &    76.73 &  57.74 &    77.01 &  58.02 &    76.93 &  57.75 \\
logit linear $\beta_I=0.5$  &    63.33 &    \textbf{63.26} &  \textbf{43.95} &    63.01 &  43.82 &    61.98 &  42.99 &    62.84 &  43.70 &    63.07 &  43.90 &    62.89 &  43.70 \\
logit CD $\beta_I=1.0$      &    56.29 &    55.70 &  38.81 &    \textbf{56.34} &  \textbf{39.44} &    55.05 &  38.68 &    55.92 &  39.26 &    55.78 &  39.38 &    55.81 &  39.31 \\
logit CD $\beta_I=2.0$      &    53.32 &    53.20 &  37.31 &    53.02 &  38.05 &    52.58 &  37.77 &    \textbf{53.24} &  38.11 &    53.09 &  \textbf{38.19} &    52.63 &  37.96 \\
logit CD $\beta_I=0.5$      &    57.01 &    56.00 &  39.21 &    \textbf{56.94} &  \textbf{39.70} &    56.13 &  39.11 &    56.41 &  39.54 &    56.55 &  39.63 &    56.35 &  39.55 \\
probit linear $\beta_I=1.0$ &    72.12 &    72.00 &  \textbf{53.28} &    \textbf{72.03} &  53.15 &    70.67 &  51.31 &    71.67 &  52.97 &    71.94 &  53.26 &    71.83 &  53.10 \\
probit linear $\beta_I=2.0$ &    81.28 &    \textbf{81.32} &  \textbf{64.41} &    81.03 &  64.32 &    79.45 &  61.44 &    80.85 &  64.13 &    81.07 &  64.37 &    81.00 &  64.15 \\
probit linear $\beta_I=0.5$ &    67.42 &    \textbf{67.48} &  \textbf{48.33} &    67.25 &  48.24 &    66.19 &  47.17 &    67.24 &  48.06 &    67.26 &  48.30 &    67.18 &  48.16 \\
probit CD $\beta_I=1.0$     &    60.30 &    59.61 &  41.00 &    59.89 &  42.02 &    58.61 &  41.43 &    59.81 &  41.95 &    \textbf{60.06} &  \textbf{42.07} &    59.56 &  41.84 \\
probit CD $\beta_I=2.0$     &    57.53 &    57.15 &  39.12 &    \textbf{57.21} &  40.33 &    56.63 &  39.90 &    56.91 &  40.26 &    56.94 &  \textbf{40.43} &    56.89 &  40.27 \\
probit CD $\beta_I=0.5$     &    62.03 &    60.91 &  42.34 &    61.46 &  \textbf{42.92} &    60.81 &  42.28 &    \textbf{61.96} &  42.78 &    61.35 &  42.86 &    61.41 &  42.77 \\
\bottomrule
\end{tabular}
}
\end{table}

\begin{table}
\centering
\caption{Accuracy and GMPCA results by model for each synthetic dataset on the test set}
\label{tab:experiment-1-test}
\resizebox{\textwidth}{!}{
\begin{tabular}{rlllllllllllll}
\toprule
{} &  Maximum & \multicolumn{2}{c}{MNL} & \multicolumn{2}{c}{SVM} & \multicolumn{2}{c}{RF} & \multicolumn{2}{c}{XGBoost} & \multicolumn{2}{c}{NN} & \multicolumn{2}{c}{DNN} \\
\cmidrule(lr){3-4} \cmidrule(lr){5-6} \cmidrule(lr){7-8} \cmidrule(lr){9-10} \cmidrule(lr){11-12} \cmidrule(lr){13-14} 
{} & Accuracy & Accuracy &  GMPCA & Accuracy &  GMPCA & Accuracy &  GMPCA & Accuracy &  GMPCA & Accuracy &  GMPCA & Accuracy &  GMPCA \\
\midrule
logit linear $\beta_I=1.0$  &    66.79 &     66.3 &  \textbf{47.45} &     66.8 &  47.24 &     64.2 &  46.13 &     65.8 &  46.99 &     \textbf{66.9} &  \textbf{47.45} &     67.1 &  47.36 \\
logit linear $\beta_I=2.0$  &    77.47 &     79.4 &  60.46 &     78.9 &  60.27 &     77.5 &  57.59 &     79.7 &  59.81 &     79.6 &  \textbf{60.49} &     \textbf{79.9} &  60.48 \\
logit linear $\beta_I=0.5$  &    62.07 &     62.1 &  43.55 &     \textbf{62.5} &  \textbf{43.63} &     60.5 &  42.64 &     61.6 &  43.40 &     \textbf{62.5} &  43.62 &     61.7 &  43.35 \\
logit CD $\beta_I=1.0$      &    55.70 &     56.1 &  39.16 &     57.4 &  39.78 &     55.4 &  38.83 &     57.6 &  39.71 &     57.2 &  \textbf{39.80} &     \textbf{57.9} &  \textbf{39.80} \\
logit CD $\beta_I=2.0$      &    52.93 &     \textbf{53.9} &  37.67 &     53.3 &  38.16 &     52.3 &  37.95 &     53.1 &  \textbf{38.52} &     53.7 &  38.51 &     52.3 &  37.97 \\
logit CD $\beta_I=0.5$      &    57.48 &     \textbf{57.3} &  39.28 &     56.8 &  39.86 &     57.0 &  39.26 &     56.0 &  39.67 &     56.7 &  \textbf{39.87} &     55.9 &  39.70 \\
probit linear $\beta_I=1.0$ &    72.28 &     \textbf{73.9} &  \textbf{53.53} &     73.5 &  53.28 &     71.3 &  51.74 &     73.1 &  53.18 &     73.5 &  53.45 &     72.4 &  52.97 \\
probit linear $\beta_I=2.0$ &    81.24 &     \textbf{80.5} &  \textbf{65.03} &     80.0 &  64.85 &     78.6 &  62.16 &     80.3 &  64.44 &     80.1 &  64.99 &     79.3 &  64.18 \\
probit linear $\beta_I=0.5$ &    67.70 &     \textbf{67.4} &  \textbf{48.08} &     \textbf{67.4} &  47.99 &     65.9 &  47.27 &     65.9 &  47.62 &     67.2 &  48.00 &     66.1 &  47.91 \\
probit CD $\beta_I=1.0$     &    61.09 &     59.1 &  40.49 &     60.1 &  41.62 &     59.4 &  40.98 &     59.7 &  \textbf{41.56} &     \textbf{60.3} &  41.52 &     59.8 &  41.12 \\
probit CD $\beta_I=2.0$     &    57.45 &     58.5 &  39.82 &     57.6 &  40.73 &     58.0 &  40.11 &     57.5 &  40.59 &     \textbf{58.8} &  \textbf{40.80} &     58.0 &  40.73 \\
probit CD $\beta_I=0.5$     &    62.45 &     62.4 &  42.85 &     63.0 &  \textbf{43.58} &     62.0 &  42.90 &     62.4 &  43.39 &     62.7 &  43.53 &     \textbf{63.2} &  43.36 \\
\bottomrule
\end{tabular}
}
\end{table}

\subsection{Experiment 2: Analysis of the probability function}
This experiment is designed to evaluate the quality of the estimation of the probability function $\mathbb{P}(i| \mathbf{x}_n )$ obtained by the \gls{ml} models. In other applications \citep{NiC05} it has been found that, even though they show very good performance in classification tasks, the \gls{ml} models present calibration problems with the probability estimates. A reliable probability estimate is essential to the derivation of model outputs.

\begin{figure}[!h]
	\begin{center}
		\includegraphics[width=\textwidth]{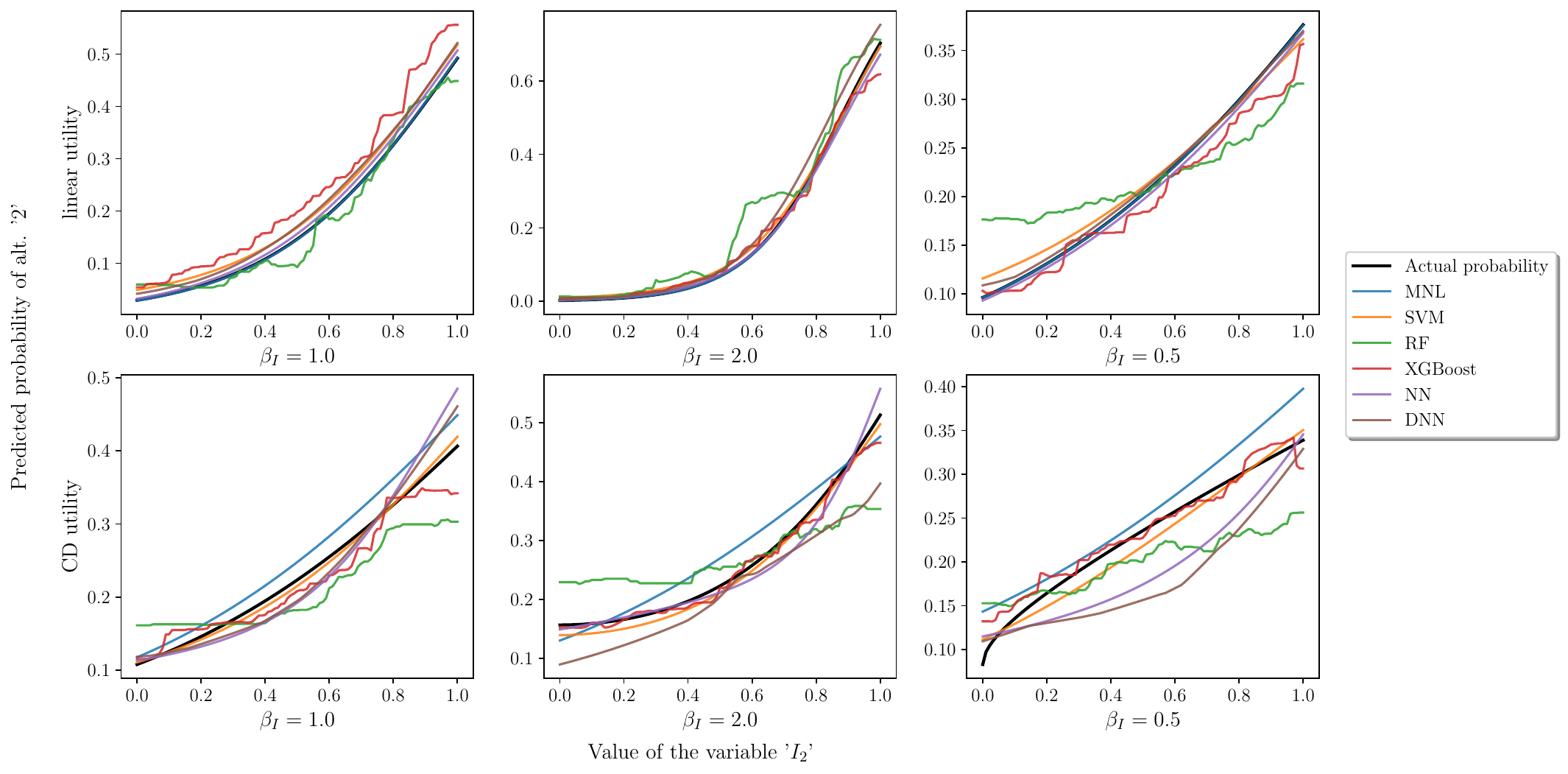}
		\caption{Probability estimates for logit models}
		\label{fig:expe2_logit}
	\end{center}
\end{figure}

\begin{figure}[!h]
	\begin{center}
		\includegraphics[width=\textwidth]{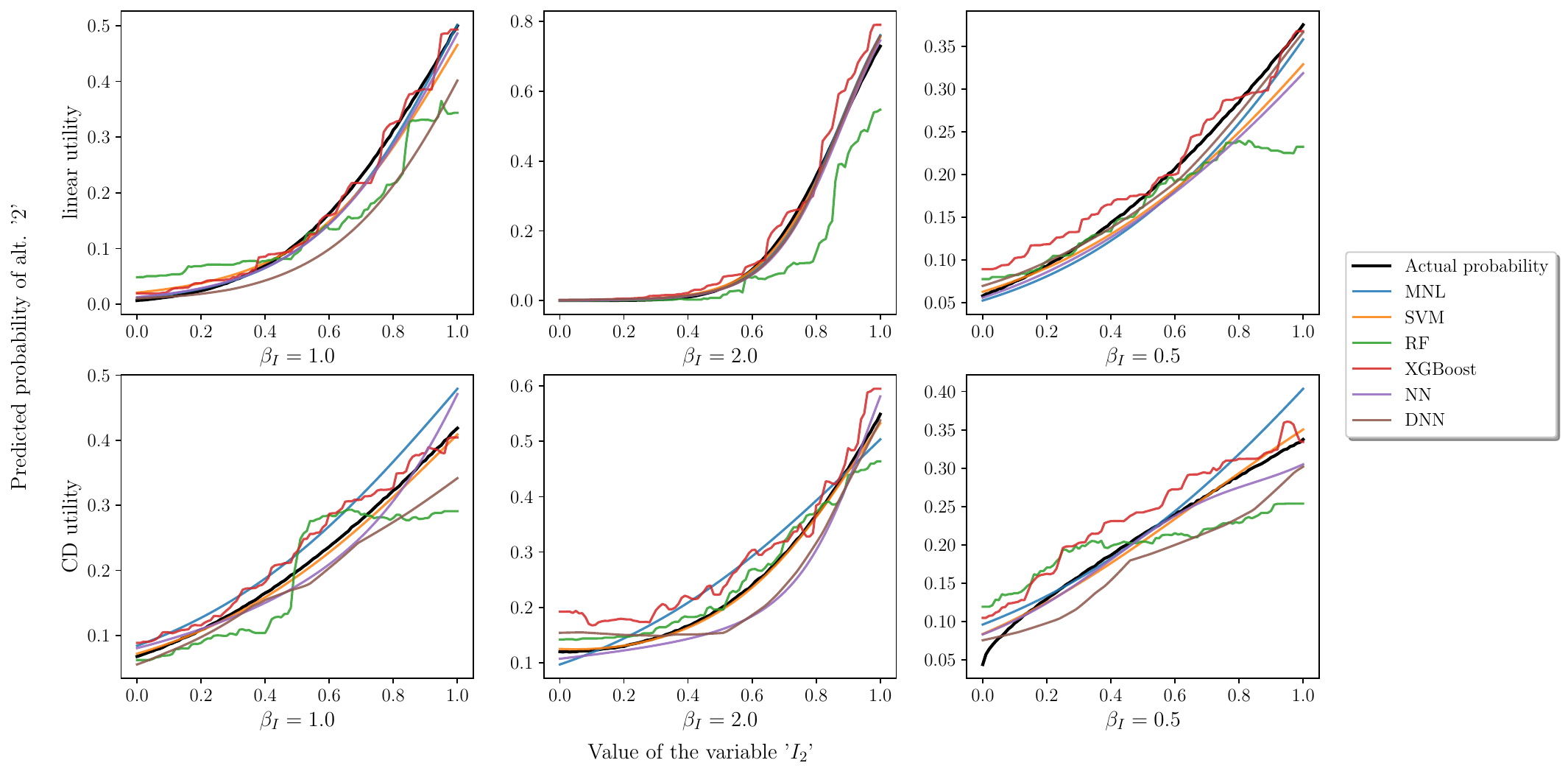}
		\caption{Probability estimates for probit models}
		\label{fig:expe2_probit}
	\end{center}
\end{figure}

For this analysis, an individual $\mathbf{x}_n=(0.25,0.25, 0.5,I_{n,2}, 0.75,0.75 )$ is considered, with $I_{n,2}\in[0,1]$ who has five of the six attributes fixed. Now, the income this individual would receive in the alternative $2$ in his domain is varied, and the change in the probability function $\mathbb{P}(i=2| \mathbf{x}_n)$ of individual $n$ selecting alternative $i=2$ is determined. Figures \ref{fig:expe2_logit} and \ref{fig:expe2_probit} show how the probability function of that individual changes depending on the value of $I_{n,2}$ for logit and probit models, respectively. The actual probability for each of the synthetic datasets can be obtained using logit and probit models, which is shown in black. Finally, the estimates obtained with the \gls{mnl} and \gls{ml} models are plotted in different colours. 

In general, all \gls{ml} methods, which do not assume probabilistic assumptions, are able to approximate the actual probability of the logit and probit models, both for linear and CD utilities. As previously noted in Section \ref{sect:RUMvsML}, ensemble methods that rely on decision trees, such as  \gls{rf} and \gls{xgboost}, may not be the most appropriate techniques for estimating monotonic and differentiable probabilities, due to their estimation procedure. Another noteworthy observation is that \gls{nn} demonstrate superior performance compared to  \gls{dnn}. Perhaps the explanation, based on the sources of error analysed in Section \ref{sect:RUMvsML}, is that for simple phenomena, a well-estimated, simple neural network is preferable to a more complex but poorly-estimated one.

The \gls{mnl} models return good estimates of the probability function for problems with linear utilities, but in the case of CD utilities a misspecification error is observed. This is an important argument supporting the use of the \gls{ml} models in these situations. Some computational experiments, not shown in this article, were also carried out, where CD utilities have been correctly specified (assuming a situation where the modeller knows the function describing the underlying phenomenon) and in this case the \gls{mnl} models accurately estimate all the probabilities.

\begin{figure}[!h]
	\begin{center}
		\includegraphics[width=\textwidth]{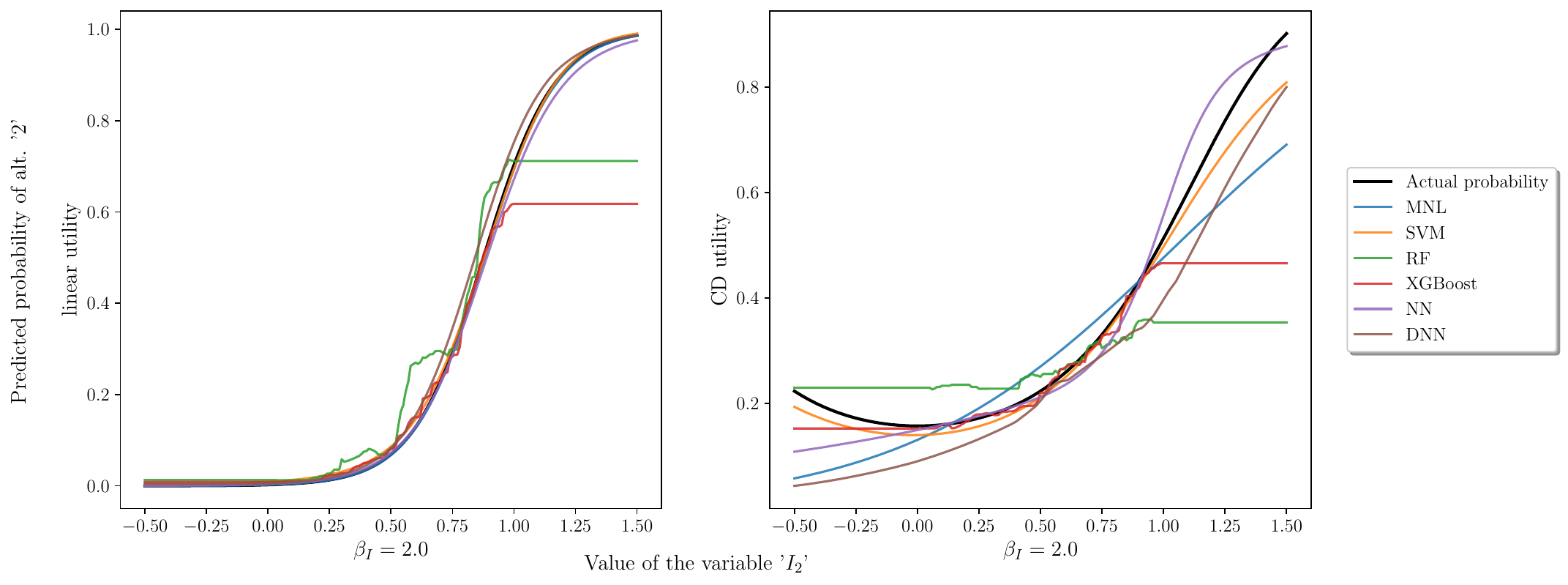}
		\caption{Model extrapolation}
		\label{fig:expe3_extrapolation}
	\end{center}
\end{figure}

Besides the previous analysis, it is interesting to examine what happens to the probability functions when we estimate values outside the training set domain. To do this, Figure \ref{fig:expe3_extrapolation} shows the probability estimates for the values $I_{n,2} \in [-0.5, 1.5]$. For simplicity, only the plots of the linear and CD utilities for $\beta_I = 2.0$ are given, as they are the most representative examples. It is observed that for the linear utilities, all models extrapolate the probabilities properly, except for the tree-based models (\gls{rf} and \gls{xgboost}). In the case of the CD utilities, although the neural network based models perform better than the tree-based models, the result is not excellent. However, a test was performed by varying the random values of the parameters in the initialisation of the \gls{nn}, and it can be improved in some cases. The values of \gls{mnl} are not satisfactory, since it does not manage to extrapolate the non-linear behaviour. The model that best fits the real probability function in this case is \gls{svm}.

It can be concluded from this experiment that the \gls{ml} methods offering the most accurate probability estimation are the \gls{svm}, \gls{nn} and \gls{xgboost} methods. If we consider the quality of the estimates, the differentiability and the computational cost of training and prediction of the model, the \gls{svm} and \gls{nn} methods are recommended. It is also observed that the \gls{mnl} models are robust when the type of error and/or in the specification of the utility function is not determined properly. 

\subsection{Experiment 3: Behavioural analysis}
\label{sect:Exp3}
Models are built to answer questions that the modeller has about reality. In the case of mode choice models, these are questions of the type {\sl what if}. For example, what would happen if we changed the price or travel time of an alternative?. This requires evaluating hypothetical situations different from the current one. This experiment recreates this situation by analysing the capacity of the different models to estimate two very useful indicators in the field of transport, namely market shares and \gls{wtp}.

\subsubsection{Market shares}
The first part of this experiment will calculate the so-called market shares in three different scenarios. The market shares for alternative $i$ are defined by:
\begin{equation}
  MS_i=\frac{1}{N_{\hbox{\tiny test}}} \sum_{n=1}^{N_{\hbox{\tiny test}}}s_i(\mathbf{x}_{n}) * 100\%,
\end{equation}
where $N_{\hbox{\tiny test}}$ is the number of test samples and $\mathbf{x}_n$ its attributes (or feature vector).

In this experiment, three different scenarios are considered:
\begin{enumerate}
  \item[\bf S1:] This is the baseline scenario, where market shares are forecasted for the current situation, which produces the synthetic data. The actual market shares for the twelve synthetic datasets are $MS^*=\left (\frac{100}{3}, \frac{100}{3}, \frac{100}{3} \right )$.
  \item[\bf S2:] This scenario is designed to promote alternative $1$. Hence, the attribute vector of alternative $1$ is modified as follows $\mathbf{x}'_n=\mathbf{x}_n+(0.3,0.3,0,0,0,0)$. This results in an average of $51\%$ of the data falling outside the domain where the models were trained on, i.e. $[0,1]^6$. 
  \item[\bf S3:] This scenario is a variation of {\bf S2}, where instead of increasing the features of all the individuals equally, the features of the alternative $1$ are increased proportionally $\mathbf{x}'_n=(1.3x_{n1},1.3I_{n1},x_{n2},I_{n2},x_{n3},I_{n3})$. On average, this leads to $40.86\%$ of the data falling outside the domain where the models have been trained, i.e. $[0,1]^6$. 
\end{enumerate}

The actual market shares, denoted by $MS^*$, have been estimated using Monte Carlo simulation with $n=50,000,000$ iterations. Table \ref{tab:estimated_share_makets} shows the values of $MS^*$ obtained for the scenarios {\bf S2} and {\bf S3} for the twelve synthetic datasets. It is observed that scenario {\bf S2} is more different from the current scenario {\bf S1} than scenario {\bf S3}. Scenarios {\bf S2} and {\bf S3} have been introduced to test the extrapolation capabilities of the methods.

To simplify the analysis of the results of this experiment, the average error of the estimation of the market shares is reported for each alternative, i.e. $\frac{1}{3}\|MS-MS^*\|_1$, where $\|\cdot \|_1$ is the $L_1$-norm. This index is expressed as the difference in percentage. The results are shown in Table \ref{tab:experiment_3_MS_table}. It can be seen that all methods, except for \gls{dnn}, reproduce the market shares of the current scenario {\bf S1} with an average error of less than $0.5\%$. Analysing the extrapolation capacity of the methods (estimates outside the domain of the training data) it is observed that the \gls{rf} and \gls{xgboost} methods are not capable of correctly estimating scenarios {\bf S2} and {\bf S3}.

The \gls{mnl} method obtains good estimates, except in the most adverse situation: non-linear utilities, especially in scenario {\bf S2}. The \gls{nn} method obtains good results in all scenarios and test problems. The \gls{svm} method returns acceptable results in all situations, although it is not the most accurate. The \gls{dnn} method gives mixed results, as it combines very good results with more modest ones. This may be due to the difficulty of training the \gls{dnn} methods properly, whether due to an optimisation error or an insufficient number of examples that increase the generalisation error.

\begin{table}
\centering
\caption{True market shares $(MS^*)$ \label{tab:estimated_share_makets}}
\label{tab:share markets}
\resizebox{0.7\textwidth}{!}{
\begin{tabular}{lllllllll}
\toprule
& & & \multicolumn{3}{c}{\bf S2} & \multicolumn{3}{c}{ \bf S3} \\
\cmidrule(lr){4-6} \cmidrule(lr){7-9} 
{\bf Model}& {\bf Utility}& $\beta_I$ & $i=1$ &  $i=2$ &  $i=3$  &  $i=1$ & $i=2$ & $i=3$ \\
\midrule
&     & 1.0	 &   66.782  & 16.609 & 16.609 &  50.045 & 24.978 & 24.978 \\ 
&Linear	 & 2.0	 & 68.695  & 15.653 & 15.653 &  50.991 & 24.504 & 24.504 \\ 
Logit &	 & 0.5	 & 62.311  & 18.845 & 18.845 &  47.821 & 26.089 & 26.089 \\ 
\cmidrule(lr){3-9}
&	 & 1.0	 & 60.962  & 19.519 & 19.519 &  45.023 & 27.489 & 27.489 \\ 
&CD	 & 2.0	 & 60.142  & 19.929 & 19.929 &  45.586 & 27.207 & 27.207 \\ 
&	 & 0.5	 & 60.107  & 19.947 & 19.947 &  44.769 & 27.615 & 27.615 \\ 
\midrule
&	 & 1.0	 & 68.486  & 15.757 & 15.757 &  50.885 & 24.557 & 24.557 \\ 
&Linear	 & 2.0	 & 69.768  & 15.116 & 15.116 &  51.524 & 24.238 & 24.238 \\ 
Probit&	 & 0.5	 & 64.013  & 17.993 & 17.993 &  48.773 & 25.613 & 25.613 \\ 
\cmidrule(lr){3-9}
&	 & 1.0	 & 62.567  & 18.717 & 18.717 &  45.749 & 27.126 & 27.126 \\ 
&CD	 & 2.0	 & 61.503  & 19.249 & 19.249 &  46.207 & 26.897 & 26.897 \\ 
&	 & 0.5	 & 61.745  & 19.128 & 19.128 &  45.452 & 27.274 & 27.274 \\ 
\bottomrule
\end{tabular}
}
\end{table}

\begin{table}
\centering
\caption{Average error of the estimated market shares by model on the synthetic datasets}
\label{tab:experiment_3_MS_table}
\resizebox{\textwidth}{!}{
\begin{tabular}{rllllllllllllllllll}
\toprule
{} & \multicolumn{3}{c}{MNL} & \multicolumn{3}{c}{SVM} & \multicolumn{3}{c}{RF} & \multicolumn{3}{c}{XGBoost} & \multicolumn{3}{c}{NN} & \multicolumn{3}{c}{DNN} \\
\cmidrule(lr){2-4} \cmidrule(lr){5-7} \cmidrule(lr){8-10} \cmidrule(lr){11-13} \cmidrule(lr){14-16} \cmidrule(lr){17-19} 
{} &    S1 &    S2 &    S3 &    S1 &    S2 &    S3 &    S1 &    S2 &    S3 &      S1 &    S2 &    S3 &    S1 &    S2 &    S3 &    S1 &    S2 &    S3 \\
\midrule
logit linear $\beta_b=1.0$  &  \textbf{0.24} &  0.73 &  0.36 &  \textbf{0.24} &  1.30 &  0.67 &  \textbf{0.24} &  6.04 &  4.14 &    \textbf{0.24} &  4.67 &  3.26 &  0.27 &  \textbf{0.72} &  \textbf{0.34} &  0.81 &  0.64 &  0.52 \\
logit linear $\beta_b=2.0$  &  0.41 &  \textbf{0.93} &  \textbf{0.82} &  0.43 &  1.44 &  1.07 &  0.44 &  6.95 &  4.88 &    \textbf{0.40} &  5.04 &  3.87 &  0.41 &  1.15 &  0.95 &  1.28 &  1.24 &  1.32 \\
logit linear $\beta_b=0.5$  &  0.46 &  1.39 &  0.89 &  0.47 &  2.39 &  1.44 &  0.49 &  7.59 &  4.74 &    0.45 &  4.93 &  3.38 &  \textbf{0.42} &  1.62 &  1.00 &  0.43 &  \textbf{1.15} &  \textbf{0.75} \\
logit CD $\beta_b=1.0$      &  0.31 &  3.26 &  \textbf{0.34} &  0.33 &  1.92 &  0.56 &  0.32 &  7.79 &  3.46 &    0.30 &  4.43 &  1.74 &  0.29 &  \textbf{0.79} &  1.43 &  \textbf{0.25} &  1.81 &  0.39 \\
logit CD $\beta_b=2.0$      &  0.50 &  4.66 &  1.48 &  0.49 &  2.58 &  0.71 &  \textbf{0.39} &  7.98 &  3.75 &    0.51 &  3.74 &  1.41 &  0.49 &  1.31 &  \textbf{0.57} &  2.00 &  \textbf{1.20} &  1.85 \\
logit CD $\beta_b=0.5$      &  0.40 &  \bf 3.04 & \bf  0.37 &  0.40 &  3.26 &  1.61 &  0.40 &  7.46 &  4.00 &    0.41 &  4.69 &  2.62 & \bf 0.39 &   3.18 &  1.48 &  1.49 &  3.21 &  1.80 \\
probit linear $\beta_b=1.0$ &  \bf 0.10 &  \bf  0.30 &  \bf 0.22 &  \bf  0.10 &  0.66 &  0.45 &  \bf 0.10 &  6.64 &  4.68 &    0.11 &  3.58 &  3.05 &  0.15 &  0.61 &  0.28 &  2.18 &  1.58 &  1.88 \\
probit linear $\beta_b=2.0$ &  0.36 &  0.39 &  0.28 &  0.34 &  0.89 &  0.60 &  0.32 &  5.85 &  4.33 &    0.37 &  3.47 &  2.97 &  0.40 &  0.87 &  0.52 &  \bf 0.31 &  \bf 0.24 &  \bf 0.13 \\
probit linear $\beta_b=0.5$ &  0.10 &  \bf 0.18 & \bf 0.12 & \bf  0.07 &  0.53 &  0.40 &  0.09 &  5.48 &  3.85 &    0.09 &  4.14 &  3.17 &  0.10 & \bf  0.18 &  0.20 &  2.83 &  3.04 &  2.70 \\
probit CD $\beta_b=1.0$     &  0.45 &  1.75 &  0.85 &  \bf 0.43 &  1.18 &  0.35 &  0.49 &  4.81 &  2.51 &    0.45 &  3.43 &  1.70 &  0.46 &  \bf 0.57 & \bf  0.27 &  0.58 &  1.04 &  0.60 \\
probit CD $\beta_b=2.0$     &  0.27 &  3.75 &  1.12 &  0.28 &  1.69 &  \bf 0.70 &  \bf 0.24 &  6.90 &  3.74 &    0.25 &  3.83 &  2.23 &  0.28 &  \bf 0.64 &  0.89 &  1.74 &  1.47 &  1.78 \\
probit CD $\beta_b=0.5$     &  0.16 &  1.07 &  0.87 &  0.17 &  1.18 &  0.24 &  0.18 &  5.38 &  3.12 &    0.15 &  4.39 &  2.42 & \bf  0.09 &  \bf 0.40 & \bf  0.11 &  0.75 &  1.88 &  1.05 \\
\bottomrule
\end{tabular}
}
\end{table}

\subsubsection{Willingness to pay}
The \gls{wtp} index quantifies how much a user would be willing to pay to consume an additional unit of product $x$. The mathematical expression of this index for the utility functions defined on this study is:
\begin{eqnarray}
    \label{WTP-CD}
  \gls{wtp}_{ni}=\frac{\beta_x I_{ni}}{\beta_I x_{ni}},&\hbox{\quad Cobb-Douglas}\\
    \label{WTP-Linear}
  \gls{wtp}_{ni}=\frac{\beta_x}{\beta_I },& \hbox{\quad Linear}
\end{eqnarray}
\noindent where the variable $I_{ni}$ is considered to play the role of the cost of alternative $i$ for user $n$.

It is observed in the expression for linear utilities, Equation~(\ref{WTP-Linear}), that the \gls{wtp} is the same for all decision makers $n$, while for non-linear Cobb-Douglas utilities, Equation~(\ref{WTP-CD}) shows that the \gls{wtp} differs between decision makers. As the value of \gls{wtp} is different for each individual, two metrics will be analysed in this experiment. The first is the empirical distribution of the sample of estimates $\left \{\gls{wtp}_{ni} \right \}_{n=1}^{N_{train}}$ and the second is a global estimate of the \gls{wtp} for the whole population. All the methods will be compared in both metrics. 

\begin{figure}[!h]
	\begin{center}
		\includegraphics[width=0.90\textwidth, angle=0, origin=c]{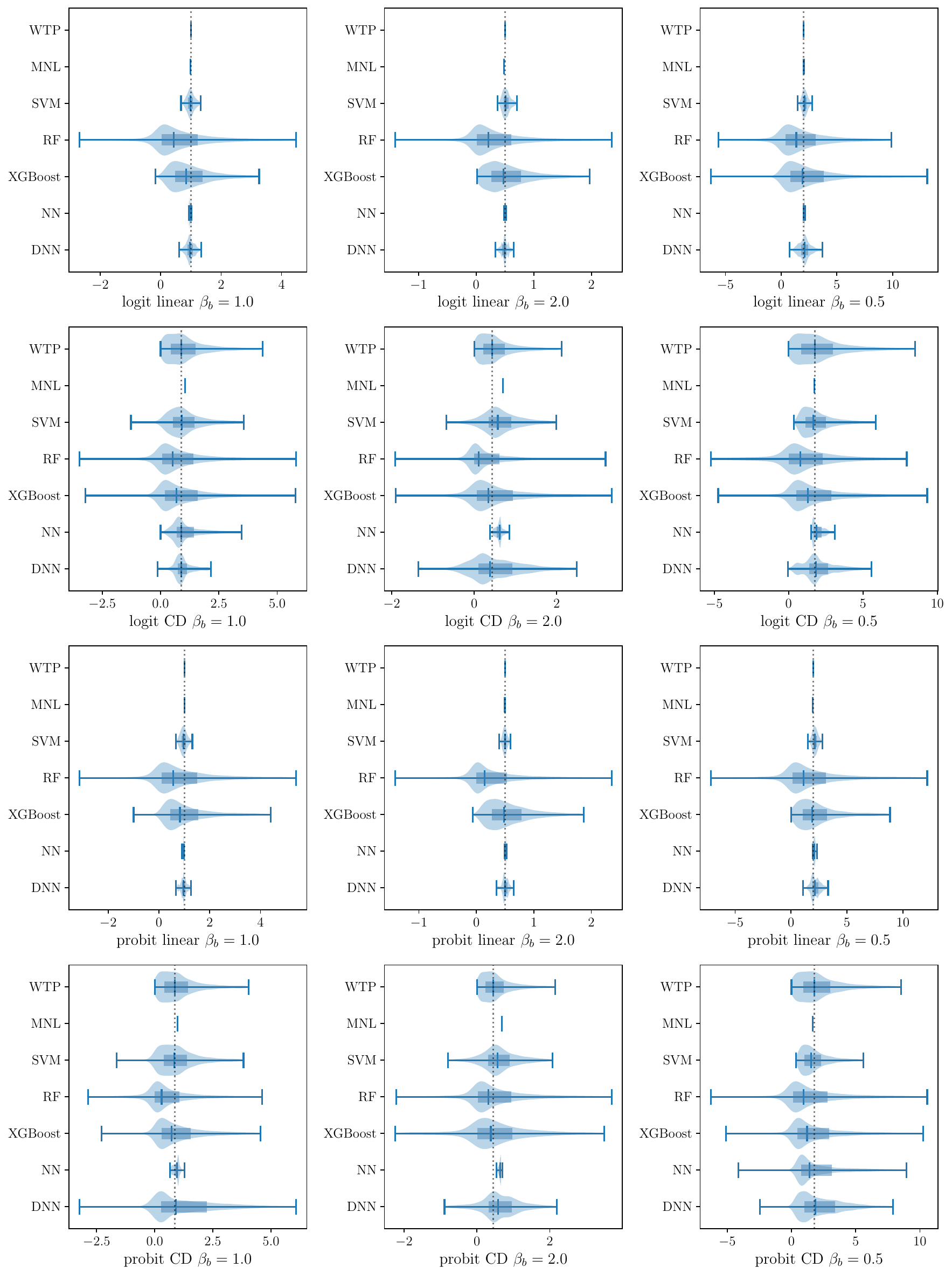}
		\caption{Comparison of the WTP of the different models}
		\label{fig:Experiment-3-WTP}
	\end{center}
\end{figure}

Equation~(\ref{eq:wtp-trick}) was used for the calculation of $\gls{wtp}_{ni}$, where the derivatives have been numerically approximated using the expression in Equation~(\ref{numerical_derivative}) and taking the value $h=0.025$. This calculation was performed considering $i=1$ for all the \gls{ml} methods, and for each decision maker $n$. Since a numerical approximation of the derivative is used, some of the estimates obtained are invalid: It is obtained \textit{NaN} (not a number) or infinite values (when the estimate of a derivative is $0$). The average number of invalid observations in percents has been `MNL': $0 \%$, `SVM': $0 \%$, `RF': $1.0069 \%$, `XGBoost': $0.0373 \%$, `NN': $0.0008 \%$, `DNN': $0.0007 \%$. Furthermore, the data has been filtered by removing the outliers, using the interquartile range (IQR) with $k=1.5$. Figure \ref{fig:Experiment-3-WTP} shows the distribution of the actual \gls{wtp} for each synthetic dataset along with the distribution of the \gls{wtp} predictions obtained from each \gls{ml} method. The violin plot is a graphical representation used to compare probability distributions. It comprises a box that encompasses the range between the first and third quartiles, the median, and an estimated density plot on each side. To assess which method yields a distribution that more closely resembles the true distribution of the \gls{wtp}, the appearance of the actual \gls{wtp} in the first line should be compared with those returned by the methods set out in the following lines.

It is observed that the best methods for estimating the \gls{wtp} in the case of linear utilities are the \gls{mnl}, \gls{nn} and \gls{svm}. In the non-linear case, the best methods are the \gls{svm}, \gls{mnl} and the \gls{dnn}. That is, in the case of higher complexity of the utility functions, \gls{dnn} are more suitable than conventional \gls{nn}. It is also observed that the \gls{rf} and \gls{xgboost} methods, as their $s_i$ functions are neither monotonic nor differentiable, have greater variance, which sometimes leads to predicting a wrong sign for the \gls{wtp}.

\begin{figure}[!h]
	\begin{center}
		\includegraphics[width=0.90\textwidth, angle=0, origin=c]{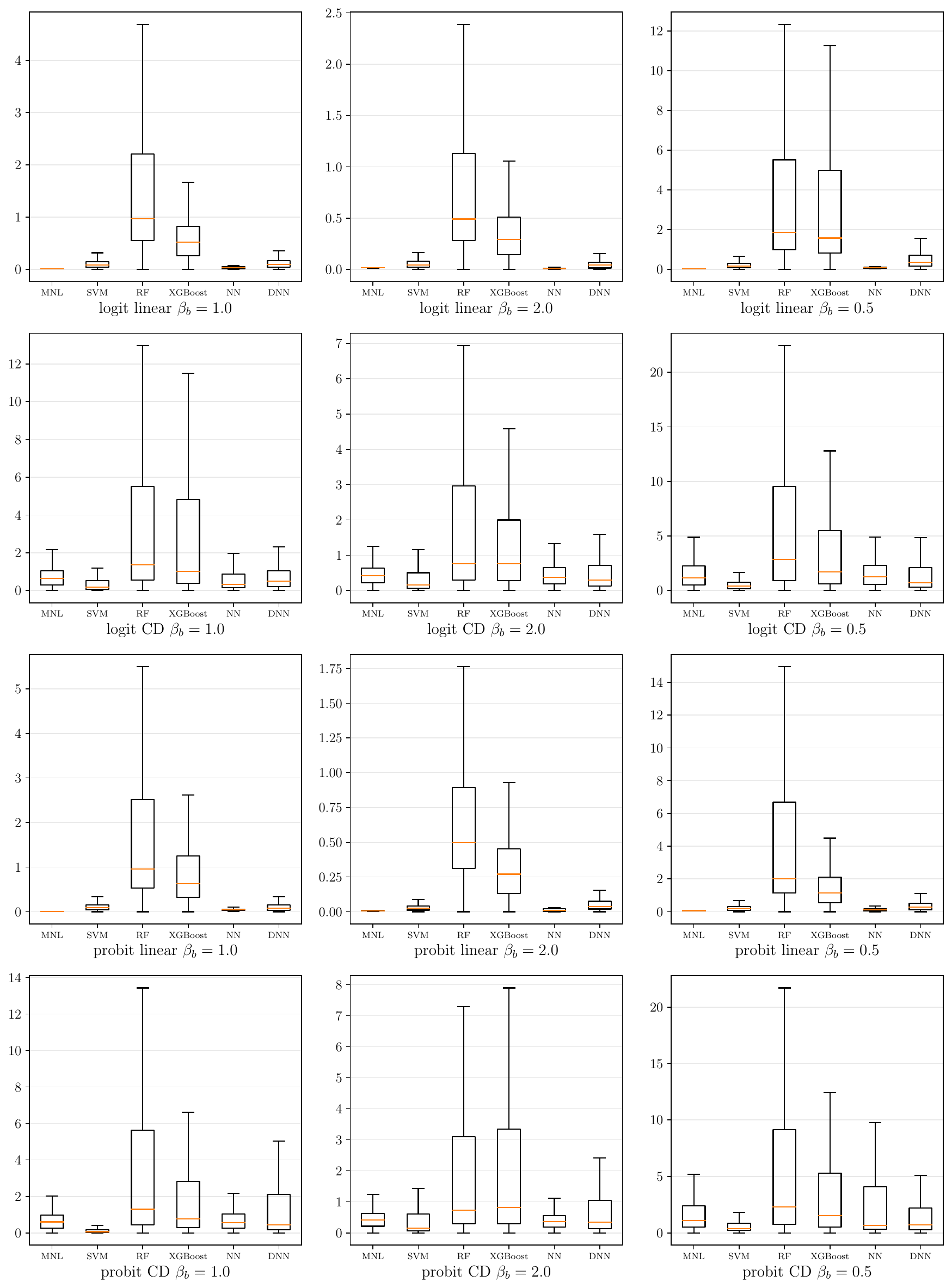}
		\caption{Difference in the WTP of the different models with respect to the actual WTP}
		\label{fig:Experiment-3-WTP-diff}
	\end{center}
\end{figure}

To further explore the conclusions obtained, Figure \ref{fig:Experiment-3-WTP-diff} shows the error distribution, i.e. the difference in absolute value of the \gls{wtp} of the different models with respect to the actual \gls{wtp} of each observation for each synthetic dataset. The box plot is a statistical tool that summarises the distribution of a variable and it is used to visualise the interquartile range and the median of the errors in estimating \gls*{wtp}. An efficient estimator should be unbiased and have minimum variance. The most optimal methods can be identified by observing the narrowest boxes (interquartile range) located at 0. This figure shows that the \gls{svm} performs adequately in both linear and non-linear utilities. The base \gls{mnl} model performs well, although in the non-linear case its performance declines.

Finally, in order to simplify the comparison, a value $\gls{wtp}_{*}$ will be introduced into the experiment, which allows the \gls{wtp} to be measured for the whole population. In particular, the median of the \gls{wtp} on the population was considered. The reason for choosing the median instead of the average value is because for Cobb-Douglas utilities the $\mathbb{E}(\gls{wtp}_{ni})=\infty$. \ref{App:WTP-CD} shows that in the case of the Cobb-Douglas utility the median of the \gls{wtp} on the population can be computed as:
\begin{equation}
  \gls{wtp}_{*}= \underset{n}{\hbox{Median}} \left (\gls{wtp}_{ni} \right )= \frac{\beta_x}{\beta_I}.
\end{equation}
Note that this expression also coincides with the one obtained in the case of linear utilities. 

Table \ref{tab:experiment_3_WTP_table} shows the actual value of $\gls{wtp}_{*}$ and the estimates obtained with the different methods on the synthetic data. The method achieving the best estimate is highlighted in bold. This experiment re-synthesises more clearly the results reported in the previous experiment. The \gls{svm} is the best-performing method in general, the \gls{mnl} also performs very well for linear utilities and \gls{dnn} obtains very good estimates for non-linear utilities, especially when the error term is normally distributed. 

\begin{table}
\centering
\caption{Estimated WTP by model on the synthetic datasets}
\label{tab:experiment_3_WTP_table}
\begin{tabular}{rlllllll}
\toprule
{} &  true &   MNL &   SVM &    RF &  XGBoost &    NN &   DNN \\
\midrule
logit linear $\beta_b=1.0$  &   1.0 &  \textbf{0.99} &  \textbf{0.99} &  0.48 &     0.92 &  0.97 &  0.96 \\
logit linear $\beta_b=2.0$  &   0.5 &  0.48 &  0.52 &  0.24 &     0.52 &  \textbf{0.50} &  0.49 \\
logit linear $\beta_b=0.5$  &   2.0 &  \textbf{2.03} &  2.11 &  1.41 &     2.29 &  2.08 &  2.16 \\
logit CD $\beta_b=1.0$      &   1.0 &  1.06 &  \textbf{0.98} &  0.55 &     0.77 &  0.94 &  0.91 \\
logit CD $\beta_b=2.0$      &   0.5 &  0.70 &  \textbf{0.59} &  0.13 &     0.40 &  0.62 &  \textbf{0.41} \\
logit CD $\beta_b=0.5$      &   2.0 &  1.73 &  1.79 &  0.77 &     1.49 &  \textbf{1.88} &  1.87 \\
probit linear $\beta_b=1.0$ &   1.0 &  \textbf{1.00} &  0.97 &  0.64 &     0.93 &  0.96 &  0.97 \\
probit linear $\beta_b=2.0$ &   0.5 &  0.49 &  \textbf{0.50} &  0.17 &     0.52 &  0.51 &  0.49 \\
probit linear $\beta_b=0.5$ &   2.0 &  \textbf{1.94} &  2.14 &  1.15 &     2.12 &  2.11 &  2.14 \\
probit CD $\beta_b=1.0$     &   1.0 &  0.98 &  0.95 &  0.32 &     0.79 &  0.96 &  \textbf{1.00} \\
probit CD $\beta_b=2.0$     &   0.5 &  0.68 &  \textbf{0.57} &  0.35 &     0.40 &  0.63 &  0.58 \\
probit CD $\beta_b=0.5$     &   2.0 &  1.65 &  1.65 &  1.00 &     1.52 &  1.52 &  \textbf{2.03} \\
\bottomrule
\end{tabular}
\end{table}

\subsection{Experiment 4: Real datasets}\label{sect:exp4}
Experiment $4$ repeats Experiments 1 and 3 carried out on synthetic data but now on three real transport mode choice datasets. In contrast to the previous experiments, the expected results are unknown, so the results can only be evaluated with respect to their internal consistency, and taking into account the conclusions obtained in the previous experiments. These real datasets are more complex than synthetic data because probabilistic assumptions may not be satisfied, the utility functions are non-linear, the number of attributes is higher and there are multiple decision makers.

Tables \ref{tab:experiment_4_train-1-train} and \ref{tab:experiment-4-test} show the accuracy and \gls{gmpca} indices obtained on training and test data, respectively. The results show that the \gls{xgboost} method exhibits the best performance. The \gls{xgboost} improves the accuracy of the \gls{mnl} on the test data (Optima, NTS and LPMC datasets) by $5.6\%, 3.9\%$ and $2.2 \%$ and the \gls{gmpca} by $7.8\%, 4.4\%$ and $3.0\%$. The key observation is that the differences in performance between methods is moderate, corroborating the results obtained in the synthetic data. This highlight is contrary to results collected in several articles in the literature, which show that \gls{ml} methods outperform \gls{mnl} with a large difference in accuracy index. For example, for the NTS dataset, \cite{HaH17} reports an accuracy of $57\%$ for the \gls{mnl} and $92\%$ for the \gls{rf}. This discrepancy with the results obtained in this study can be explained by several methodological errors. The first observation is that in the NTS dataset, each individual reported several observations (several trips made by the same individual). This fact was not taken into account by \cite{HaH17} when splitting the dataset into training and test, which leads to data leakage due to the simultaneous use of observations of the same individuals as both training and test data. The second reason is that \cite{HaH17} balanced the training and test data, which leads to a situation where the test data comes from a different distribution than in reality. However, if re-sampling is only applied to the training set, as opposed to the work by \cite{HaH17}, it can be confirmed that the authors' model overestimates by five to ten percent. 

When class balance is applied to the data, the probabilities of the classes are modified with the direct consequence of introducing a bias in the market shares. In terms of analysing user behaviour, it is essential to estimate the probabilities as accurately as possible, and this is the reason why data balancing is never used in \gls{dcm}. In the \gls{ml} community it is usual to apply a balancing operation on the training set when the data is highly imbalanced. This operation is motivated when the focus is on the classification and not on the probability estimation. Moreover, it is found that the balancing operation reduces the performance of the \gls{mnl} and increases that of the \gls{ml} methods.

Finally, the applicability of the algorithms is analysed, Table \ref{tab:experiments-time} shows the average estimation time needed to estimate each of the models on all the datasets. The computational cost of the \gls{mnl} depends on the number of covariates. This is the reason that, for synthetic datasets, it is observed that \gls{mnl} and \gls{rf} are the fastest algorithms. However, in real problems the \gls{mnl} method does not have a significantly lower computational cost than \gls{ml}. In our computational tests it is found that the \gls{svm} has to be applied in conjunction with the Nyström method on the NTS and LPMC datasets because otherwise the CPU execution time and RAM memory requirements would prohibit it. Furthermore, it is worth noting that \gls{xgboost} incurs a significantly higher computational cost than \gls{rf} when applied to the NTS and LPMC datasets. This may be attributed to certain hyperparameters which heavily impact the performance of the model. Specifically, for \gls{xgboost}, the number of boosting rounds tends to increase substantially for both datasets (see Table \ref{tab:hyperparameters-space}), possibly due to their larger number of observations and variables. Conversely, \gls{rf} is capable of achieving good results with a smaller number of decision trees. As a result, on the Optima problem, the performance difference between \gls{xgboost} and \gls{rf} is minimal, while it becomes more pronounced for the NTS and LPMC datasets.

\begin{table}
\centering
\caption{Accuracy and GMPCA results by model for each real dataset on the training set (using 5-CV)}
\label{tab:experiment_4_train-1-train}
\resizebox{\textwidth}{!}{
\begin{tabular}{rllllllllllll}
\toprule
{} & \multicolumn{2}{c}{MNL} & \multicolumn{2}{c}{SVM} & \multicolumn{2}{c}{RF} & \multicolumn{2}{c}{XGBoost} & \multicolumn{2}{c}{NN} & \multicolumn{2}{c}{DNN} \\
\cmidrule(lr){2-3} \cmidrule(lr){4-5} \cmidrule(lr){6-7} \cmidrule(lr){8-9} \cmidrule(lr){10-11} \cmidrule(lr){12-13} 
{} & Accuracy &  GMPCA & Accuracy &  GMPCA & Accuracy &  GMPCA & Accuracy &  GMPCA & Accuracy &  GMPCA & Accuracy &  GMPCA \\
\midrule
Optima &    75.87 &  55.95 &    77.31 &  57.19 &    76.86 &  57.99 &    \textbf{78.83} &  \textbf{60.65} &    77.69 &  57.23 &    77.23 &  56.37 \\
NTS    &     65.8 &  43.97 &    65.02 &  45.07 &    68.32 &   47.1 &    \textbf{69.16} &  \textbf{48.08} &    68.44 &  47.34 &    67.51 &  46.51 \\
LPMC   &    73.03 &  49.64 &    74.76 &  51.74 &    74.67 &  51.17 &    \textbf{75.69} &  \textbf{52.68} &    74.49 &  51.76 &    74.57 &  51.84 \\
\bottomrule
\end{tabular}
}
\end{table}

\begin{table}
\centering
\caption{Accuracy and GMPCA results by model for each real dataset on the test set}
\label{tab:experiment-4-test}
\resizebox{\textwidth}{!}{
\begin{tabular}{rllllllllllll}
\toprule
{} & \multicolumn{2}{c}{MNL} & \multicolumn{2}{c}{SVM} & \multicolumn{2}{c}{RF} & \multicolumn{2}{c}{XGBoost} & \multicolumn{2}{c}{NN} & \multicolumn{2}{c}{DNN} \\
\cmidrule(lr){2-3} \cmidrule(lr){4-5} \cmidrule(lr){6-7} \cmidrule(lr){8-9} \cmidrule(lr){10-11} \cmidrule(lr){12-13} 
{} & Accuracy &  GMPCA & Accuracy &  GMPCA & Accuracy &  GMPCA & Accuracy &  GMPCA & Accuracy &  GMPCA & Accuracy &  GMPCA \\
\midrule
Optima &    72.01 &  51.71 &    74.65 &   55.0 &    \textbf{78.17} &  58.24 &    77.64 &  \textbf{59.54 }&    75.53 &  48.76 &    75.53 &  54.03 \\
NTS    &    65.88 &  44.35 &    65.59 &  45.53 &    68.99 &  47.54 &    \textbf{69.82} &  \textbf{48.72} &     69.0 &  47.67 &    68.63 &  47.46 \\
LPMC   &    72.54 &  48.85 &     74.2 &  50.89 &    73.59 &  50.16 &    \textbf{74.72} &  \textbf{51.85} &    74.25 &  51.03 &    74.05 &  51.16 \\
\bottomrule
\end{tabular}
}
\end{table}

\begin{table}
\centering
\caption{Average estimation time by model (seconds)}
\label{tab:experiments-time}
\begin{tabular}{rllllll}
\toprule
{} &     MNL &     SVM &    RF & XGBoost &    NN &     DNN \\
\midrule
Synthetic data (avg.) &  0.18 &  4.77 &  0.31 &     7.52 &  1.69 &  16.01 \\
Optima &    0.88 &    1.66 &  0.17 &    0.87 &  1.84 &     2.1 \\
NTS    &  360.85 &  820.56 &  1.39 &  115.14 &  9.21 &  118.22 \\
LPMC   &  318.17 &  365.75 &  2.19 &   75.77 &  3.78 &   29.37 \\
\bottomrule
\end{tabular}
\end{table}

Next we will evaluate the market shares for the different transport modes available. The alternatives of the LPMC dataset are `walk', `cycle', `pt' and `drive' but have been renamed to `Walk', `Bike', `Public Transport' and `Car' to maintain consistency with the NTS dataset. Table \ref{tab:experiment_4_MS_table} reports the market shares obtained and the behaviour is similar to that obtained in the synthetic data. All methods estimate the same market shares (with a difference between estimates of less than $1\%$) with the exception of the \gls{dnn} method in the NTS problem and the \gls{xgboost} method for the Optima.

\begin{table}
\centering
\caption{Estimated market shares by model for each real dataset on the test set}
\label{tab:experiment_4_MS_table}
\resizebox{\textwidth}{!}{
\begin{tabular}{rlllllllllll}
\toprule
{} & \multicolumn{3}{c}{Optima} & \multicolumn{4}{c}{NTS} & \multicolumn{4}{c}{LPMC} \\
\cmidrule(lr){2-4} \cmidrule(lr){5-8} \cmidrule(lr){9-12} 
{} & Public transport & Private modes & Soft modes &       Walk &   Bike & Public Transport &        Car &   Walk &  Bike & Public Transport &        Car \\
\midrule
MNL     &        27.82 &     66.93 &       5.25 &  16.40 &  24.16 &             4.17 &  55.27 &  16.84 &  2.86 &        36.24 &  44.07 \\
SVM     &        27.98 &     66.34 &       5.68 &  16.00 &  24.32 &             3.84 &  55.85 &  16.74 &  2.92 &        36.10 &  44.24 \\
RF      &        27.43 &     67.42 &       5.16 &  16.20 &  24.43 &             4.05 &  55.32 &  16.94 &  2.91 &        36.35 &  43.80 \\
XGBoost &        26.61 &     68.16 &       5.23 &  16.21 &  24.32 &             4.01 &  55.45 &  16.98 &  2.91 &        36.01 &  44.00 \\
NN      &        27.09 &     67.14 &       5.78 &  16.14 &  24.35 &             4.06 &  55.45 &  16.84 &  2.86 &        36.19 &  44.12 \\
DNN     &        27.88 &     66.98 &       5.14 &  14.39 &  23.68 &             3.92 &  58.01 &  16.52 &  3.01 &        36.98 &  43.49 \\
\bottomrule
\end{tabular}
}
\end{table}

The experiments will conclude by computing a special case of \gls{wtp}: the \gls{vot}. This index calculates how much money a traveller would be willing to pay to reduce one unit (hour, in this case) in his travel time. In order to calculate this magnitude, the dataset is required to contain the time and cost attributes of the alternatives. As the NTS dataset is not based on surveys, it does not contain this information and therefore cannot be used in this experiment. The \gls{vot} has been computed in both `Public Transport' and the `Car' alternative. In the LPMC dataset, the `Public Transport' alternative includes both bus and train modes, hence, the magnitude of travel time (in vehicle) is different between users. Moreover, the time is perceived differently (and consequently the \gls{vot} for each public transport mode) depending on whether the traveller is waiting, in the vehicle or walking to connect the public transport network with his/her origin/destination. For these reasons, in the LPMC dataset we have used the access time to the public transport network as the time variable, as this variable is shared by all public transport travellers. 

To estimate the \gls{vot}, similarly to the synthetic datasets, the derivatives have been numerically approximated using Equation~(\ref{numerical_derivative}). The value of $h$ chosen is a $5\%$ of the standard deviation of the attribute (time or cost). 

The \gls{mnl} method has the disadvantage over the \gls{ml} methods that it requires the specification of the utility function. The modeller could choose a linear function but this choice is not without problems. It is found in both datasets that a linear model can lead to an inconsistent sign in the \gls{vot} (negative output). For example, in the case of the Optima dataset, if the distance attribute is omitted in the car alternative, the \gls{vot} obtained is negative. This is due to the fact that users prefer the car alternative for long duration trips and this is captured by the model in the travel cost variable in the absence of the distance variable. A second example, in the `Public Transport' alternative of the LPMC dataset there is one variable, `cost\_transit', that accounts for the estimated final cost of public transport route, including the train and bus discounts. If `cost\_transit' is introduced in the utility function simultaneously with the variables that measure the fare discounts (such as `faretype' or `bus\_scale'), then the sign of the `cost\_transit' variable is confused on the estimation of the \gls{mnl} model.

\begin{figure}[!h]
	\begin{center}
		\includegraphics[width=0.90\textwidth, angle=0, origin=c]{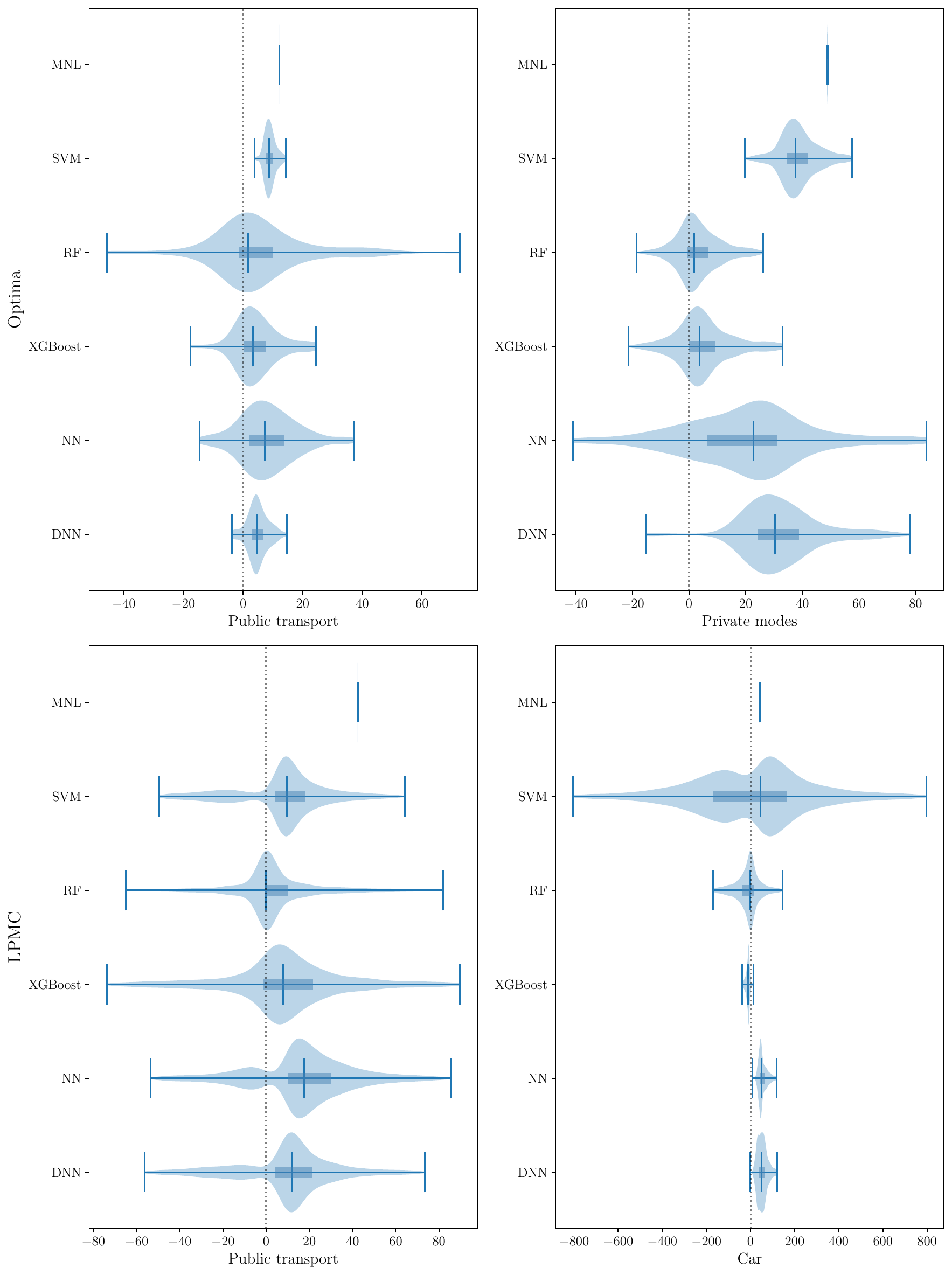}
		\caption{Comparison of the VOT of the different models on the real datasets}
		\label{fig:Experiment-4-WTP}
	\end{center}
\end{figure}

Once this problem has been identified and solved, the results obtained are shown in Figure \ref{fig:Experiment-4-WTP}. Then, the first question is to identify which methods can be valid for estimating the \gls{vot}. Since the \gls{vot} is approximated numerically, some methods such as those based on decision trees may give problems when calculating the derivative of the probability. Firstly, the invalid values estimated by some methods, such as infinity values or division by zero, were filtered out. The number of non-valid observations in the Optima dataset is $0\%$ for the \gls{mnl} and \gls{svm} methods, $49.96\%$ for the \gls{rf}, $56.98\%$ for the \gls{xgboost}, $0.91\%$ for the \gls{nn}, and $0.27\%$ for the \gls{dnn}. In the case of the LPMC dataset the non-valid observations are $0\%$ for the \gls{mnl}, \gls{svm}, and \gls{nn} methods, $40.57\%$ for the \gls{rf}, $42.92\%$ for the \gls{xgboost}, and $0.01\%$ for the \gls{dnn}.

Table \ref{tab:experiment_4_WTP_table} shows the median of the \gls{vot} estimates of all the individuals in the sample. The fact that a method estimates many individuals with negative \gls{vot} values is an indication of the invalidity of the method. An estimate is considered to be inconsistent if the $0$ value is in the interval defined within the first and third quartiles. Accordingly, the inconsistent estimates have been labelled with $\dag$ .   Table \ref{tab:experiment_4_WTP_table} shows that the \gls{rf} and \gls{xgboost} (the one with the highest performance) always generate inconsistent estimates. The \gls{svm} is inconsistent only for the alternative `Car' of the LPMC dataset. Only the \gls{nn} and \gls{dnn} are consistent in all situations. The larger discrepancy in the estimation of the \gls{vot} by the \gls{nn} in the Optima dataset may be due to the fact that they report a lower value of the \gls{gmpca} in the test set.

The \gls{mnl} method can be considered as the baseline method and the \gls{ml} methods as tools to correct the estimate given by the \gls{mnl}. Although the methodology for doing so needs to be investigated, a first criterion may be to report as the \gls{vot} value the consistent estimate closest to the one obtained by the \gls{mnl}, which is marked in bold in the table. It is observed that this approach can correct the estimates of \gls{mnl} in three of the cases, but in the case of `Public Transport' alternative for the LPMC dataset the estimate given by all the methods is significantly different from the one obtained by \gls{mnl}.

\begin{table}
\centering
\caption{Median of the estimated VOT values by model for each real dataset on the test set}
\label{tab:experiment_4_WTP_table}
\begin{tabular}{rllll}
\toprule
{} & \multicolumn{2}{c}{Optima} & \multicolumn{2}{c}{LPMC} \\
\cmidrule(lr){4-5} \cmidrule(lr){2-3} 
{} & Public transport & Private modes & Public Transport &        Car \\
\midrule
{\bf MNL}     &        12.12 &            48.81&        42.39 &  41.62 \\
SVM     &         \bf 8.74 &            \bf 37.56 &         9.60 &  45.16 $\dag$ \\
RF      &         1.73 $\dag$ &      1.75 $\dag$&         0.00$\dag$ &  -4.02 $\dag$\\
XGBoost &         3.39 $\dag$ &       3.67$\dag$ &         7.77 $\dag$ & -10.33 $\dag$ \\
NN      &         7.27 &            22.79 &        \bf 17.46 &  \bf 49.10 \\
DNN     &         4.63 &            30.27 &        11.99 &  50.94 \\
\bottomrule
$\dag$ Inconsistent estimation
\end{tabular}
\end{table}

\subsubsection{Post-hoc interpretation of ML and MNL approaches}
The \gls{mnl} methods provide statistical tests that permit evaluation of whether a regressor affects the choice response. Explainable \gls{ml} methods offer alternative approaches such as \gls{shap} values to perform this task. The goal of this experiment is to determine whether applying \gls{shap} values to \gls{ml} models produces results consistent with those obtained through statistical methods by \gls{mnl}.

The \gls{shap} value \citep{LuL17} is a model-agnostic method that can be employed to explain the predictions made by any \gls{ml} model. This method allows the contribution of each feature in the prediction process to be quantified, enhancing understanding of how the model arrives at its predictions. \gls{shap} values have three key advantages: Firstly, they provide global interpretability by not only indicating the importance of features, but also revealing whether each feature has a positive or negative impact on predictions. Secondly, they enable local interpretability, as \gls{shap} values are calculated for individual predictions, which reveals how each feature contributes to specific predictions. This level of granularity is not available in other techniques that provide only aggregated results for the entire dataset. Lastly, \gls{shap} values can be employed to explain a wide range of models, including linear models, tree-based models, and neural networks. This versatility makes \gls{shap} values an extremely useful tool for explaining the inner workings of the \gls{ml} models.

The computation of \gls{shap} values involves a concept called Shapley values from cooperative game theory. To compute \gls{shap} values, an intuitive approach is to consider all possible subsets of features and evaluate their contributions. However, this becomes computationally infeasible as the number of features grows. To address this, \gls{shap} values rely on a more efficient sampling-based approach. The algorithm starts by creating a reference or baseline prediction, typically the average prediction of the dataset. It then iterates through each data point, and samples different subsets of features. For each subset, it evaluates the contribution of the selected features by comparing the prediction with and without those features. This process is repeated multiple times to obtain a representative sample of subsets. Next, the algorithm computes the Shapley value for each feature by considering all possible permutations of the order in which the features are added to the subsets. It calculates the marginal contribution of each feature in each permutation and averages these contributions over all permutations. This averaging process ensures fairness and consistency in the distribution of importance across features. Finally, the Shapley values thus computed are scaled, to guarantee that the sum of \gls{shap} values aligns with the prediction of the model. 
Taking into account an alternative $j$ in our analysis, all of these processes lead to the following expression for \gls{shap} values:
\begin{equation}
    \phi_k( \mathbf{x}_{nk})= \sum_{S \subseteq \{1,2,\cdots,k-1,k+1,\cdots K\} }\dfrac{|S|! (K-|S|-1)!}{K!} \left ( \mathbb{E} \left [s_j( \mathbf{x}_{nk}| \mathbf{x}_{n S\cup\{k\}}) \right ]-\mathbb{E} \left[s_j( \mathbf{x}_{nk}| \mathbf{x}_{n S}) \right] \right ),
\end{equation}
where $\mathbb{E} \left[s_j( \mathbf{x}_{nk}| \mathbf{x}_{n S}) \right]$ is the expected value of the $s_j(\mathbf{x}_{nk})$ conditioned by the features present in the set $S$.

This experiment will analyse the \gls{shap} values obtained by each model on the real datasets. Specifically, a kernel explainer is employed within the \gls{shap} method to compute the importance of each feature. The goal is to examine whether the features identified as important by different methods correspond to those identified using the \gls{mnl} coefficients. Due to limitations of space and to maintain consistency with Experiment 3, where only the \gls{vot} of the Optima and LPMC datasets were analysed, the same datasets will be reported in this experiment. The complete results for all datasets are available at the GitHub repository, along with the tables containing the \gls{mnl} coefficients obtained.

To determine the impact of a feature, the focus is on evaluating the disparity in probabilities when a feature is present versus when it is ``missing''. The \gls{mnl} model allows for the interpretation of the effect of the explanatory variable $\mathbf{x}_{nk}$ on the outcome $y_n$ through the use of regression coefficients. When all features are standardised, with $\mathbb{E}(\mathbf{x}_{nk})=0$ and $\mathbb{V}(\mathbf{x}_{nk})=1$, the coefficients estimated are referred to as standardised or beta coefficients. This enables a direct comparison of these beta coefficients. The importance of a feature, $\mathbf{x}_{nk}$, can be assessed based on the absolute value of its beta coefficient. A larger value $|\beta_k|$ indicates greater importance.

The importance of a feature can also be assessed through its \gls{shap} value, which is directly related to its proportional effect determined by coefficient $\beta_k$ in the case of the \gls{mnl} model. For \gls{mnl} models, the relationship between a feature $\mathbf{x}_{nk}$ and its corresponding \gls{shap} value, see Corollary 1 of \citet{LuL17}, can be expressed as follows:
\begin{equation}
    \label{eq:linear-SHAP}
    \phi_k (\mathbf{x}_{nk})=\beta_k \mathbf{x}_{nk}- \mathbb{E}(\beta_k\mathbf{x}_{nk})=
    \beta_k \mathbf{x}_{nk}- \beta_k \mathbb{E}(\mathbf{x}_{nk})=\beta_k \mathbf{x}_{nk}.
\end{equation}

\begin{figure}[hp]
	\begin{center}
		\includegraphics[width=1\textwidth, angle=0, origin=c]{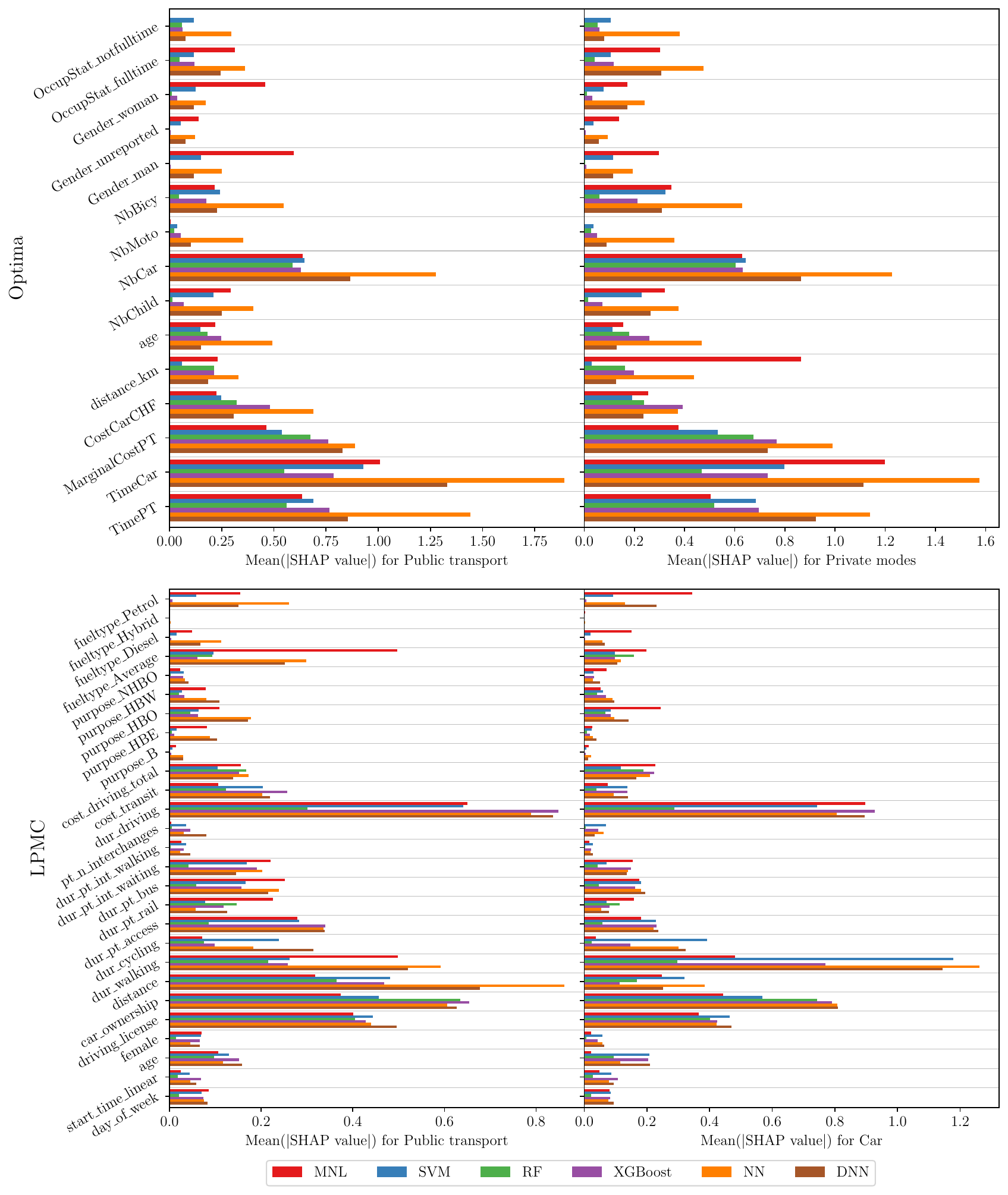}
		\caption{Comparison of the SHAP values of the different models on the Optima and LPMC datasets}
		\label{fig:Experiment-5-Comparative}
	\end{center}
\end{figure}

The mean absolute \gls{shap} value can be used to determine the importance of each feature. Figure \ref{fig:Experiment-5-Comparative} compares this index for each model, specifically for the alternatives: Public Transport and Private modes (Car). This is achieved by using the sample estimate of $\mathbb{E} \left [ | \phi_k (\mathbf{x}_{nk}) | \right  ]$. In the case of linear \gls{mnl}, this can be obtained by applying Equation~(\ref{eq:linear-SHAP}):
\begin{equation}
   \label{eq:expected-average-mnl}
   \mathbb{E} \left [ | \phi_k (\mathbf{x}_{nk}) | \right  ] = |\beta_k| \mathbb{E} (| \mathbf{x}_{nk}|).
\end{equation}

This study determined the importance of a variable in a given \gls{mnl} model if the absolute coefficient value, $|\beta_k|$, is greater than or equal to $1$. This criterion is applicable because all the features were standardised before model estimation. Using Equation~(\ref{eq:expected-average-mnl}), it can be equivalently stated that for \gls{mnl} models, this criterion corresponds to considering that its \gls{shap} value exceeds $0.5$. When the covariates $\mathbf{x}_{nk}$ follow a normal distribution, then $\mathbb{E} \left [ | \phi_k (\mathbf{x}_{nk}) | \right ] = 0.499 |\beta_k|$. In other words, considering variables with \gls{shap} values greater than or equal to $\mathbb{E} \left [ | \phi_k (\mathbf{x}_{nk}) | \right ] \ge 0.5$ is equivalent to considering variables with \gls{mnl} model coefficients $|\beta_k| \ge 1$.

Analysing Figure \ref{fig:Experiment-5-Comparative}, it is observed that certain variables emerge as highly relevant for the Optima dataset, namely {\tt NbCar, MarginalCostPT, TimeCar}, and {\tt TimePT}. This outcome aligns with the analysis of the beta coefficients in the \gls{mnl} model. However, a specific discrepancy arises regarding the variable {\tt NbCar} (number of cars in the household). This inconsistency may arise from the discrete nature and non-normal distribution of this variable. As a result, the approximation $\mathbb{E}(|\mathbf{x}_{nk}|)=0.499$ is clearly erroneous in this case.

Examining the results for the LPMC dataset, the variables {\tt distance, dur-driving, dur-walking}, and {\tt car-ownership} are identified as the most important ones for the prediction. These variables correspond to those identified by the \gls{mnl} model, except for {\tt car-ownership}. Once again, this discrepancy can be attributed to the binary nature of the variable.

Overall, there is consensus among the methods as they consistently identify the same variables as the most important ones in the decision process. It can be observed that the methods returning lower values of the \gls{gmpca} index on the test set (Table~\ref{tab:experiment-4-test}), are also those that show more inconsistent, in this case higher, values of the \gls{shap} index. A clear example of this is the \gls{nn} and \gls{mnl} models on the Optima dataset.

\begin{figure}[hp]
	\begin{center}
		\includegraphics[width=1\textwidth, angle=0, origin=c]{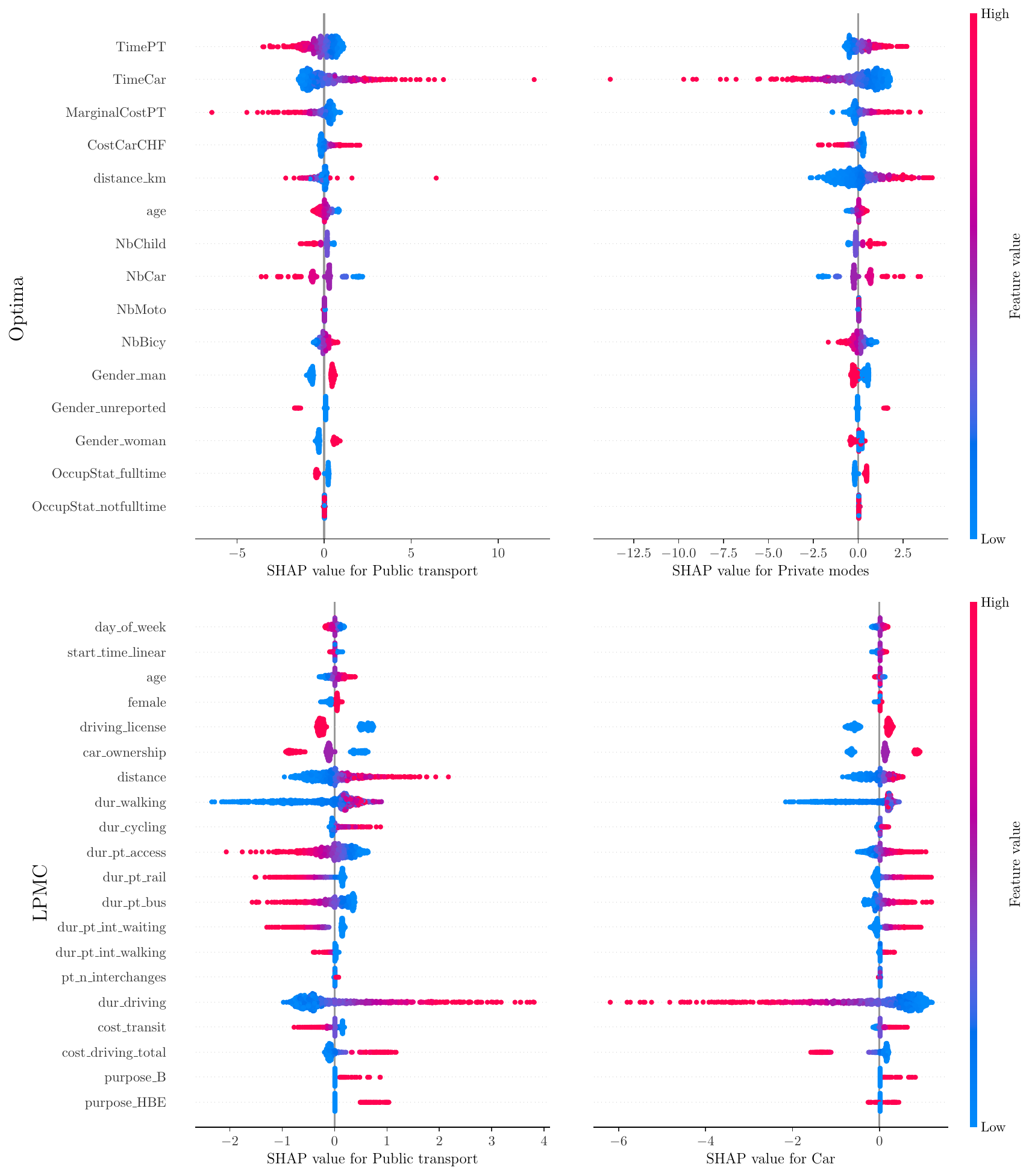}
		\caption{Bee-swarm plot of the SHAP values of the MNL model}
		\label{fig:Experiment-5-MNL-beeswarm}
	\end{center}
\end{figure}

\begin{figure}[hp]
	\begin{center}
		\includegraphics[width=1\textwidth, angle=0, origin=c]{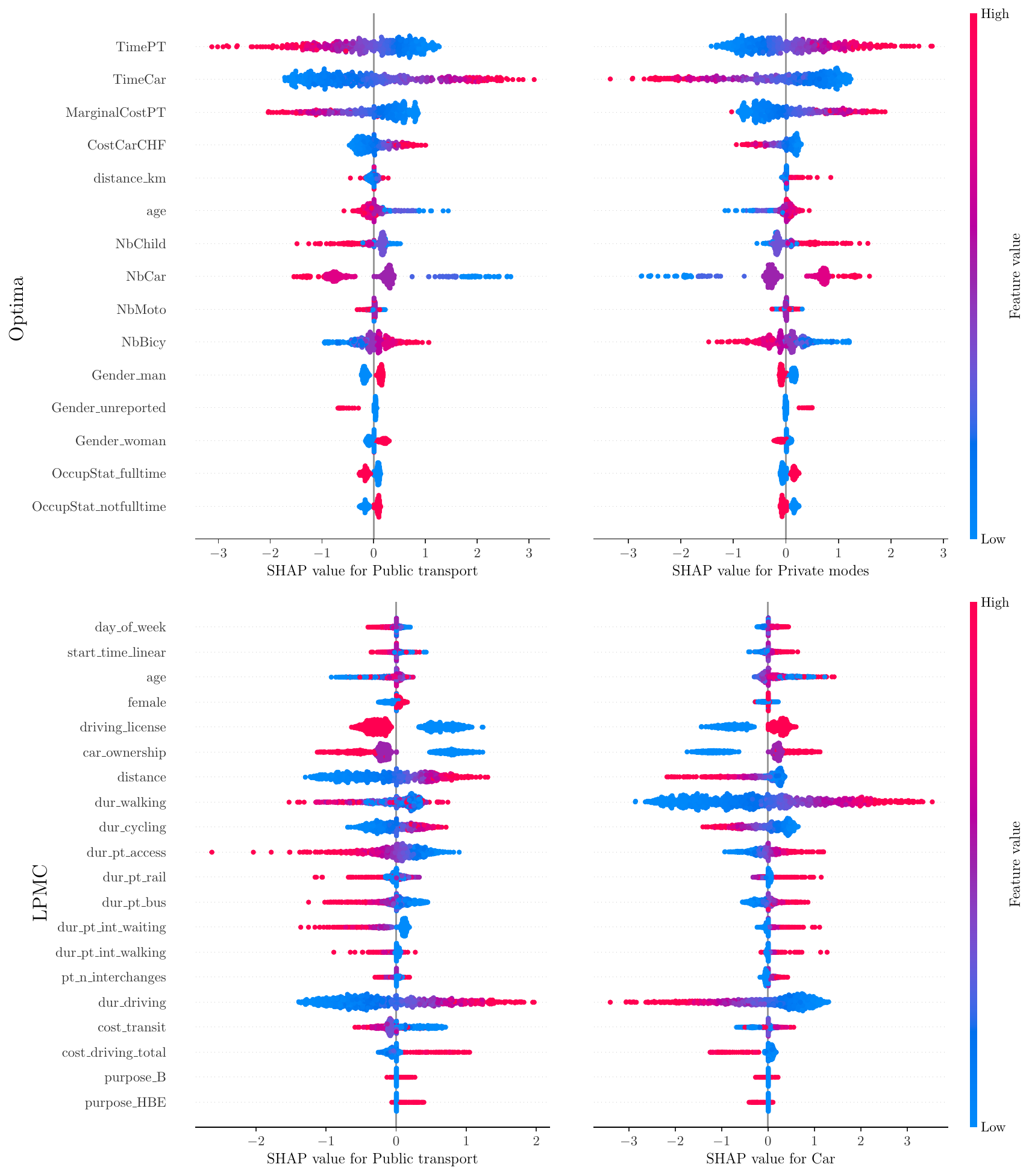}
		\caption{Bee-swarm plot of the SHAP values of the SVM model}
		\label{fig:Experiment-5-SVM-beeswarm}
	\end{center}
\end{figure}

\begin{figure}[hp]
	\begin{center}
		\includegraphics[width=1\textwidth, angle=0, origin=c]{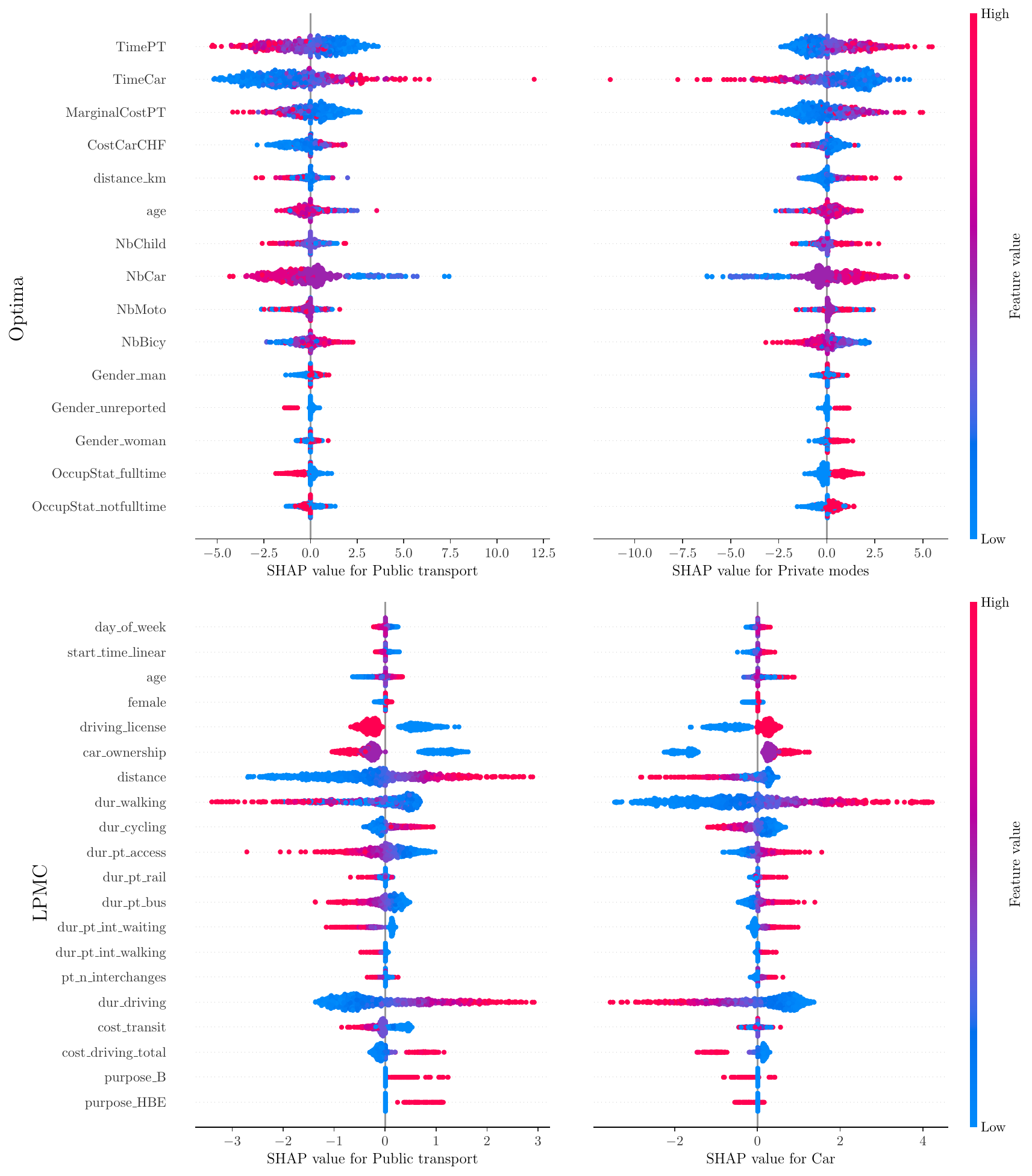}
		\caption{Bee-swarm plot of the SHAP values of the NN model}
		\label{fig:Experiment-5-NN-beeswarm}
	\end{center}
\end{figure}

To analyse the direction of the effect of covariates, which is equivalent to the sign of the coefficients on the \gls{mnl} models, a bee swarm plot was generated. This technique allows the use of overlapping scatterplots to represent the distribution and importance of \gls{shap} values for the different features in each dataset. Figures \ref{fig:Experiment-5-MNL-beeswarm}, \ref{fig:Experiment-5-SVM-beeswarm}, and \ref{fig:Experiment-5-NN-beeswarm} depict the bee swarm plots for the \gls{mnl}, \gls{svm}, and \gls{nn}, respectively. The reason for selecting these three models is that the \gls{mnl} model can act as the baseline, while the \gls{svm} and \gls{nn} were the \gls{ml} methods that returned the best performances in calculating the \gls{vot} in the previous experiment. However, the bee swarm plots for the remaining methods are provided as supplementary material on the GitHub repository.

The interpretation of these plots is as follows: a positive \gls{shap} value indicates an increase in the probability of choosing the alternative. By examining the colour, it can be determined whether high or low values of the covariate correspond to an increase in the probability. If the colour follows the same order as the scale shown, it indicates a positive correlation, while a contrary order indicates a negative correlation. For instance, when analysing the {\tt TimePT} variable for the Public Transport alternative in the Optima dataset, it is observed that the points start as red and transition to blue, indicating an inverse order compared to the scale, and thus a negative correlation. This suggests that as the time spent on public transport increases, users are less likely to choose this alternative. Therefore, \gls{shap} values enable a post-hoc analysis of \gls{ml} models that helps determine such correlations.

Furthermore, the inclusion of bee swarm plots for the \gls{svm} and \gls{nn} models demonstrates the ability of these methods to capture nonlinear effects. For example, when examining the {\tt Age} variable for Private modes, it is seen that children and older adults tend not to choose the car, while middle-aged individuals do. This pattern cannot be adequately captured by linear terms alone, such as in \gls{mnl} (see Figure \ref{fig:Experiment-5-MNL-beeswarm}).

\section{Conclusions and future work}\label{sect:conclusions}
This study compares the \gls{mnl} and the most promising \gls{ml} methods for the transport mode choice problem. This comparison uses both synthetic data, which provides a ground truth for comparison, and real problems. This comparison has covered both performance aspect (canonical comparison) and behavioural analysis. The theoretical analysis presented in this study demonstrates that the correct estimation of the econometric indicators of \gls{ml} models cannot be guaranteed by the training problem alone.

Regarding the performance study, it has been found that all methods are able to achieve maximum accuracy for all synthetic problems, i.e. they are able to learn the systematic part of the utility. The performance improvements of the methods of \gls{ml} versus \gls{mnl} appear mainly on real data, although they are moderate. The maximum differences observed are $5.6\%, 3.9\%$ and $2.2 \%$ for accuracy and $7.8\%, 4.4\%$ and $3.0\%$ for \gls{gmpca}, with \gls{xgboost} as the most promising method in terms of performance. 

The conclusion drawn in this paper is that there is a slight predictive performance advantage of \gls{ml} classifiers over linear Logistic Regression models, but this advantage is much more modest than that suggested by previous research studies, which is mainly attributable to failures in the validation schemes of the methods in these studies \citep{Hil21}. This result is consistent with the empirical findings reported in \citet{WMH21}.

Regarding the behavioural analysis on the synthetic data, it is shown that the \gls{xgboost} and \gls{rf} exhibit poor extrapolation properties, suffer from numerical problems in the calculation of derivatives as they fail to estimate monotonic probabilities (increasing or decreasing) against attributes, and have lower performance in the calculation of the \gls{wtp} than the \gls{svm}, \gls{nn} or \gls{dnn} methods. These extrapolation problems of the \gls{rf} method were also identified out by \cite{ZYY20}. On the other hand, the \gls{mnl} method with linear utilities has shown very good results in almost all problems and experiments. Only in the case of problems with non-linear utilities and with a use outside the data domain (extrapolation) does it seem advisable to use \gls{ml} methods, with \gls{nn} and \gls{svm} being the best options. More specifically, in the estimation of market shares on the synthetic data for the $S2$ scenario with non-linear utilities is where the performance of the \gls{mnl} method declines, and \gls{nn} and the \gls{svm} method are more appropriate.

The \gls{svm} is not a probabilistic classification method, therefore this study introduces probabilities by means of cross-validation and calibration. The result is a robust method that obtains the best results for the estimation of \gls{wtp} on synthetic data and also obtains consistent results on almost all real datasets.

Analysing the real datasets, it is found that \gls{xgboost}, which is the best method with respect to predictive performance, returns inconsistent results in the calculation of \gls{vot}. This suggests that the \gls{ml} methods should be analysed simultaneously from a performance and behavioural analysis perspective. Another piece of evidence that shows that an analysis only on the performance of \gls{ml} methods is insufficient is the large discrepancy in the estimates of \gls{vot} compared to the difference in the performance measures accuracy and \gls{gmpca}. The only \gls{ml} methods that do not show inconsistent behaviour in all real scenarios are \gls{nn} and \gls{dnn}. These methods also showed the best results in estimating the \gls{wtp} for a given type of non-linear utilities in the synthetic data. It seems that for highly non-linear cases these are the most recommended methods. 

It has also been shown that \gls{ml} methods offer alternative approaches such as \gls{shap} values to assess the importance of predictors and the direction of their relationships, yielding results consistent with those obtained using \gls{mnl}. This enables post-hoc analysis of the models, allowing for further insights into the relationships between variables.

The results obtained from the \gls{mnl} in this study are robust across all the situations analysed. This, together with its strong theoretical framework, position it as a method of reference. These findings indicate that \gls{ml} methods do not aim to replace \gls{rum} methods but instead highlight the potential of combining \gls{ml} and \gls{rum} techniques for modelling travel mode choice. Therefore, two areas of further research need to be explored to integrate \gls{ml} techniques into practice: i) the selection of the appropriate \gls{ml} methods for a given problem, and ii) the integration of estimates obtained from different methods.

Regarding i), the so-called {\sl automated machine learning} or AutoML, may be a promising  avenue to explore. AutoML automates the slow and iterative tasks involved in developing a \gls{ml} model. It can begin with a pool of algorithms that have shown acceptable performance and the ability to derive econometric indicators in the article and obtain the best algorithm for the problem. AutoML relaxes the dependence of algorithm results on data, and allows for considering a set of algorithms as a solution, rather than just one algorithm.

Regarding ii), the \gls{ml} methods may serve as an exploratory tool used to identify nonlinear effects. There is a need for methodologies to identify how and when to integrate the estimates derived from the \gls{ml} methods with those obtained with the \gls{mnl}. In this paper we propose a screening of the results of the \gls{wtp} taking into account the inconsistency of the values obtained by the methods, and giving as an estimate the median of the consistently estimated values. The estimated \gls{wtp} values are close to those estimated using \gls{mnl} model.

Another line of research is to enhance the training of the \gls{dnn}. This method is the most promising in synthetic data for some non-linear situations, however, the main challenge of this method is that it is highly sensitive during the training process, an issue pointed out in \cite{WWZ20}. It can be proven theoretically that the maximum likelihood estimate of the \gls{mnl} reproduces the observed market shares. This property is also fulfilled in the numerical experiments for all the \gls{ml} methods, except on some occasions for the \gls{dnn} (in half of the synthetic dataset and in the NTS dataset) and the \gls{xgboost} (in the Optima dataset). This may be an indication that a global optimum of the training problem has not been found for these cases, as \gls{dnn} are non-convex and high-dimensional.

Finally, all the codes developed in the experimental results section, as well as the synthetic datasets and the scripts to generate and process the different datasets, are available in a GitHub repository. This allows the reproducibility of the experiments and their application to other contexts or areas or research.

\section*{Acknowledgements}
This work was supported by grant PID2020-112967GB-C32 funded by MCIN/AEI/10.13039/501100011033 and by {\sl ERDF A way of making Europe}. The research of Mart\'in-Baos has been supported by the FPU Predoctoral Program of the Spanish Ministry of University with reference FPU18/00802.

\printglossary[type=\acronymtype,title=Abbreviations]

\bibliographystyle{elsarticle-harv} 
\bibliography{references}






\appendix

\section{Proof of Theorem \ref{th1}}
\label{App:Proof-th1}

\begin{proof}
Let $\varepsilon_0>0$ be an arbitrary value. From the condition of null measure set of ${\cal X}_o$, there exists a countable set $\{B_1,\cdots, B_s\}$ of open Cartesian products in $\mathbb{R}^d$ satisfying ${\cal X}_o \subseteq B=\bigcup_{k=1}^s B_k$. Furthermore, the measure of this set satisfies  $\mu(B) < \varepsilon_0$. Note that ${\cal X}-B$ is a compact set because $\cal X$ is compact by hypothesis and $B$ is an open set.

Define the function $\phi(\mathbf{x})=\underset{i\neq j}{\min}|P_i(\mathbf{x})-P_j(\mathbf{x})|$, which is continuous since the probability functions $\{P_j(\mathbf{x})\}_{j \in C}$ are continuous. Consider the following optimisation problem:
\begin{equation}
\label{eq:th1}
    \underset{\mathbf{x}\in {\cal X}-B}{\min} \phi(\mathbf{x}).
\end{equation}
The Weierstrass extreme value theorem, due to the minimisation of a continuous function over the compact set ${\cal X}-B$, ensures that the  problem (\ref{eq:th1}) has a minimum:
\begin{equation}
    \phi(\mathbf{x}^*)=\underset{\mathbf{x}\in {\cal X}-B}{\min} \phi(\mathbf{x}).
\end{equation}
Since $\mathbf{x}^* \notin B$, it follows that $\mathbf{x} \notin {\cal X}_o$. Therefore, $\phi(\mathbf{x}^*)>0.$
Given $\varepsilon=\frac{\phi(\mathbf{x}^*)}{2}>0$, there exists $\boldsymbol{\omega}$ such that
\begin{eqnarray}
\left |P_i (\mathbf{x})-s_i(\mathbf{x}; \boldsymbol{\omega }) \right | \le  \frac{\phi(\mathbf{x}^*)}{2}, \, \, \hbox{ for all } i.
\end{eqnarray}

We will prove
\begin{equation}
\label{eq:th2}
    f(\mathbf{x})=h(\mathbf{x};\boldsymbol{\omega}), \,\,\hbox{ for all } \mathbf{x} \in {\cal X}-B.
\end{equation}

Given $\mathbf{x} \in {\cal X} - B$. Assume that $f(\mathbf x)=i$, hence $P_i(\mathbf{x})\ge P_j(\mathbf{x})$. The definition of $\phi$ and $\mathbf{x}^*$ gives:
\begin{equation}
    P_i(\mathbf{x})- \frac{\phi(\mathbf{x}^*)}{2} \ge  P_j(\mathbf{x})+ \frac{\phi(\mathbf{x}^*)}{2} \hbox{ with } i\ne j.
\end{equation}
Using the relationship (\ref{eq:th1}) gives:
\begin{equation}
\label{eq:th4}
    s_i(\mathbf{x};\boldsymbol{\omega }) \ge P_i(\mathbf{x})- \frac{\phi(\mathbf{x}^*)}{2} \ge  P_j(\mathbf{x})+ \frac{\phi(\mathbf{x}^*)}{2} \ge  s_j(\mathbf{x};\boldsymbol{\omega }) \hbox{ with } i\ne j.
\end{equation}

From (\ref{eq:th4}) we deduce that $h(\mathbf{x};\boldsymbol{\omega})= \underset{j\in C}{\hbox{arg max }} \,\,s_{j}(\mathbf{x};\boldsymbol{\omega})=i$. Therefore, from the optimality of $\mathbf{\omega}_D$
and the definition of $R_D(\mathbf{\omega})$, we obtain
\begin{eqnarray}
\nonumber
R_D(\mathbf{\omega}_D)\le R_D(\mathbf{\omega}) = \mathbb{E}_{(\mathbf{x},y)\sim \mathbb{P}({\cal X},C) }\ell (h(\mathbf{x}; \boldsymbol \omega),y)
= \int_{\cal X} \sum_{i\in C} \ell(h(\mathbf{x};\boldsymbol{\omega}),i)P_i(\mathbf{x}) \psi(\mathbf{x}) {\rm d} {\mathbf x}\\ \label{eq:th3}
=\int_{{\cal X}-B} \sum_{i\in C} \ell(h(\mathbf{x};\boldsymbol{\omega}),i)P_i(\mathbf{x}) \psi(\mathbf{x}) {\rm d} {\mathbf x}+
\int_{B} \sum_{i\in C} \ell(h(\mathbf{x};\boldsymbol{\omega}),i)P_i(\mathbf{x}) \psi(\mathbf{x}) {\rm d} {\mathbf x}.
\end{eqnarray}

By using the non-negativity of the integrating function and Equation~(\ref{eq:th2}), we can bound the Equation~(\ref{eq:th3}) as follows:
\begin{eqnarray}
\nonumber
R_D(\mathbf{\omega}_D)\le R_D(\mathbf{\omega}) \le 
\int_{{\cal X}} \sum_{i\in C} \ell(f(\mathbf{x}),i)P_i(\mathbf{x}) \psi(\mathbf{x}) {\rm d} {\mathbf x}+
\int_{B} \sum_{i\in C} \ell(h(\mathbf{x};\boldsymbol{\omega}),i)P_i(\mathbf{x}) \psi(\mathbf{x}) {\rm d} {\mathbf x}\\
=\mathbb{E}_{(\mathbf{x},y)\sim \mathbb{P}({\cal X},C) }\ell (f(\mathbf{x}),y)+\int_{B} \sum_{i\in C} \ell(h(\mathbf{x};\boldsymbol{\omega}),i)P_i(\mathbf{x}) \psi(\mathbf{x}) {\rm d} {\mathbf x}.
\end{eqnarray}

The function $\psi(\mathbf{x})$ is bounded in ${\cal X}$, by hypothesis. Therefore, the integrand is a finite sum of bounded functions, so it is a bounded function, and there exists an $M>0$ such that $\sum_{i\in C} \ell(h(\mathbf{x};\boldsymbol{\omega}),i)P_i(\mathbf{x}) \psi(\mathbf{x})\le M$ for all ${\mathbf x}\in {\cal X}$. This allows us to bound the expression as follows:
\begin{eqnarray}
\nonumber R_D(\mathbf{\omega}_D)\le 
\mathbb{E}_{(\mathbf{x},y)\sim \mathbb{P}({\cal X},C) }\ell (f(\mathbf{x}),y)+ M \mu(B) \le \mathbb{E}_{(\mathbf{x},y)\sim \mathbb{P}({\cal X},C) }\ell (f(\mathbf{x}),y) +M\varepsilon_0.
\end{eqnarray}
As $\varepsilon_0$ is arbitrary and the inequality $\mathbb{E}_{(\mathbf{x},y)\sim \mathbb{P}({\cal X},C) }\ell (f(\mathbf{x}),y) \le R_D(\mathbf{\omega}_D)$ holds, the proof is complete.

\end{proof}

\section{Description of dataset variables}
\label{App:Dataset_variables}

Tables \ref{tab:dataset-variable-optima}, \ref{tab:dataset-variable-NTS}, and \ref{tab:dataset-variable-LPMC} provide a detailed description of all the variables that have been considered for the Optima, NTS, and LPMC datasets, respectively.

\begin{table}[!h]
\centering
\caption{Description of the variables utilised in the Optima dataset}
\label{tab:dataset-variable-optima}
\begin{tabular}{ll}
\toprule
Variable & Description \\ \hline
Choice & Choice variable: 0 (public transport), 1 (private mode), 2 (soft mode) \\
ID & Identifier of the individual \\
TimePT & Total duration of the travel performed in public transport (in minutes) \\
TimeCar & Total duration of the travel performed using the car (in minutes) \\
MarginalCostPT & Total cost in public transport taking into account the possible discounts \\
CostCarCHF & Total cost of a travel performed using the car (in Swiss francs) \\
distance\_km & Total distance (in kilometers) \\
age & Age of the individual \\
NbChild & Number of kids in the household \\
NbCar & Number of cars the household \\
NbMoto & Number of motorbikes in the household \\
NbBicy & Number of bikes in the household \\
Gender$^\dag$ & Gender of the individual: 1 (man), 2 (woman), 3 (unreported) \\
OccupStat$^\dag$ & Occupational status of the individual: 1 (full-time), 2-8 (not full-time) \\ \bottomrule
\multicolumn{2}{l}{$\dag$ These variables have been processed as one-hot encoding in the final dataset.}
\end{tabular}
\end{table}
\begin{table}[!h]
\centering
\caption{Description of the variables utilised in the NTS dataset}
\label{tab:dataset-variable-NTS}
\begin{tabular}{ll}
\toprule
Variable & Description \\ \hline
mode\_main & Choice variable: 0 (walk), 1 (bike), 2 (public transport), 3 (car) \\
individual\_id & Identifier of the individual inferred from the socio-economic and geographic data \\
distance & Total distance (in kilometers) \\
weekend & Is the trip done at the weekend?: 1 (yes), 0 (no) \\
age & Age of the individual \\
education & Education level of the individual: 1 (lower), 2 (middle), 3 (higher) \\
ethnicity$^\dag$ & Ethnicity of the individual: Native, Nonwestern, Western \\
license & Does the individual owns a driver's license?: 1 (yes), 0 (no) \\
male & Is the individual a male?: 1 (yes), 0 (no) \\
bicycles & Number of bikes in the household \\
cars & Number of cars in the household \\
income & Total net annual household income in $1,000$\euro: 2(less20), 3 (20to40), 4 (more40) \\
density & Density of the area where the individual lives (in $1,000$ addresses per kilometer$^2$) \\
diversity & Shannon diversity index of land use classes \\
green & Proportion of green space per post code area in $\%$ \\
precip & Daily precipitation amount in millimeters \\
temp & Daily maximum temperature in degrees Celsius \\
wind & Daily average wind speed in meters per second. \\ \bottomrule
\multicolumn{2}{l}{$\dag$ These variables have been processed as one-hot encoding in the final dataset.}
\end{tabular}
\end{table}

\begin{table}[!h]
\centering
\caption{Description of the variables utilised in the LPMC dataset}
\label{tab:dataset-variable-LPMC}
\begin{tabular}{ll}
\toprule
Variable & Description \\ \hline
travel\_mode & Choice variable: 0 (walk), 1 (bike), 2 (public transport), 3 (car) \\
household\_id & Identifier of the household \\
day\_of\_week & Day of the week when the trip took place, from 1 (Monday) to 7 (Sunday) \\
start\_time\_linear & Linearised trip start time from 0 to 24 (in hours) \\
age & Age of the individual \\
female & Is the individual female?: 1 (yes), 0 (no) \\
driving\_license & Does the individual owns a driver's license?: 1 (yes), 0 (no)  \\
car\_ownership & Number of cars in the household: 0 (no cars), 1 (less than one car per adult), \\
  &              2 (one or more cars per adult)  \\
distance & Total straight line distance (in meters) \\
dur\_walking & Predicted duration of walking route (in hours) \\
dur\_cycling & Predicted duration of cycling route (in hours) \\
dur\_pt\_access & Predicted duration of access and egress from public transport route (in hours) \\
dur\_pt\_rail & Predicted duration of rail in-vehicle time for public transport route (in hours) \\
dur\_pt\_bus & Predicted duration of bus in-vehicle time for public transport route (in hours) \\
dur\_pt\_int\_waiting & Predicted duration of waiting times on platform for public transport (in hours) \\
pt\_n\_interchanges & Number of interchanges on public transport route \\
dur\_driving & Predicted duration of driving route (in hours) \\
cost\_transit & Estimated cost of public transport route (in GBP) \\
cost\_driving\_total & Estimated cost of driving route, including fuel and congestion charge (in GBP) \\
purpose$^\dag$ & Purpose of the trip: B (employers' business), HBE (home-based education), \\
  &              HBO (home-based other), HBW (home-based work), NHBO (non-home-based \\
  &              other) \\
fueltype$^\dag$ & Fueltype of the vehicle: Petrol, Diesel, Hybrid, Average \\ \bottomrule
\multicolumn{2}{l}{$\dag$ These variables have been processed as one-hot encoding in the final dataset.}
\end{tabular}
\end{table}

\section{Hyperparameter Optimisation}
\label{App:HPO}
Currently, there are multiple software tools dedicated to hyperparameter optimisation. They usually depend on the programming language employed, the \gls{ml} algorithms used and the platforms on which they are run, among others. In this paper, the Hyperopt-Sklearn \citep{Komer2014} package, a software project for hyperparameter tuning designed for Python developers, has been used. More specifically, this software provides automatic adjustment of the hyperparameter values of any \gls{ml} algorithm of the Scikit-learn Python \gls{ml} library. It provides an easy interface to integrate Scikit-learn models into Hyperopt optimisation routines. When Hyperopt-Sklearn is used, the practitioner must define three components in order to formulate the \gls{hpo} problem. 

\begin{itemize}
  \item \textbf{Search space:} This allows the search domain for each hyperparameter of a \gls{ml} model to be defined. Thus, a hyperparameter is considered as a variable. For each of them, Hyperopt allows their type and their value to be defined. Regarding the type, a hyperparameter can be defined as integer, ordinal or categorical type. Concerning value, it is defined by means of a random variable which follows a probability distribution specified by the practitioner (uniform, log-scaling...). Furthermore, Hyperopt also allows a conditional structure to be defined in such a way that a hyperparameter takes a value from a given set. Table \ref{tab:hyperparameters-space} summarises the search spaces of each hyperparameter of the algorithms employed in this paper to solve the \gls{hpo} problem. 
  \item \textbf{Objective function:} This corresponds to a performance metric which provides the goodness of fit of the hyperparameters. Some commonly used metrics are accuracy, f1-score, cross-entropy loss, area under the ROC curve, etc. This measure is computed on held-out examples of the dataset to obtain a validation metric. Then, cross-validation is employed to obtain a more robust performance measure. In this study, $5-$fold CV is considered within the objective function.
  \item \textbf{Optimisation algorithm:} Although Hyperopt-Sklearn has been implemented to incorporate Bayesian optimisation algorithms based on Gaussian processes as well as regression trees, there is no algorithm based on these approaches implemented in the current version of the software. Three algorithms are currently available: i)~Random Search, ii)~\gls{tpe} and iii)~Adaptive \gls{tpe}. In this work, \gls{tpe} algorithm has been used due to its good results reported in the literature (see \cite{Komer2014} and \cite{Komer2019}). The interested reader is invited to consult \cite{Bergstra2011} for more details on the implementation of the \gls{tpe} algorithm. Within the same reference, a formal proof of the capabilities of the \gls{tpe} algorithm can be found, showing its superior performance compared to random search methods.
\end{itemize}

\vspace{-0.7\baselineskip}
\begin{table}[h]
\caption{Hyperparameter space of the \gls*{ml} models}
\label{tab:hyperparameters-space}
\resizebox{\textwidth}{!}{
\begin{tabular}{llllllll}
\toprule
{\bf Technique} $\mathcal{A}$ &{\bf Name of the hyperparameter } &{\bf Notation} &{\bf Type} &{\bf Search space} &{\bf OPTIMA} &{\bf NTS} & {\bf LPMC}\\
\midrule
  
\multirow{4}{1cm}{SVM} & Kernel function & $K$ & Fixed & $[\textrm{RBF}]$ & $\textrm{RBF}$ & $\textrm{RBF}$ & $\textrm{RBF}$ \\
& Samples to approximate Kernel map & $n_c$ & Uniform distribution & $[1,DatasetRows]$ & $2243$ & $228302$ & $80276$\\
& Parameter of the Gaussian function & $\gamma$ & Loguniform distribution & $[10^{-3}, 1]$ & $0.019$ & $0.008$ & $0.007$\\
& Cost (or soft margin constant) &$C$ & Loguniform distribution &  $[0.1, 10]$ & $1.475$ & $7.999$ & $6.591$\\
\hline
\multirow{5}{1cm}{RF} & Number of decision trees & $B$ & Uniform distribution  & $[1, 200]$ & $128$ & $153$ & $180$\\
& Max features for the best split&$m$ & Uniform distribution & $[2, \textrm{Nº features}]$ & $9$ & $8$ & $16$\\
& Max depth of the tree &$d$ & Uniform distribution & $[3,10]$ & $10$ & $10$ & $10$\\
& Min samples to be at a leaf node &$l$ & Uniform distribution & $[1,20]$ & $3$ & $3$ & $11$\\
& Min samples to split an internal node &$s$ & Uniform distribution & $[2,20]$ & $12$ & $15$ & $14$\\
& Goodness of split metric &$c$ & Choice & $[\textrm{Gini}|\textrm{Entropy}]$ & Gini & Entropy & Entropy\\
\hline
\multirow{10}{1cm}{XGBoost} & Maximum tree depth & $d$ & Uniform distribution & $[1,14]$ & $5$ & $7$ & $7$\\
& Minimum loss for a new split & $\gamma$ & Loguniform distribution & $[10^{-4},5]$ & $0.174$ & $4.970$ & $4.137$\\
& Minimum sum of instance weight needed in a child & $w$ & Uniform distribution & $[1,100]$ & $3$ & $1$ & $32$\\
& Maximum delta step in each tree’s weight & $\delta$ & Uniform distribution & $[0,10]$ & $2$ & $0$ & $4$\\
& Subsample ratio of the training instance & $s$ & Uniform distribution & $[0.5,1]$ & $0.581$ & $0.823$ & $0.935$\\
& Subsample ratio of columns when constructing each tree & $c_t$ & Uniform distribution & $[0.5,1]$ & $0.949$ & $0.553$ & $0.679$\\
& Subsample ratio of columns for each level & $c_l$ & Uniform distribution & $[0.5,1]$ & $0.633$ & $0.540$ & $0.629$\\
& L1 regularisation term on weights & $\alpha$ & Loguniform distribution & $[10^{-4},10]$ & $0.097$ & $0.028$ & $0.003$\\
& L2 regularisation term on weights & $\lambda$ & Loguniform distribution & $[10^{-4},10]$ & $0.690$ & $0.264$ & $0.5 \cdot 10^{-3}$\\
& Number of boosting rounds & $B$ & Uniform distribution & $[1,6000]$ & $732$ & $4376$ & $2789$\\
\hline
\multirow{8}{1cm}{NN} & Number of neurons in hidden layer & $n_1$ & Uniform distribution & $[10, 500]$ & 195 & 10 & 51\\
& Activation function & $f$ & Fixed & $[\textrm{tanh}]$ & $\textrm{tanh}$ & $\textrm{tanh}$ & $\textrm{tanh}$\\
& Solver for weights optimisation & $S$ & Choice & $[\textrm{LBFGS} |\textrm{SGD}|\textrm{Adam}]$ & $\textrm{SGD}$ & $\textrm{LBFGS}$ & $\textrm{SGD}$\\
& Initial learning rate &$\eta_{0}$ & Uniform distribution & $[10^{-4}, 1]$ & 0.521 & 0.416 & 0.041\\
& Learning rate schedule & $\eta$ & Fixed & $[\textrm{adaptive}]$ & $\textrm{adaptive}$ & $\textrm{adaptive}$ & $\textrm{adaptive}$\\
& Maximum number of iterations & $t$ & Fixed & $[10^6]$ & $10^6$ & $10^6$ & $10^6$\\
& Batch Size & $BS$ & Choice & $[128|256|512|1024]$ & $1024$ & $512$ & $1024$\\
& Tolerance for optimisation & $tol$ & Fixed & $[10^{-3}]$ & $10^{-3}$ & $10^{-3}$ & $10^{-3}$\\
\hline
\multirow{7}{1cm}{DNN} & Input dimension &$i$ & Fixed & $\textrm{Nº features}$ & $7$ & $16$ & $35$\\
& Output dimension &$o$ & Fixed & $\textrm{Nº Classes}$ & $3$ & $4$ & $4$\\
& Number of hidden layers &$depth$ & Choice & $[2|3|4|5|6|7|8|9|10]$ & $2$ & $2$ & $2$\\
& Number of neurons in hidden layers &$width$ & Choice & $[25|50|100|150|200]$ & $50$ & $25$ & $25$\\
& Dropout rate &$drop$ & Choice & $[0.1|0.01|10^{-5}]$ & $10^{-5}$ & $0.1$ & $0.1$\\
& Epochs & $e$ & Uniform distribution & $[50,200]$   & $174$ & $114$ & $122$\\ 
& Batch size &$BS$ & Choice & $[128|256|512|1024]$ & $1024$ & $256$ & $1024$\\
\bottomrule
\end{tabular}
}
\end{table}

\vspace{-1.1\baselineskip}

\section{Willingness to pay with Cobb-Douglas-type utilities}
\label{App:WTP-CD}
This appendix illustrates how to derive the \gls{wtp} for the Cobb-Douglas utility functions of the form
\begin{equation}
  V(x,I)= \left (x^{\beta_x} \right ) \left ( I^{\beta_I} \right ),
\end{equation}
where the attributes of the individuals in the population are uniformly distributed in a square ${\cal X}=[0,1] \times [0,1]$, as shown in Figure \ref{fig:wtp_CD}.

\begin{figure}[h]
  \centering
  \begin{tikzpicture}
  \draw[line width=0.3mm,->, >=latex] (0,0) -- (0,4.5);
  \draw[line width=0.3mm,->, >=latex] (0,0) -- (4.5,0);
  \node at (3.5,0.5) {\Large $\mathbf{\cal X}$};
  \node at (4,-0.5) {\Large $\mathbf{x}$};
  \node at (-0.5,4) {\Large $\mathbf{I}$};
  \node at (0,0) {$\bullet$};
  \node at (4,4) {$\bullet$};
  \node at (4,2) {$\bullet$};
  \node at (2,4) {$\bullet$};
  \node at (2,4.5) {$\gls{wtp}_n=z> \beta$};
  \node at (4.5,4.5) {$\gls{wtp}_n= \beta$};
  \node at (5.5,2) {$\gls{wtp}_n=z <\beta$};
  \draw [line width=0.5mm] (0,0) -- (4,0) -- (4,4) -- (0,4)--(0,0);
  \draw[line width=0.5mm,-, >=latex] (0,0) -- (4,4);
  \draw[line width=0.5mm,-, >=latex] (0,0) -- (4,2);
  \draw[line width=0.5mm,-, >=latex] (0,0) -- (2,4);
	\end{tikzpicture}
	\caption{Contour lines of the $\gls{wtp}_n$ \label{fig:wtp_CD}}
\end{figure}

In this scenario, the value of \gls{wtp} for an individual $\mathbf{x}_{ni}=(x_n,I_n)$ is given by the expression
\begin{equation}
  \gls{wtp}_n=\beta \dfrac{I_n}{x_n},
\end{equation}
\noindent where $\beta= \dfrac{\beta_x}{ \beta_I}$. As $\mathbf{x}_n$ is a random variable, we have to compute the probability distribution function $\gls{wtp}_n$. Note that the value of \gls{wtp} is invariant in the straight lines $I=\alpha x$ and can be obtained as
\begin{equation}
  \gls{wtp}_\alpha =\beta \dfrac{I_n}{x_n}= \beta \dfrac{\alpha I_n}{I_n}= \alpha \beta.
\end{equation}

Also note that if $\alpha'< \alpha \Longrightarrow \gls{wtp}_{\alpha'} < \gls{wtp}_\alpha$.

\vspace{0.7cm}

In order to find $\mathbb{P} (\gls{wtp}_n \le z)$, consider two cases:
\begin{itemize}
  \item[{\bf I:}] Let $z \le \beta$. Then, there exists an $\alpha$ with $0\le \alpha \le 1$ satisfying the condition $\alpha \beta = z$. Hence,
  \begin{equation*}
    \mathbb{P} (\gls{wtp}_n \le z)= \mathbb{P} (\gls{wtp}_n \le \alpha \beta )=\mathbb{P} (\gls{wtp}_n \le \gls{wtp}_\alpha )= \dfrac{1}{2} \alpha= \dfrac{z}{2\beta}.
  \end{equation*}

  \item[{\bf II:}] Let $\beta \le z $. Then, there exists an $\alpha$ with $1 \le \alpha $ satisfying the condition $\alpha \beta = z$. Hence,
  \begin{eqnarray*}
    \mathbb{P} (\gls{wtp}_n \le z)= \mathbb{P} (\gls{wtp}_n \le \beta )+ \mathbb{P} (\beta \le \gls{wtp}_n \le z)= \dfrac{1}{2}+ \int_0 ^1 \int_{\frac{I}{\alpha}} ^I 1 \rm{dx} \rm{dI}= \\
    \dfrac{1}{2}+ \int_0 ^1 \left (I-\frac{I}{\alpha} \right ) \rm{dI}= 1- \dfrac{\beta }{2z}.
  \end{eqnarray*}
\end{itemize}
These two cases are represented graphically in Figure \ref{fig:domain_region_z}.

\begin{figure}[h]
	\centering
	\subfigure[Case I: $z \le \beta$]{
  \begin{tikzpicture}
    \draw [fill=gray] (0.2,4.6) -- (0.4,4.6) -- (0.4,4.4) --(0.2,4.4)--(0.2,4.4);
    \draw[line width=0.3mm,->, >=latex] (0,0) -- (0,4.5);
    \draw[line width=0.3mm,->, >=latex] (0,0) -- (4.5,0);
    \node at (4,-0.5) {\Large $\mathbf{x}$};
    \node at (-0.5,4) {\Large $\mathbf{I}$};
    \draw [line width=0.5mm] (0,0) -- (4,0) -- (4,4) -- (0,4)--(0,0);
    \node at (0,0) {$\bullet$};
    \node at (4,2) {$\bullet$};
    \node at (0,2) {$\bullet$};
    \node at (-0.5,2) { $\alpha$};
    \draw [line width=0.25mm] (0,2) -- (4,2);
    \node at (2.5,4.5) {Region where $\gls{wtp}_n <z$};
    \draw [fill=gray] (0,0) -- (4,0) -- (4,2) --(0,0);
    \draw[line width=0.5mm,-, >=latex] (0,0) -- (4,2);
	\end{tikzpicture}
	}
	\subfigure[Case II: $z \ge \beta$]{
  \begin{tikzpicture}
    \draw [fill=gray] (0.2,4.6) -- (0.4,4.6) -- (0.4,4.4) --(0.2,4.4)--(0.2,4.4);
    \draw [fill=gray] (0,0) -- (4,0) -- (4,4) --(2,4)--(0,0);
    \draw[line width=0.3mm,->, >=latex] (0,0) -- (0,4.5);
    \draw[line width=0.3mm,->, >=latex] (0,0) -- (4.5,0);
    \node at (4,-0.5) {\Large $\mathbf{x}$};
    \node at (-0.5,4) {\Large $\mathbf{I}$};
    \node at (3,-0.5) { $I$};
    \node at (3/2,-0.5) { $\frac{I}{\alpha}$};
    \node at (-0.5,3) { $I$};
    \draw [line width=0.5mm] (0,0) -- (4,0) -- (4,4) -- (0,4)--(0,0);
    \node at (0,0) {$\bullet$};
    \node at (4,4) {$\bullet$};
    \node at (2,4) {$\bullet$};
    \node at (3,3) {$\bullet$};
    \node at (3/2,3) {$\bullet$};
    \node at (2.5,4.5) {Region where $\gls{wtp}_n <z$};
    \draw[line width=0.5mm,-, >=latex] (0,0) -- (4,4);
    \draw[line width=0.5mm,-, >=latex] (0,0) -- (2,4);
    \draw[line width=0.25mm,-., >=latex] (0,3) -- (3,3);
    \draw[line width=0.25mm,-., >=latex] (3,0) -- (3,3);
    \draw[line width=0.25mm,-., >=latex] (3/2,0) -- (3/2,3);
	\end{tikzpicture}
	}
	\caption{Graphical representation of the domain region where $\gls{wtp}_n \le z $}
    \label{fig:domain_region_z}
\end{figure}

Hence, the probability distribution function of $\gls{wtp}_n$ is:
\begin{eqnarray*}
  F(z)= \left \{
  \begin{array}{ll}
  \frac{z}{2\beta} & \hbox{if } z \le \beta \\
  1-\frac{\beta}{2z} & \hbox{if } z \ge \beta \\
  \end{array}
  \right. ,
\end{eqnarray*}
and, taking the derivative of the previous formula, the probability density function is:
\begin{eqnarray*}
  f(z)= \left \{
  \begin{array}{ll}
  \frac{1}{2\beta} & \hbox{if } z \le \beta \\
  \frac{\beta}{2z^2} & \hbox{if } z \ge \beta \\
  \end{array}
  \right. .
\end{eqnarray*}

Finally, note that
\begin{eqnarray*}
  \mathbb{E} \left (\gls{wtp}_n \right ) = \int_0^{+\infty} z f(z) {\rm dz}=+\infty, \\
  F(z)=\frac{1}{2} \Longrightarrow \hbox{median } \left ( \gls{wtp}_n \right ) = \beta=\dfrac{\beta_x}{ \beta_I}.
\end{eqnarray*}

\end{document}